\documentclass[preprint,3p,10.5pt]{elsarticle}
\usepackage{graphicx} 
\usepackage{amsmath}
\usepackage{amssymb}
\usepackage{amsfonts}
\usepackage{amsthm}
\usepackage{lipsum}
\usepackage{textcomp}
\usepackage{subcaption}
\usepackage{color}
\usepackage[normalem]{ulem}
\usepackage{booktabs}
\usepackage{multirow}
\usepackage{comment}
\usepackage{acro}
\usepackage{tikz}
\usepackage{caption}  

\usetikzlibrary{positioning, arrows.meta, decorations.pathreplacing}  

\DeclareAcronym{UQ}{
  short = UQ ,
  long = Uncertainty Quantification
}

\DeclareAcronym{MCMC}{
  short = MCMC,
  long = Markov Chain Monte Carlo
}

\DeclareAcronym{FOM}{
short = FOM,
long = Full Order Model
}
\DeclareAcronym{DA}{
    short = DA,
    long = Delayed Acceptance
}

\DeclareAcronym{MLDA}{
    short = MLDA,
    long = Multi-Level Delayed Acceptance
}

\DeclareAcronym{MFDA}{
short = MFDA,
long = Multi-Fidelity Delayed Acceptance 
}

\DeclareAcronym{MH}{
short = MH,
long = Metropolis Hastings 
}

\DeclareAcronym{PDE}{
  short = PDE,
  long = Partial Differential Equation,
  short-plural = s,  
  long-plural = s
}

\DeclareAcronym{FEM}{
short = FEM,
long = Finite Element Method
}

\DeclareAcronym{ESS}{
    short=ESS,
    long=Effective Sample Size,
}

\DeclareAcronym{ODE}{
    short=ODE,
    long=Ordinary Differential Equation,
    short-plural=s,
    long-plural=s
}

\DeclareAcronym{MEMS}{
    short = MEMS,
    long = Micro-Electro-Mechanical System,
    short-plural = ,  
    long-plural = s
}

\DeclareAcronym{LS}{
    short = LS,
    long = Least-Squares 
}

\DeclareAcronym{NN}{
short = NN,
long = Neural Network,
short-plural=s,
long-plural=s
}

\DeclareAcronym{MSE}{
short = MSE,
long = Mean Squared Error
}

\DeclareAcronym{RMSE}{
short = RMSE,
long = Root Mean Squared Error
}

\DeclareAcronym{KL}{
short = KL,
long = Karhunen–Loève
}

\DeclareAcronym{VI}{
short = VI,
long = Variational Inference
}

\usepackage[colorlinks=true,linkcolor=black,anchorcolor=black,citecolor=black,filecolor=black,menucolor=black,runcolor=black,urlcolor=black]{hyperref} 
\usepackage{cleveref}
\usepackage{natbib} 
\usepackage{algorithm}
\usepackage{algorithmic}

\begin{document}

\begin{frontmatter}

    \title{
    Multi-Fidelity Delayed Acceptance: \\  hierarchical MCMC sampling for Bayesian inverse problems \\ combining multiple solvers through deep neural networks}
    
    \author[label1]{Filippo Zacchei\corref{cor1}}
    \ead{filippo.zacchei@polimi.it}
    \author[label3,label1]{Paolo Conti}
    \author[label2]{Attilio Frangi}
    \author[label1]{Andrea Manzoni}

    \cortext[cor1]{Corresponding author: Filippo Zacchei}

    \address[label1]{Politecnico di Milano, MOX - Dept.\ of Mathematics, p.za Leonardo da Vinci, 32, Milano, 20133, Italy}
    \address[label2]{Politecnico di Milano, Dept.\ of Civil and Environmental Engineering, p.za Leonardo da Vinci, 32, Milano, 20133, Italy}
    \address[label3]{The Alan Turing Institute, London, NW1 2DB, UK}

\begin{abstract}
Inverse uncertainty quantification (UQ) tasks such as parameter estimation are computationally demanding whenever dealing with physics-based models, and typically require repeated evaluations of complex numerical solvers. When partial differential equations are involved, full-order models such as those based on the Finite Element Method can make traditional sampling approaches like Markov Chain Monte Carlo (MCMC) computationally infeasible. Although data-driven surrogate models may help reduce evaluation costs, their utility is often limited by the expense of generating high-fidelity data. In contrast, low-fidelity data can be produced more efficiently, although relying on them alone may degrade the accuracy of the inverse UQ solution.

To address these challenges, we propose a Multi-Fidelity Delayed Acceptance scheme for Bayesian inverse problems involving large-scale physics-based models. Extending the Multi-Level Delayed Acceptance framework, the method introduces multi-fidelity neural networks that combine the predictions of solvers of varying fidelity, with high-fidelity evaluations restricted to an offline training stage. During the online phase, likelihood evaluations are obtained by evaluating the coarse solvers and passing their outputs to the trained neural networks, thereby avoiding additional high-fidelity simulations.

This construction allows heterogeneous coarse solvers to be incorporated consistently within the hierarchy, providing greater flexibility than standard Multi-Level Delayed Acceptance. The proposed approach improves the approximation accuracy of the low-fidelity solvers, leading to longer sub-chain lengths, better mixing, and accelerated posterior inference. The effectiveness of the strategy is demonstrated on two benchmark inverse problems involving (i) steady isotropic groundwater flow, (ii) an unsteady reaction–diffusion system, for which substantial computational savings are obtained.
\end{abstract}


\end{frontmatter}




\section{Introduction}
\label{sec:introduction}

Parameter estimation is a crucial aspect of many engineering applications, playing a key role in the design, optimization, and control of complex systems. 
In several fields ranging from, e.g., fluid dynamics 
\cite{LykkegaardGWF2021,Willcox2008} and climate systems 
\cite{CO2_2023} to civil structures \cite{Rosafalco2021} and 
microsystems \cite{Durr_2023,ZaccheiMEMS2024}, the accuracy and
reliability of parameter estimates directly influence systems' 
performances or even safety. Indeed, despite advanced numerical
simulations have tremendously improved our ability to predict 
the behavior of these systems, mathematical models often 
involve \acp{PDE} 
\cite{LykkegaardGWF2021,Willcox2008,conti_multi-fidelity_2024}
which can be computationally intensive to solve, thus making 
many-query scenarios, like \ac{UQ} tasks, prohibitive 
\cite{Willcox2018}.

To address \ac{UQ} in parameter estimation (or inverse \ac{UQ}) a variety of methods have been developed. Bayesian approaches are particularly appealing, as they provide a principled way to update prior beliefs on the parameters using observed data, yielding a posterior distribution that reflects all sources of uncertainty. Since this posterior is rarely available in closed form, a range of computational techniques have been developed to obtain a reliable approximation. Among these, \ac{MCMC} methods \cite{Kaipio2005,geyerIntroductionMarkovChain2011,tarantola2005} provide the most general option among sampling-based methods, as they asymptotically generate samples from the posterior under mild regularity conditions. Other approaches include, for instance, Importance Sampling \cite{tokdar_importance_2010}, which estimates expectations by reweighting samples from a proposal distribution, Sequential Monte Carlo methods \cite{doucet2001} that rely on ensembles of weighted particles to explore evolving posterior distributions, as well as Kalman filters and their nonlinear extensions \cite{rosafalco2024ekf,Simon2006}, that are widely used for recursive Bayesian updates in sequential or state-space models. 

While more computationally intensive than frequentist techniques such as maximum likelihood estimation, 
Bayesian methods offer significant advantages: they yield richer probabilistic information, including credibility intervals, sensitivity to outliers, and the ability to capture complex posterior features such as multi-modality and skewness, thus providing a more robust framework for \ac{UQ} \cite{schoot2014}. 

\ac{VI} \cite{blei2017} has emerged as a scalable alternative, formulating posterior approximation as an optimization problem. Although efficient, \ac{VI} may introduce bias due to the restricted
expressiveness of the variational family. In contrast, \ac{MCMC} methods impose fewer assumptions but
require many forward model evaluations, leading to slow convergence when models are expensive.

To mitigate these limitations, we propose a novel sampling scheme, which we refer to as \ac{MFDA}. This method accelerates \ac{MCMC}-based inference by integrating surrogate models of varying fidelity via multi-fidelity fusion \cite{Willcox2018} using \acp{NN} regression, and by incorporating filtering \cite{Willcox2018} strategies inspired by the \ac{MLDA} framework \cite{lykkegaard_multilevel_2023}. 

\subsection{Existing approaches: multi-level methods}

The classical \ac{MCMC} algorithm explores the parameter space through an \textit{outer loop} \cite{Willcox2018} process: at each iteration, a new candidate parameter is proposed based on the previous sample, and the forward model is evaluated to compute the likelihood of the candidate. The parameter is then accepted (and treated as being a sample drawn from the target posterior distribution) or rejected. 
Therefore, the effectiveness of \ac{MCMC} sampling is based fundamentally on {\em (i)} the rapid and accurate evaluation of the likelihood function and {\em (ii)} the efficient exploration of the parameter space. 

The rapid evaluation of the likelihood function can be addressed using low-fidelity surrogates as approximations to accelerate the evaluation process. 
While replacing the high-fidelity model ($\mathbf{f}_\text{HF}$) with a low-fidelity counterpart ($\mathbf{f}_\text{LF}$) substantially reduces computational efforts, this might come with a loss in accuracy that can be quite hard to assess, as it usually depends on the characteristics of the specific surrogate model in use \cite{pagani2017efficient,HU2024112970}.  Data-fit surrogates can provide accurate approximations but require substantial high-fidelity data for training \cite{ZaccheiMEMS2024,alarra2023,OLEARYROSEBERRY2024112555}, which may be impractical. 

Regarding the exploration of the parameter space, \ac{MCMC} samples often exhibit high autocorrelation,  leading to an effective number of samples being only a small fraction of the total number of forward model evaluations \cite{lykkegaard_multilevel_2023,gelman2013bayesian,cui2014}. Several strategies have been proposed to mitigate this limitation, such as, e.g, 
updating the proposal distribution using the information obtained during the process, as implemented in the adaptive Metropolis algorithm \cite{bj1080222083}, or 
estimating gradient information from the forward model, as in the Metropolis-adjusted Langevin algorithm \cite{girolami2011riemann}, Hamilton Monte Carlo \cite{Betancourt2018} or No-U-Turn-Sampler \cite{Hoffman2011}. However, these methods often pose a series of computational challenges. For instance, the required gradient information can be expensive to obtain \cite{cao2025derivative}, thereby limiting their applicability in complex or high-dimensional settings.

The idea of combining different solvers to enhance sampling-based Bayesian inference has led to several strategies, among which multi-level methods \cite{giles2015multilevel,beskos2017multilevel} have gained substantial attention in the last decade. These techniques were originally introduced in the context of forward UQ, leveraging hierarchies of numerical discretizations, typically generated by varying mesh resolution in a single solver, to balance accuracy and computational cost \cite{cliffe2011multilevel}.
In this setting, a high-fidelity model is used to ensure the required accuracy, while coarser discretizations offer inexpensive but less accurate approximations.

Multi-level strategies have subsequently been adapted to improve the efficiency of sampling methods in inverse UQ, primarily through filtering-based schemes. One prominent class involves Delayed Acceptance algorithms \cite{Fox2005,cao2025derivative}, where low-cost models are used to discard unlikely parameter proposals before evaluating them with high-fidelity simulations. Strategies like Multi-level \ac{MCMC} \cite{dodwell2015} and \ac{MLDA}\cite{lykkegaard_multilevel_2023} further extend this idea using chains that pass through models of increasing fidelity, retaining proposals only if they pass acceptance tests at all levels: by doing so, the number of expensive model evaluations is reduced by rejecting poor candidates earlier in the chain. 

However, the efficiency of \ac{MLDA} is highly sensitive to the consistency of the posterior approximations across fidelity levels. When the coarse forward models introduce non-negligible bias, proposals that appear acceptable at coarse levels may be rejected at finer levels, resulting in poor mixing. This issue is particularly pronounced when the coarse models differ from the high-fidelity model not only in discretization but also in physical representation or numerical formulation. Consequently, \ac{MLDA} faces a fundamental limitation: one must either use very short sub-chain lengths, shifting a significant portion of the computational budget back to the high-fidelity model, or employ more accurate (and therefore more expensive) coarse models, which reduces the computational advantage of the multi-level approach.

To improve the alignment between fidelity levels, multi-fidelity fusion methods can be employed. Unlike filtering-based approaches, fusion methods evaluate multiple models simultaneously and combine their outputs to produce better predictions.
Classical techniques include control variates \cite{bratley_guide_1987,hammersley_monte_1964,nelson_control_1987} and
co-kriging \cite{annels_geostatistical_1991,myers_matrix_1982,perdikaris_multi-fidelity_2015},
although these approaches may scale poorly with problem dimension. In contrast \acp{NN} have proven effective for multi-fidelity fusion in complex, high-dimensional problems. Their capacity to learn nonlinear mappings between low- and high-fidelity models makes them well-suited for improving the consistency of surrogate-based approximations \cite{meng_composite_2020,Meng_2021,liu_multi-fidelity_2019,motamed_multi-fidelity_2020,Guo_2022,krouglova2025multifidelity}.

\subsection{Our proposed strategy: multi-fidelity delayed acceptance MCMC}

In this work, we build upon the filtering structure of the \ac{MLDA} framework and extend it by incorporating a multi-fidelity fusion strategy based on \acp{NN}. The proposed approach consists of two stages. In the offline stage, the \acp{NN} are trained to learn a corrective mapping that reduces the discrepancy between the coarse solvers and the high-fidelity model. This is the only stage in which evaluations of the high-fidelity solver are required. 

In the online stage, we perform multi-level sampling while retaining the  filtering structure of \ac{MLDA}, but using only the coarse solvers. At each fidelity level, the outputs of all coarser solvers up to that level are provided as input to the corresponding \ac{NN}, which produces a corrected prediction. This improves the accuracy of the lowest fidelity levels and ensures that the finest level in the hierarchy provides an approximation that is sufficiently consistent with the high-fidelity model.

We refer to this enhanced approach as \ac{MFDA}. By leveraging the structure and correlation across models of varying fidelity, \ac{MFDA} provides a scalable tradeoff between two extremes: purely data-driven surrogates, which incur high offline training costs but offer fast online evaluations, and solver-based \ac{MLDA} schemes, which require no offline phase but heavily rely on costly online high fidelity computations during sampling.

The proposed approach offers several advantages. First, by improving the consistency between posterior approximations across fidelity levels, it enables longer sub-chains at coarse levels, reduces sample autocorrelation, and increases the overall sampling efficiency. Second, the corrective capability of the multi-fidelity \acp{NN} allows the use of lower-cost and less accurate coarse models that would otherwise lead to poor performance in standard \ac{MLDA} schemes. Third, the numerical results indicate that the \acp{NN} perform best when trained on a hierarchy of coarse model outputs, rather than relying solely on the most accurate surrogate, highlighting the benefit of exploiting correlations across multiple fidelity levels.

To our knowledge, the proposed \ac{MFDA} framework represents one of the first systematic integration of multi-fidelity data fusion and filtering strategies within an \ac{MCMC}-based approach for inverse uncertainty quantification.

The structure of the paper is as follows. Section \ref{sec:met1} presents the backward \ac{UQ} framework based on \ac{MCMC} techniques, beginning with standard methods and progressing to the \ac{MLDA} approach. This is followed by the introduction of multi-fidelity \acp{NN} and the development of the proposed \ac{MFDA} scheme. Sections \ref{sec:res1} and \ref{sec:res2} detail the results of two numerical experiments offering an empirical assessment of all the listed advantages of \ac{MFDA}. Finally, Section \ref{sec:conclusion} concludes the work and outlines prospective avenues for further research.


\section{Methodology}\label{sec:met1}
In this section, we review the foundations of Bayesian inverse problems and standard \ac{MCMC} sampling strategies, highlighting their limitations. We then introduce surrogate models and the \ac{MLDA} method,  finally showing how to extend this framework using \ac{NN}-based multi-fidelity fusion. This will yield our \ac{MFDA} scheme.

\subsection{Bayesian inverse problem for PDEs}

Inverse problems deal with the use of actual measurements or observational data to infer the properties of a system described by a mathematical model \cite{kirsch_introduction_2021,tarantola2005}. These properties are often encoded in a vector of input parameters $\boldsymbol{\theta}$, whereas the model $\mathcal{G}(\boldsymbol{\theta})$ allows us to express all the available knowledge about the way data can be explained in terms of the input parameters. Under the assumption of additive, independent noise, observations are linked to input parameters through the following relationship:
\begin{equation}
\label{eq:inverse_problem_1}
\boldsymbol{y}^{\text{obs}} = \mathcal{G}(\boldsymbol{\theta}) + \boldsymbol{\varepsilon},
\end{equation}
where $\mathcal{G}: \mathbb{R}^m \rightarrow \mathbb{R}^d$ denotes the parameter-to-observable map, while $\boldsymbol{\varepsilon}$ accounts for observational noise and model mis-specification.  
The task of computing $\mathcal{G}(\boldsymbol{\theta})$ for a known input $\boldsymbol{\theta}$ defines the so-called \textit{forward problem}, which very often involves the solution of a differential problem, in the form of either a system of \acp{ODE} or \acp{PDE}; in this work we focus on the latter. Usually, $\mathcal{G}(\boldsymbol{\theta})$ consists of a set of solution components, whenever mimicking, for instance, data collected at a set of sensors, or more general outputs depending on the \ac{PDE} solution and involving, e.g., spatial averages, fluxes, or derivatives.  
In this context, the input $\boldsymbol{\theta} \in \boldsymbol{\Theta} \subset \mathbb{R}^m$ typically denotes a vector of model coefficients affecting the differential operator, as well as boundary or initial conditions, and is referred to as the set of model \textit{parameters}.

A common instance of an inverse problem occurs when the model $\mathcal{G}$ is approximated by a high-fidelity simulator $\mathbf{f}_\text{HF}$, which provides a highly accurate numerical approximation to the \ac{PDE}, leading to the reformulation of \eqref{eq:inverse_problem_1} as:
\begin{equation}
\label{eq:inverse_problem_statistical}
\boldsymbol{y}^{\text{obs}} = \mathbf{f}_\text{HF}(\boldsymbol{\theta}) + \boldsymbol{\varepsilon},
\end{equation}
where $\boldsymbol{\varepsilon}$ is typically assumed to follow a zero-mean Gaussian with covariance $\boldsymbol{\Sigma}_\varepsilon \in \mathbb{R} ^{d\times d}$.
By casting the inverse problem in a Bayesian framework, the goal is to determine the so-called \textit{posterior distribution} \(\pi(\boldsymbol{\theta}|\boldsymbol{y}^{\text{obs}})\)
of the parameters $\boldsymbol{\theta}$ given  \(\boldsymbol{y}^{\text{obs}}\), which according to the \textit{Bayes theorem}  reads as:
\begin{equation}
\label{eq:BayesTheorem}
\pi(\boldsymbol{\theta}|\boldsymbol{y}^{\text{obs}} ) = \frac{\pi(\boldsymbol{y}^{\text{obs}} |\boldsymbol{\theta}) \pi(\boldsymbol{\theta})}{\pi(\boldsymbol{y}^{\text{obs}} )}.
\end{equation}
Here \(\pi(\boldsymbol{y}^{\text{obs}}|\boldsymbol{\theta})\) denotes the \textit{likelihood} of the data given the parameters, \(\pi(\boldsymbol{\theta})\) is the \textit{prior distribution} of the parameters, encoding all the available knowledge on $\boldsymbol{\theta}$ before acquiring the data,  and \(\pi(\boldsymbol{y}^{\text{obs}})\) is the \textit{marginal distribution} (or {\em evidence}), which plays the role of a normalizing constant, and is given by 
\begin{equation}
\label{eq:marginal}
\pi\left(\boldsymbol{y}^{\text{obs}}\right)=\int_{\boldsymbol{\Theta}} \pi(\boldsymbol{\theta}) \pi\left(\boldsymbol{y}^{\text{obs}} \mid \boldsymbol{\theta}\right) d \boldsymbol{\theta}.
\end{equation}

The integral in Eq.~\eqref{eq:marginal} cannot be computed analytically in general, so that a closed-form expression for the posterior distribution is therefore unavailable. 
A widely used strategy to approximate samples from the posterior distribution relies on \ac{MCMC} methods \cite{robert2011short}. 
These methods aim at generating a Markov chain in the input parameter space $\boldsymbol{\Theta}$, whose invariant distribution approximates the target posterior. Among the various \ac{MCMC} algorithms, the \ac{MH} scheme is one of the most widely used options \cite{hastings}. After  initializing the chain, \ac{MH} proceeds by generating candidate samples $\boldsymbol{\theta}'$, given the previous sample $\boldsymbol{\theta}$, from a proposal distribution $q(\cdot\mid\boldsymbol{\theta})$ and accepting or rejecting each candidate based on the acceptance probability
\begin{equation}
    \alpha(\boldsymbol{\theta}' \mid \boldsymbol{\theta}) = \min\left\{1, \frac{\pi(\boldsymbol{y}^{\text{obs}} \mid \boldsymbol{\theta}') \pi(\boldsymbol{\theta}')q(\boldsymbol{\theta}'\mid\boldsymbol{\theta})}{\pi(\boldsymbol{y}^{\text{obs}} \mid \boldsymbol{\theta}) \pi(\boldsymbol{\theta})q(\boldsymbol{\theta}\mid\boldsymbol{\theta}')} \right\}.
\end{equation}
This mechanism ensures convergence of the Markov chain to the target posterior distribution over successive iterations \cite{robert2011short}. The full procedure is reported in Algorithm~\ref{alg:MH}.

\begin{algorithm}[t]
\caption{Metropolis-Hastings (MH)}
\label{alg:MH}
    \hspace*{\algorithmicindent} \textbf{Input}:  Likelihood $\pi(\boldsymbol{y}^{\text{obs}}\mid\cdot)$, prior distribution $\pi(\cdot)$, proposal distribution $q(\cdot \mid \cdot)$, initial sample $\boldsymbol{\theta}_0$,  number of samples $N$ \\
    \hspace*{\algorithmicindent} \textbf{Output}: Chain of samples $\{\boldsymbol{\theta}_j\}_{j=1}^N.$
    \begin{algorithmic}[1]
    \FOR{$j = 1$ to $N$}
        \STATE Propose $\boldsymbol{\theta}' \sim q(\boldsymbol{\theta}' \mid \boldsymbol{\theta}_{j-1})$.
        \STATE Compute acceptance probability:
        \[
    \alpha(\boldsymbol{\theta}' \mid \boldsymbol{\theta}) = \min\left\{1, \frac{\pi(\boldsymbol{y}^{\text{obs}} \mid \boldsymbol{\theta}') \pi(\boldsymbol{\theta}')q(\boldsymbol{\theta}'\mid\boldsymbol{\theta})}{\pi(\boldsymbol{y}^{\text{obs}} \mid \boldsymbol{\theta}) \pi(\boldsymbol{\theta})q(\boldsymbol{\theta}\mid\boldsymbol{\theta}')} \right\}.\]
        \STATE Accept $\boldsymbol{\theta}'$ with probability $\alpha$; set $\boldsymbol{\theta}_j = \boldsymbol{\theta}'$ if accepted, otherwise $\boldsymbol{\theta}_j = \boldsymbol{\theta}_{j-1}$.
    \ENDFOR
\end{algorithmic}
\end{algorithm}

Evaluating the likelihood at each iteration requires solving the high-fidelity model $\mathbf{f}_{\text{HF}}$ as defined in Eq.~\eqref{eq:inverse_problem_statistical}. By assuming that $\boldsymbol{\varepsilon} \sim N({\bf 0}, \boldsymbol{\Sigma}_\varepsilon)$ 
 the likelihood takes the form:
\begin{equation}
\pi(\boldsymbol{y}^{\text{obs}} \mid \boldsymbol{\theta}) \propto \exp \left( -\frac{1}{2} \left\| \boldsymbol{\Sigma}_{\boldsymbol{\varepsilon}}^{-\frac{1}{2}} \left(\mathbf{f}_{\text{HF}}(\boldsymbol{\theta}) - \boldsymbol{y}^{\text{obs}} \right) \right\|^2 \right),
\end{equation}
where $\| \cdot \|$ denotes the Euclidean norm. Every step of an \ac{MCMC} algorithm thus requires the solution of the high-fidelity problem: this is the main reason why solving Bayesian inverse problems involving differential models is indeed an intensive task. To overcome this bottleneck, one can introduce a surrogate model $\mathbf{f}_{\text{LF}}$ to approximate $\mathbf{f}_{\text{HF}}$ and define a corresponding approximate likelihood. This reduces the computational burden at the price of introducing an approximation in the posterior distribution. Unfortunately, the intrinsic ill-posedness of inverse problems might require extremely accurate surrogate models
\cite{manzoni2016accurate,pagani2017efficient,cui2014}. When a single surrogate is not sufficiently accurate or efficient, a structured approach based on multi-fidelity model management offers a viable alternative.

\subsection{Multi-Fidelity Model Management in Bayesian Inverse Problems}

In many inverse problems involving \acp{PDE}, a high-fidelity model $\mathbf{f}_{\text{HF}}$ is often available alongside one or more low-fidelity approximations. The objective of multi-fidelity model management is to design sampling strategies that combine these models efficiently, aiming to retain the accuracy of the high-fidelity model while reducing computational costs through surrogate evaluations \cite{Willcox2018}.

\subsubsection{Multi-Fidelity Filtering: Multi-Level Delayed Acceptance}
We consider a setting in which a high-fidelity model $\mathbf{f}_{\text{HF}}$ is available, along with a hierarchy of low fidelity numerical approximations, referred to as \textit{surrogate models}, denoted by $\mathbf{f}_\text{LF}^{(1)}, \mathbf{f}_\text{LF}^{(2)}, \ldots, \mathbf{f}_\text{LF}^{(L)}$, ordered by increasing accuracy and computational cost. Each surrogate model $\mathbf{f}_\text{LF}^{(l)}$ defines in principle an approximate likelihood:
\begin{equation}
\tilde{\pi}_\text{LF}^{(l)}(\boldsymbol{y} \mid \boldsymbol{\theta}) \propto \exp \left( -\frac{1}{2} \left\| \boldsymbol{\Sigma}_{\boldsymbol{\varepsilon}}^{-\frac{1}{2}} \left(\mathbf{f}_\text{LF}^{(l)}(\boldsymbol{\theta}) - \boldsymbol{y} \right) \right\|^2 \right), \quad l = 1, \ldots, L.
\end{equation}

These models are integrated into a multi-level \ac{MCMC} scheme of $L+1$ levels. Here, each level $l\le L$ uses the surrogate $\mathbf{f}_\text{LF}^{(l)}$ and its corresponding likelihood $\tilde{\pi}_\text{LF}^{(l)}$. At the level $L+1$, the high-fidelity model $\mathbf{f}_{\text{HF}}$ and the corresponding likelihood $\pi$ is employed to ensure accurate sampling. At each level, a candidate $\boldsymbol{\theta}'$ is proposed by generating a sub-chain of length $J_{l-1}$ (also referred to as sub-sampling rate) from the previous level $l-1$. For simplicity, we assume the subchain length to be fixed beforehand, as it often occurs in practice  \cite{lykkegaard_multilevel_2023}. The acceptance probability at level $l$ is then computed as:
\begin{equation}
\label{eq:acc}
\alpha_l(\boldsymbol{\theta}' \mid \boldsymbol{\theta}) = \min\left\{1, \frac{\tilde{\pi}_\text{LF}^{(l)}(\boldsymbol{y}^{\text{obs}}\mid\boldsymbol{\theta}') \tilde{\pi}_\text{LF}^{(l-1)}(\boldsymbol{y}^{\text{obs}}\mid\boldsymbol{\theta}) \pi(\boldsymbol{\theta}')}{\tilde{\pi}_\text{LF}^{(l)}(\boldsymbol{y}^{\text{obs}}\mid\boldsymbol{\theta}) \tilde{\pi}_\text{LF}^{(l-1)}(\boldsymbol{y}^{\text{obs}}\mid\boldsymbol{\theta}') \pi(\boldsymbol{\theta})} \right\}, \quad l=2,\ldots,L.
\end{equation}

At the coarsest level $l=1$, a standard \ac{MH} algorithm is used employing the cheapest model. As the level $l$ increases, finer and more expensive models are queried. At the finest level $l=L+1$ we have: 
\begin{equation}
\label{eq:acc1}
\alpha_{L+1}(\boldsymbol{\theta}' \mid \boldsymbol{\theta}) = \min\left\{1, \frac{{\pi}(\boldsymbol{y}^{\text{obs}}\mid\boldsymbol{\theta}') \tilde{\pi}_\text{LF}^{(L)}(\boldsymbol{y}^{\text{obs}}\mid\boldsymbol{\theta}) \pi(\boldsymbol{\theta}')}{{\pi}(\boldsymbol{y}^{\text{obs}}\mid\boldsymbol{\theta}) \tilde{\pi}_\text{LF}^{(L)}(\boldsymbol{y}^{\text{obs}}\mid\boldsymbol{\theta}') \pi(\boldsymbol{\theta})} \right\}.
\end{equation}

\begin{figure}[t]
    \centering
    \includegraphics[width=\linewidth]{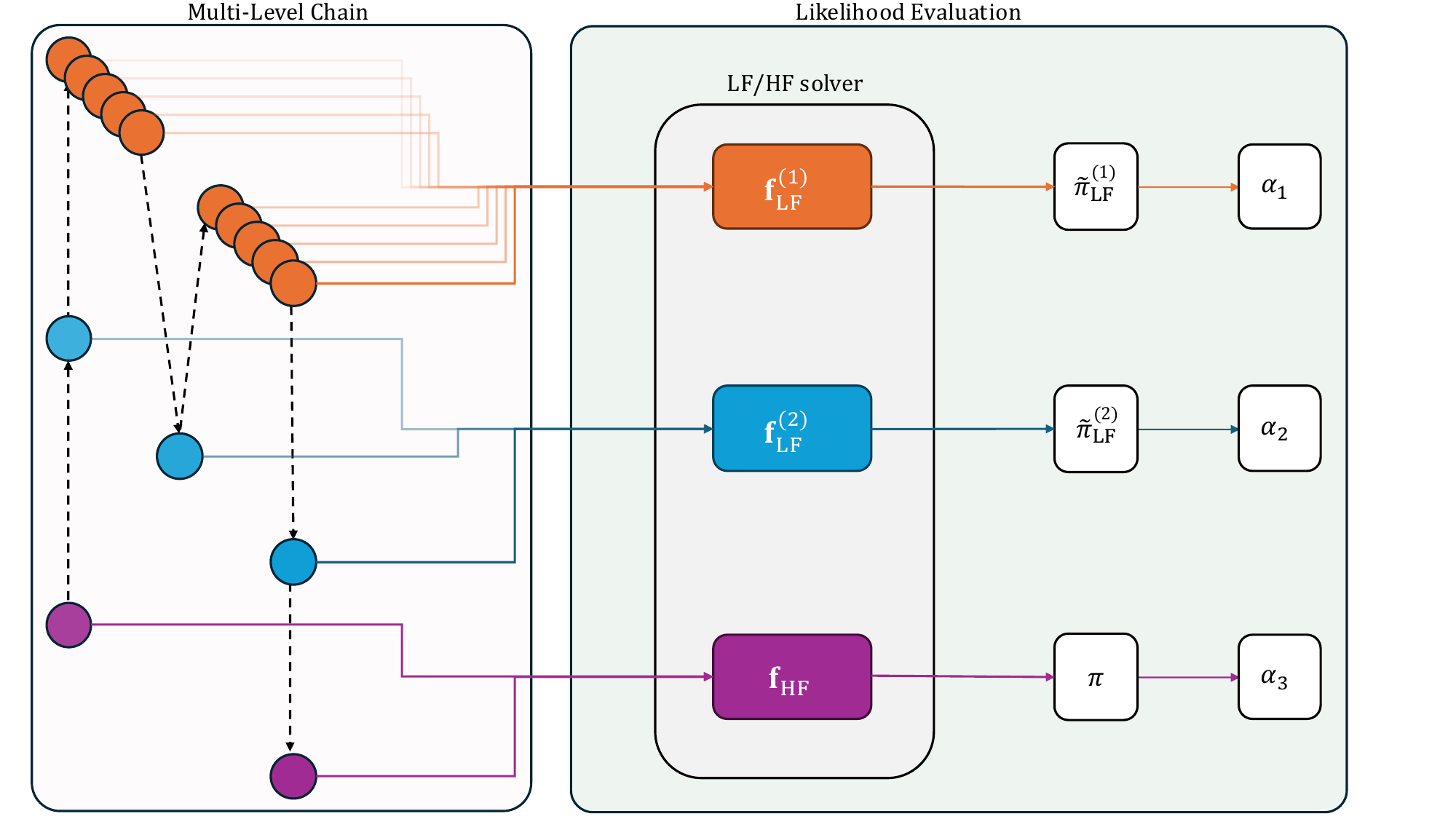}
    \caption{Schematic representation of the \ac{MLDA} scheme, for an instance of two low-fidelity solvers and one high-fidelity solver. At each level, a candidate $\boldsymbol{\theta}'$ is proposed by generating a sub-chain of length $J_{l-1}$. Each coarse level $l$ uses the surrogate $\mathbf{f}_\text{LF}^{(l)}$ and its corresponding likelihood $\tilde{\pi}_\text{LF}^{(l)}$, while the finest level uses  $\mathbf{f}_\text{HF}$ and its corresponding likelihood ${\pi}$.}
\label{fig:MLDA_schematic}
\end{figure}

This hierarchical design allows for the early rejection of poor candidates using cheaper models, significantly reducing the number of high-fidelity evaluations while maintaining sampling accuracy. Moreover, the nested sub-chains at different levels effectively reduce sample autocorrelation, improving the mixing of the chain.

Importantly, it can be shown that the chain at the finest level satisfies detailed balance with respect to the posterior distribution defined by the high-fidelity model \cite{lykkegaard_multilevel_2023, lee2025delayed}. Note also that the approach remains transparent to the particular choice of the forward model, allowing it to flexibly handle both linear and nonlinear, stationary or transient simulations without significant alterations  \cite{LykkegaardGWF2021,Fox2005,seelinger2025democratizing,madrigal2023analysis}. 
 A schematic representation of the method is provided in Fig.~\ref{fig:MLDA_schematic}, and the detailed algorithm is presented in Algorithm~\ref{alg:MLDA}.
 
\begin{algorithm}[t!]
\caption{Multi-level Delayed Acceptance (MLDA)}
\label{alg:MLDA}
    \hspace*{\algorithmicindent} \textbf{Input}:  
    High-fidelity likelihood $\pi(\boldsymbol{y}^{\text{obs}}\mid\cdot)$,  
    coarse likelihoods $\tilde{\pi}_\ell(\boldsymbol{y}^{\text{obs}}\mid\cdot)$ for $\ell=1,\dots,L$,  
    prior distribution $\pi(\cdot)$,  
    symmetric proposal distribution $q(\cdot \mid \cdot)$,  
    initial sample $\boldsymbol{\theta}_0$,  
    number of fine-level samples $J$,  
    sub-chain lengths $J_\ell$ for $\ell=1,\dots,L$.\\
    \hspace*{\algorithmicindent} \textbf{Output}: Chain of samples $\{\boldsymbol{\theta}_j\}_{j=1}^J$.
\begin{algorithmic}[1]
    \STATE Initialise $\boldsymbol{\theta}^\ell_0 = \boldsymbol{\theta}_0$ and $j_\ell \gets 0$ for all $\ell=1,\dots,L$.
    \FOR{$j = 0$ to $J-1$}
        \STATE \textbf{Level 1:} Run a sub-chain of length $J_1$ using Algorithm~\ref{alg:MH}, starting from current state $\boldsymbol{\theta}^{1}_0$ and proposal $q(\boldsymbol{\theta}' \mid \boldsymbol{\theta})$
        \STATE Set $\ell = 2$ 
        \WHILE{$\ell \le L$}
        \STATE Set proposed state $\tilde{\boldsymbol{\theta}}^\ell \leftarrow
        \boldsymbol{\theta}^{\ell-1}_{J_{\ell-1}}$
            \STATE Compute acceptance probability between current state $\boldsymbol{\theta}^\ell_{j_\ell}$ and proposed state $\tilde{\boldsymbol{\theta}}^\ell$:
            \[
            \alpha_\ell = \min\left\{1, 
            \frac{\tilde{\pi}_\ell(\boldsymbol{y}^{\text{obs}} \mid \tilde{\boldsymbol{\theta}}^\ell) \ \tilde{\pi}_{\ell-1}(\boldsymbol{y}^{\text{obs}} \mid \boldsymbol{\theta}^\ell_{j_\ell})}
                 {\tilde{\pi}_\ell(\boldsymbol{y}^{\text{obs}} \mid \boldsymbol{\theta}^\ell_{j_\ell}) \ \tilde{\pi}_{\ell-1}(\boldsymbol{y}^{\text{obs}} \mid \tilde{\boldsymbol{\theta}}^\ell)}
            \right\}
            \]
            \STATE With probability $\alpha_\ell$, accept: $\boldsymbol{\theta}^\ell_{{j_\ell}+1} \leftarrow \tilde{\boldsymbol{\theta}}^\ell$;  
            \quad otherwise reject: $\boldsymbol{\theta}^\ell_{{j_\ell}+1} \leftarrow \boldsymbol{\theta}^\ell_{j_\ell}$
            \STATE Increment $j_\ell \leftarrow j_\ell + 1$
            \IF{$j_\ell = J_\ell$}
            \STATE Set $\ell\leftarrow\ell+1$
            \ELSE
            \STATE Reset $j_k = 0$, $\boldsymbol{\theta}^k_0 \leftarrow \boldsymbol{\theta}^\ell_{j_\ell}$
                for all $1 \leq k < \ell$, and return to Step 3
            \ENDIF
        \ENDWHILE
        \STATE Set final proposal: $\tilde{\boldsymbol{\theta}} \leftarrow \boldsymbol{\theta}^{L}_{J_{L}}$
        \STATE Compute final acceptance probability between current $\boldsymbol{\theta}_j$ and proposed $\tilde{\boldsymbol{\theta}}$:
        \[
        \alpha_{L+1} = \min\left\{1, 
        \frac{\pi(\boldsymbol{y}^{\text{obs}} \mid \tilde{\boldsymbol{\theta}}) \ \tilde{\pi}_{L}(\boldsymbol{y}^{\text{obs}} \mid \boldsymbol{\theta}_j)}
             {\pi(\boldsymbol{y}^{\text{obs}} \mid \boldsymbol{\theta}_j) \ \tilde{\pi}_{L}(\boldsymbol{y}^{\text{obs}} \mid \tilde{\boldsymbol{\theta}})}
        \right\}
        \]
        \STATE With probability $\alpha_{L+1}$, accept: $\boldsymbol{\theta}_{j+1} \leftarrow \tilde{\boldsymbol{\theta}}$;  
        \quad otherwise reject: $\boldsymbol{\theta}_{j+1} \leftarrow \boldsymbol{\theta}_j$
        \STATE Reset $j_\ell = 0$, $\boldsymbol{\theta}^\ell_0 \leftarrow \boldsymbol{\theta}_{j+1}$ for all $1 \leq \ell \leq L$
    \ENDFOR
\end{algorithmic}
\end{algorithm}

\subsection{Multi-Fidelity Fusion through Neural Networks for Regression}\label{mfnn}

Although \ac{MLDA} can significantly reduce reliance on costly high-fidelity evaluations, its practical efficiency is still inhibited by discrepancies among different fidelity levels. Specifically, when low-fidelity approximations poorly represent their high-fidelity counterparts, the acceptance probability at the second stage deteriorates significantly. As a consequence, frequent rejections occur unless extremely low sub-sampling rates are adopted, which in turn result in suboptimal acceptance rates, inadequate mixing, and ultimately diminished effective sample sizes \cite{lykkegaard_multilevel_2023}.
To mitigate this problem, the current work introduces a multi-fidelity regression strategy leveraging artificial \acp{NN}, extending the recent framework of \cite{Guo_2022, conti2025progressive}. 

Here, the fundamental assumption is the existence of a nonlinear relationship $\mathcal{F}_{\mathrm{MF}}$ linking $n$ low-fidelity model outputs to their high-fidelity counterpart \cite{perdikaris2017nonlinear}, represented as follows:

\begin{equation}
\mathbf{f}_{\mathrm{HF}}(\boldsymbol{\theta})=\mathcal{F}_{\mathrm{MF}}\left(\boldsymbol{\theta}, \mathbf{f}_{\mathrm{LF}}^{(1)}(\boldsymbol{\theta}),\mathbf{f}_{\mathrm{LF}}^{(2)}(\boldsymbol{\theta}),\ldots,\mathbf{f}_{\mathrm{LF}}^{(n)}(\boldsymbol{\theta})\right).
\end{equation}

The key benefit of using \acp{NN} with respect to other multi-fidelity regression frameworks lies in the simultaneous and efficient incorporation of multiple low-fidelity approximations as inputs. Traditionally, even in hierarchical settings, multi-fidelity regression methods such as auto-regressive co-kriging \cite{perdikaris_multi-fidelity_2015} incorporate only a single lower-fidelity model at each stage. Although these schemes can be extended to multiple fidelity levels by sequentially stacking such mappings, the resulting structure restricts information flow to a one-directional correction chain. This  simplifies implementation and reduces memory requirements, but prevents the method from  exploiting complementary information across different solvers. In contrast, the inherent flexibility and nonlinear representational capacity of \acp{NN} allow efficient fusion of multiple low-fidelity sources within a unified model, enhancing predictive accuracy without increasing numerical complexity. An example of the effectiveness of \acp{NN} in combining multiple sources of information is provided in \cite{conti2025progressive}.

For this reason, to model this multi-fidelity mapping, a densely connected feedforward \ac{NN} architecture is adopted. Such a network, consisting of $N_L$ layers, defines a parametric function $\mathbf{f}_{\text{MF}}(\cdot; \mathbf{W}, \mathbf{b})$, where $\mathbf{W} = \{\mathbf{W}^{(1)}, \ldots, \mathbf{W}^{(N_L)}\}$ and $\mathbf{b} = \{\mathbf{b}^{(1)}, \ldots, \mathbf{b}^{(N_L)}\}$ denote weight matrices and biases, respectively. Each network layer performs an affine transformation followed by a nonlinear activation function, as follows:
\begin{equation}
\mathbf{z}^{(k)} = \sigma \left( \mathbf{W}^{(k)} \mathbf{z}^{(k-1)} + \mathbf{b}^{(k)} \right), \quad k = 1, \ldots, N_L,
\end{equation}
with $\mathbf{z}^{(0)}$ representing the input vector (i.e., $\mathbf{z}_0 = [\mathbf{\theta}, \mathbf{f}_\texttt{LF}^{(1)}(\mathbf{\theta}), ...]^\top$ is the concatenation of parameters $\boldsymbol{\theta}$ and outputs from multiple low-fidelity models), and $\sigma(\cdot)$ indicating a suitable nonlinear activation function. 

\begin{figure}[t]
    \centering
\includegraphics[width=0.85\linewidth]{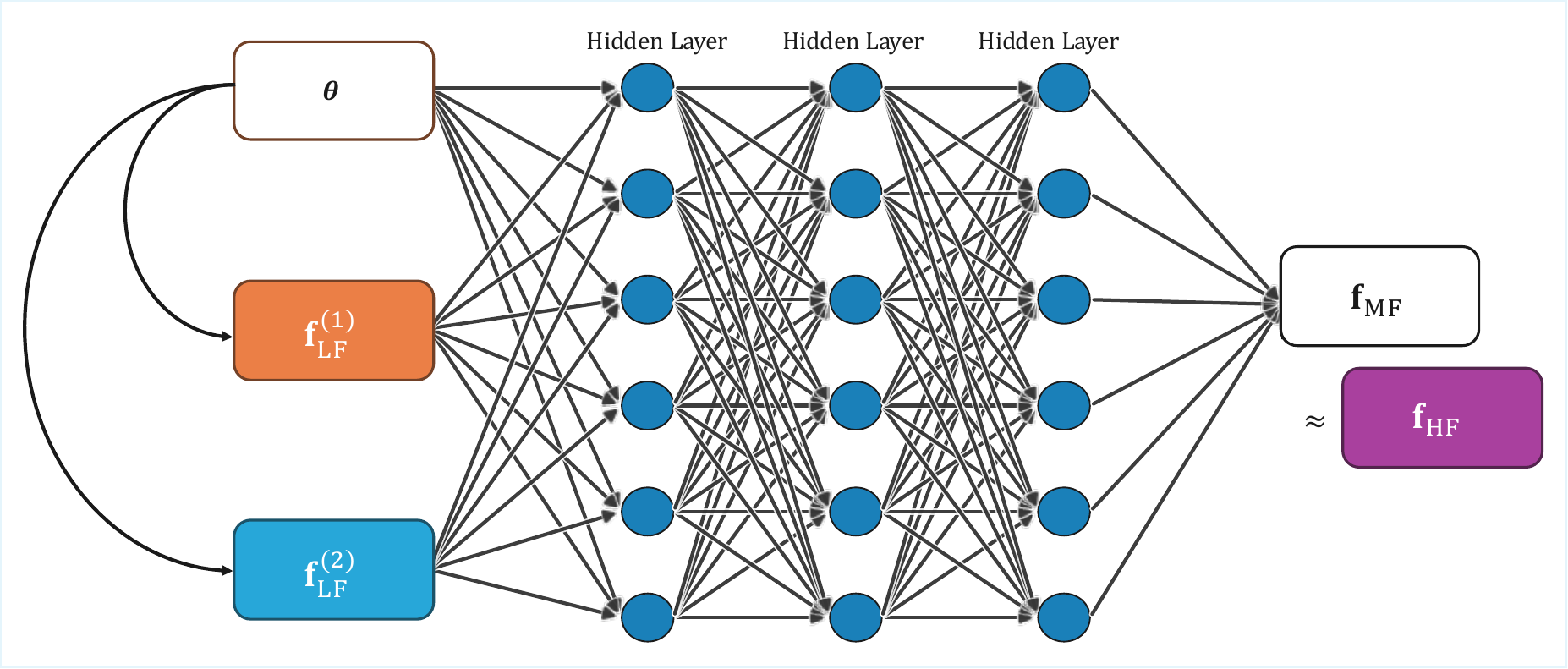}    \caption{Schematic representation of a multi-fidelity \ac{NN}, for a generic instance of two low-fidelity solvers and a reference high-fidelity solver. Input parameters $\boldsymbol{\theta}$ are passed to the low-fidelity solvers. The outputs of the solvers and the input parameter $\boldsymbol{\theta}$ are passed to the first hidden layer of the NN. The subsequent hidden layers perform nonlinear transformations to capture complex correlations across fidelities. The final network output serves as an approximation of the high-fidelity model.}
    \label{fig:mfnn}
\end{figure}

A schematic representation of a multi-fidelity \ac{NN}, for a generic instance of two low-fidelity solvers and a reference high-fidelity solver, is presented in Figure \ref{fig:mfnn}. Input parameters $\boldsymbol{\theta}$ are passed to the low-fidelity solvers. The outputs of the solvers and the input parameter $\boldsymbol{\theta}$ are passed to the first hidden layer of the NN. The subsequent hidden layers perform nonlinear transformations to capture complex correlations across fidelities. The final network output serves as an approximation of the high-fidelity model.

Training of the network parameters is performed offline. 
Specifically, a set of $N_\text{train}$ parameter samples $\{\boldsymbol{\theta}_j\}_{j=1}^{N_\text{train}}$ is generated, and for each sample, both the high-fidelity and all $n$ low-fidelity solvers are evaluated. 
The resulting dataset, consisting of $N_\text{train}$ paired high- and multi-fidelity evaluations, is used to train the \ac{NN} by minimizing the \ac{MSE} between the high-fidelity model outputs and the network predictions. 
The optimization problem can be expressed as:
\begin{equation}
\mathbf{W},\mathbf{b} = 
\arg \min_{\mathbf{W}, \mathbf{b}} 
\frac{1}{N_\text{train}} 
\sum_{j=1}^{N_\text{train}} 
\left\|
\mathbf{f}_{\text{HF}}(\boldsymbol{\theta}_j)
-
\mathbf{f}_{\text{MF}}\!\left(
\boldsymbol{\theta}_j,
\mathbf{f}_\text{LF}^{(1)}(\boldsymbol{\theta}_j),
\ldots,
\mathbf{f}_\text{LF}^{(n)}(\boldsymbol{\theta}_j);
\mathbf{W},\mathbf{b}
\right)
\right\|^2.
\label{eq:training_loss}
\end{equation}
This minimization is carried out using standard stochastic gradient–based optimizers, such as stochastic gradient descent and its variants (e.g., Adam~\cite{kingma2015adam}), as commonly employed in deep learning frameworks.

\begin{figure}[t]
    \centering
    \includegraphics[width=\linewidth]{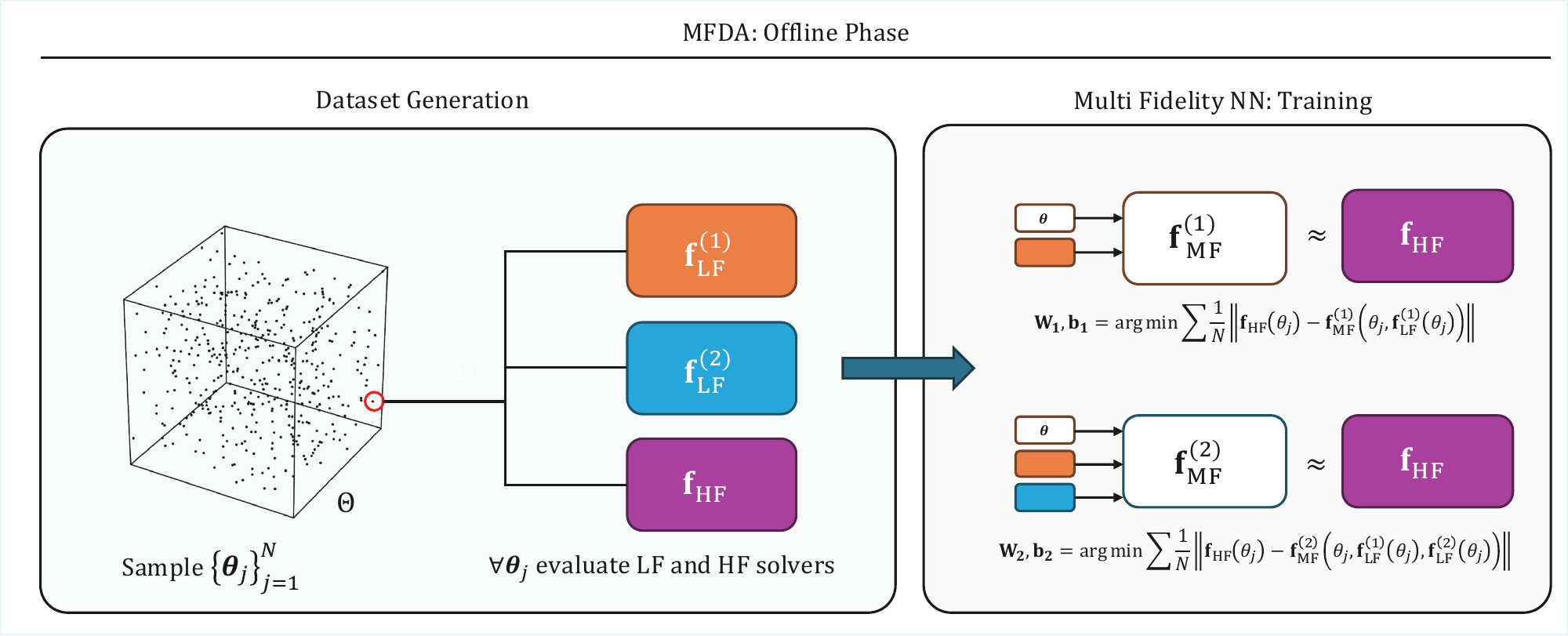}
    \includegraphics[width=\linewidth]{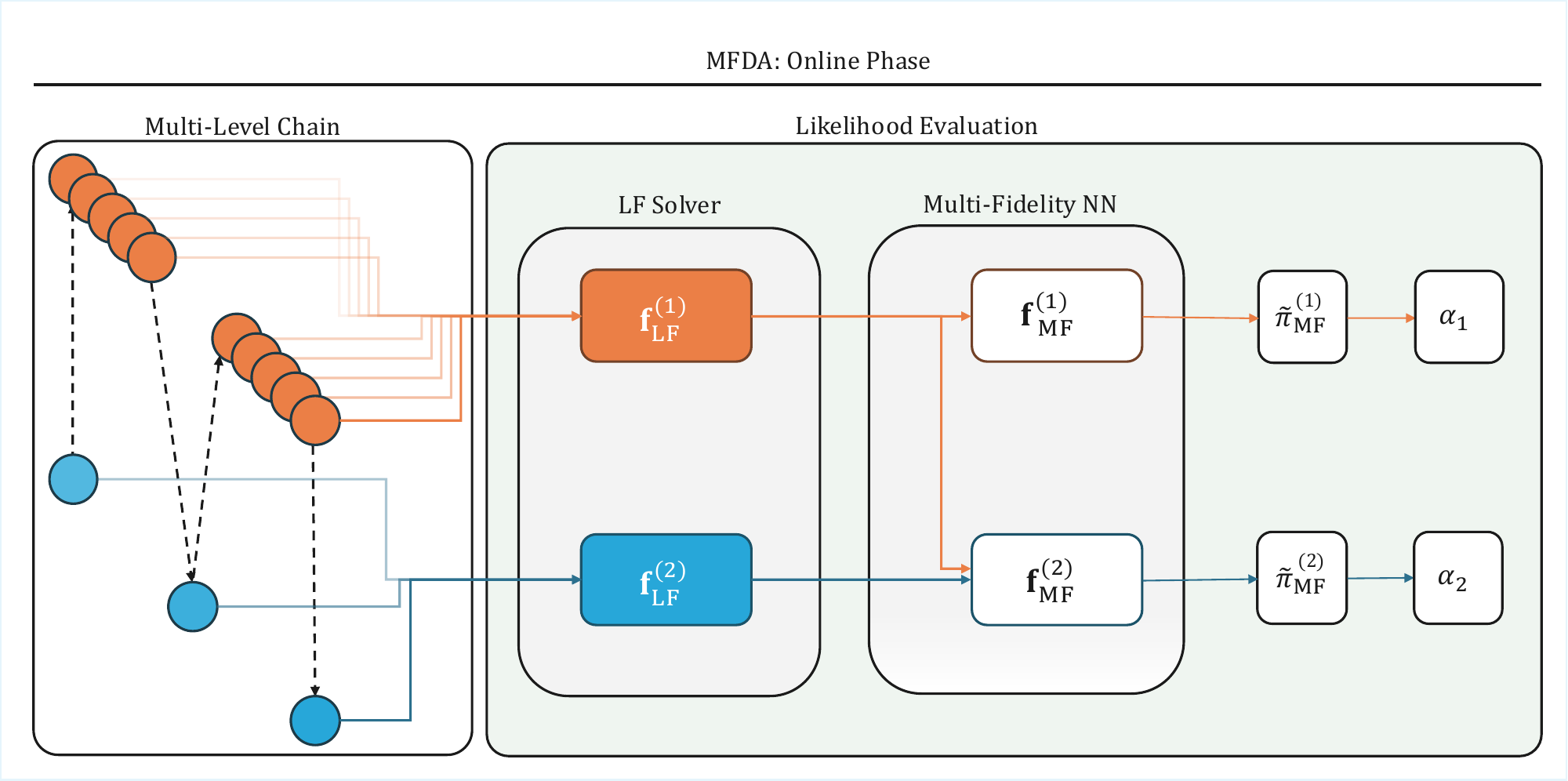}
    \caption{Schematic representation of the Multi-Fidelity Delayed Acceptance (\ac{MFDA}) framework for an instance of two low-fidelity solvers and one high-fidelity solver. First row: offline phase. A dataset of parameter samples is generated and each solver is evaluated for every parameter instance. Then multi-fidelity neural networks are trained to approximate the high-fidelity reference. Second row: online phase. A Markov chain of parameter samples is generated using a multi-level structure. At each level we evaluate the corresponding low-fidelity solvers and neural networks to compute the likelihood and acceptance rate.}
    \label{fig:MFDA_schematic}
\end{figure}

\begin{algorithm}[h!]
\caption{Multi-Fidelity Delayed Acceptance (MFDA)}
\label{alg:MFDA}
\hspace*{\algorithmicindent} \textbf{Input}:  
Number of training samples $N_{\mathrm{train}}$;    
high-fidelity model $\mathbf{f}_{\mathrm{HF}}$;  
low-fidelity surrogate models $\mathbf{f}_{\mathrm{LF}}^{(\ell)}$ for $\ell=1,\dots,L$;  
prior distribution $\pi(\cdot)$,  
symmetric proposal distribution $q(\cdot \mid \cdot)$;  
initial state $\boldsymbol{\theta}_0$;  
sub-chain lengths $J_\ell$ for $\ell=1,\dots,L-1$, and number of fine-level samples $J$.\\
\hspace*{\algorithmicindent} \textbf{Output}: Sample chain $\{\boldsymbol{\theta}_j\}_{j=1}^J$.
\begin{algorithmic}[1]
    \STATE \textbf{Offline Phase (Training):}
    \STATE Collect training data $\{(\boldsymbol{\theta}_i, \mathbf{f}_{\mathrm{HF}}(\boldsymbol{\theta}_i))\}_{i=1}^{N_{\mathrm{train}}}$.
    \FOR{$\ell = 1$ to $L$}
        \STATE Evaluate and store $\mathbf{f}_{\mathrm{LF}}^{(\ell)}(\boldsymbol{\theta}_i)$ for all $i$.
        \STATE Train multi-fidelity surrogate $\mathbf{f}_{\mathrm{MF}}^{(\ell)}$  
              using inputs $\big(\boldsymbol{\theta}_i, \mathbf{f}_{\mathrm{LF}}^{(1)}, \dots, \mathbf{f}_{\mathrm{LF}}^{(\ell)}\big)$.
    \ENDFOR
    \STATE
    
    \STATE \textbf{Online Phase (Inference):}
    \STATE Initialise $\boldsymbol{\theta}^\ell_0 = \boldsymbol{\theta}_0$ and $j_\ell \gets 0$ for all $\ell=1,\dots,L$.
    \STATE Initialise cache for low fidelity surrogate models evaluations.
    \FOR{$j = 0$ to $J-1$}
        \STATE \textbf{Level 1:} Run a sub-chain of length $J_1$ using Algorithm~\ref{alg:MH}, starting from $\boldsymbol{\theta}^1_{0}$.  
        \STATE Set $\ell \gets 2$
        \WHILE{$\ell < L$}
            \STATE Set proposal $\tilde{\boldsymbol{\theta}}^\ell \gets \tilde{\boldsymbol{\theta}}^{\ell-1}$.
            \STATE Evaluate $\mathbf{f}_{\mathrm{LF}}^{(\ell)}(\tilde{\boldsymbol{\theta}}^\ell)$ and store in cache.
            \STATE Retrieve surrogate outputs and compute  
                  $\mathbf{f}_{\mathrm{MF}}^{(\ell)}(\tilde{\boldsymbol{\theta}}^\ell)$ and likelihood $\tilde{\pi}_{\mathrm{MF}}^{(\ell)}(\boldsymbol{y} \mid \tilde{\boldsymbol{\theta}}^\ell)$ from Eq. \eqref{eq:MFlikelihood}.
            \STATE Compute acceptance probability:
            \[
            \alpha_\ell = \min\left\{1,\,
            \frac{\tilde{\pi}_{\mathrm{MF}}^{(\ell)}(\boldsymbol{y} \mid \tilde{\boldsymbol{\theta}}^\ell) \,
                  \tilde{\pi}_{\mathrm{MF}}^{(\ell-1)}(\boldsymbol{y} \mid \boldsymbol{\theta}^\ell_{j_\ell})}
                 {\tilde{\pi}_{\mathrm{MF}}^{(\ell)}(\boldsymbol{y} \mid \boldsymbol{\theta}^\ell_{j_\ell}) \,
                  \tilde{\pi}_{\mathrm{MF}}^{(\ell-1)}(\boldsymbol{y} \mid \tilde{\boldsymbol{\theta}}^\ell)} \right\}.
            \]
            \STATE With probability $\alpha_\ell$, accept: $\boldsymbol{\theta}^\ell_{j_\ell+1} \gets \tilde{\boldsymbol{\theta}}^\ell$;  
                   else set $\boldsymbol{\theta}^\ell_{j_\ell+1} \gets \boldsymbol{\theta}^\ell_{j_\ell}$.
            \STATE Increment $j_\ell \gets j_\ell + 1$.
            \IF{$j_\ell = J_\ell$}
            \STATE Set $\ell\leftarrow\ell+1$
            \ELSE
            \STATE Reset $j_k = 0$, $\boldsymbol{\theta}^k_0 \leftarrow \boldsymbol{\theta}^\ell_{j_\ell}$
                for all $1 \leq k < \ell$, and return to Step 12
            \ENDIF
        \ENDWHILE
        \STATE \textbf{Final Level $L$:} Let $\tilde{\boldsymbol{\theta}} \gets \tilde{\boldsymbol{\theta}}^{L-1}$.
        \STATE Compute $\mathbf{f}_{\mathrm{MF}}^{(L)}(\tilde{\boldsymbol{\theta}})$ and likelihood $\tilde{\pi}_{\mathrm{MF}}^{(L)}(\boldsymbol{y} \mid \tilde{\boldsymbol{\theta}})$.
        \STATE Compute final acceptance probability:
        \[
        \alpha_L = \min\left\{1,\,
        \frac{\tilde{\pi}_{\mathrm{MF}}^{(L)}(\boldsymbol{y} \mid \tilde{\boldsymbol{\theta}}) \,
              \tilde{\pi}_{\mathrm{MF}}^{(L-1)}(\boldsymbol{y} \mid \boldsymbol{\theta}_j)}
             {\tilde{\pi}_{\mathrm{MF}}^{(L)}(\boldsymbol{y} \mid \boldsymbol{\theta}_j) \,
              \tilde{\pi}_{\mathrm{MF}}^{(L-1)}(\boldsymbol{y} \mid \tilde{\boldsymbol{\theta}})} \right\}.
        \]
        \STATE With probability $\alpha_L$, set $\boldsymbol{\theta}_{j+1} \gets \tilde{\boldsymbol{\theta}}$;  
               else set $\boldsymbol{\theta}_{j+1} \gets \boldsymbol{\theta}_j$.
        \STATE Reset $\boldsymbol{\theta}^\ell_0 \gets \boldsymbol{\theta}_{j+1}$ and $j_\ell \gets 0$ for all $\ell = 1,\dots,L$.
    \ENDFOR
\end{algorithmic}
\end{algorithm}

\subsection{A remark on low-fidelity solvers}

In standard multi-level frameworks concerning physical problems and PDEs, low-fidelity solvers are typically constructed by systematically reducing the resolution or complexity of a high-fidelity model. Common strategies include mesh coarsening in finite element or finite volume schemes, simplified discretizations, or algebraic multi-level formulations such as multigrid approaches \cite{cliffe2011multilevel,giles2015multilevel,beskos2017multilevel,lykkegaard_multilevel_2023}. 

However, in many practical scenarios, lower-fidelity models may differ from the high-fidelity solver not only in resolution but also in their underlying physical assumptions, boundary conditions, or numerical formulations \cite{Behrou,CLEEMAN2023118125,Ebers,KIM2023109163}. In such cases, correlations across fidelity levels can become strongly nonlinear \cite{perdikaris2017nonlinear}. As a result, the filtering at the lower levels of \ac{MLDA} scheme fails to reflect the structure of the fine posterior distribution, thereby limiting the effectiveness of \ac{MLDA} approach.

To address this challenge, non-linear information fusion techniques \cite{perdikaris2017nonlinear,conti2025progressive}, such as the \ac{NN}–based strategy adopted in this work, provide a flexible solution. By integrating multiple low-fidelity models within a unified nonlinear regression framework, the proposed approach can effectively capture complex dependencies between low- and high-fidelity responses. As a result, the \ac{NN}-based fusion yields surrogate representations that more closely align the coarse-level approximations with the high-fidelity posterior.

\subsection{Multi-Fidelity Delayed Acceptance}
\label{sec:mfda}

The multi-fidelity \acp{NN} are embedded into the multi-level \ac{MCMC} sampling framework of $L$ levels to form the proposed \ac{MFDA} scheme. For each fidelity level $l=1,\ldots,L$, we construct a multi-fidelity \ac{NN}
\begin{equation}
\label{eq:MF_l}
\mathbf{f}_{\mathrm{MF}}^{(l)}(\boldsymbol{\theta})
=
\mathbf{f}_{\mathrm{MF}}\!\left(
\boldsymbol{\theta},
\mathbf{f}_{\mathrm{LF}}^{(1)}(\boldsymbol{\theta}),
\ldots,
\mathbf{f}_{\mathrm{LF}}^{(l)}(\boldsymbol{\theta})
\right),
\end{equation}
which takes as inputs the parameter vector and the outputs of all solvers up to level $l$. Each $\mathbf{f}_{\mathrm{MF}}^{(l)}$ follows the architecture introduced in Section~\ref{mfnn}. The algorithm operates in two stages:
\begin{itemize}
    \item {Offline phase (training)}: we generate a set of parameter samples $\{\boldsymbol{\theta}_j\}_{j=1}^{N_{\mathrm{train}}}$ (e.g., via Latin hypercube sampling) and evaluate all available solvers at these points. We train each network $\mathbf{f}_{\mathrm{MF}}^{(l)}$ independently by minimizing the mean squared error between the high-fidelity outputs and the network predictions, as in~\eqref{eq:training_loss}. This is the only part where we rely on the high-fidelity model. The number of training points should be selected so that the surrogate at the finest level $l = L$ attains the desired accuracy, as this model replaces the high-fidelity solver at the final stage of the multi-level chain. 
    \item {Online phase (inference)}:  In this phase, parameter samples are drawn using the standard \ac{MLDA} procedure, with likelihood evaluations replaced by their multi-fidelity \ac{NN} counterparts:
    
    \begin{equation}\label{eq:MFlikelihood}
    \tilde{\pi}_{\text{MF}}^{(l)}(\boldsymbol{y}^{\text{obs}} \mid \boldsymbol{\theta}) \propto \exp \left( -\frac{1}{2} \left\| \boldsymbol{\Sigma}_{\boldsymbol{\varepsilon}}^{-\frac{1}{2}} \left(\mathbf{f}_{\text{MF}}^{(l)}(\boldsymbol{\theta}) - \boldsymbol{y}^{\text{obs}} \right) \right\|^2 \right), \quad l = 1, \ldots, L.
    \end{equation}

    These adjusted multi-fidelity likelihoods naturally replace standard likelihoods in the acceptance criterion of the \ac{MLDA} framework. Specifically, the modified acceptance probability at fidelity level $l$ becomes:

    \begin{equation}
    \label{eq:acc2}
    \alpha_l(\boldsymbol{\theta}’ \mid \boldsymbol{\theta}) = \min\left\{1, \frac{\tilde{\pi}_{\text{MF}}^{(l)}(\boldsymbol{y}^{\text{obs}} \mid \boldsymbol{\theta}’) \tilde{\pi}_{\text{MF}}^{(l-1)}(\boldsymbol{y}^{\text{obs}} \mid \boldsymbol{\theta}) \pi(\boldsymbol{\theta}’)}{\tilde{\pi}_{\text{MF}}^{(l)}(\boldsymbol{y}^{\text{obs}} \mid \boldsymbol{\theta}) \tilde{\pi}_{\text{MF}}^{(l-1)}(\boldsymbol{y}^{\text{obs}} \mid \boldsymbol{\theta}’) \pi(\boldsymbol{\theta})} \right\}, \quad l=2,\ldots,L.
    \end{equation}
    At level $l=1$ we use \ac{MH} scheme using $\tilde{\pi}_{\text{MF}}^{(1)}$. In the proposed MFDA implementation, only the $L$ coarse solvers are used online: likelihood evaluations at all levels rely on the multi-fidelity \acp{NN}, and the high-fidelity model is not called during sampling.
    Preserving the filtering structure enables reuse of coarse-level evaluations as inputs to finer-level networks, avoiding redundant computations. Hence, the only additional online cost relative to standard \ac{MLDA} is the (typically negligible) \ac{NN} inference, while the high-fidelity solver is not called online. 
    \end{itemize}

The Markov chain at the finest level satisfies detailed balance
with respect to the posterior associated with the surrogate likelihood
$\tilde{\pi}_{\text{MF}}^{(L)}$, rather than with the exact high-fidelity
posterior. We expect that the use of multi-fidelity inputs allows one to construct
sufficiently accurate surrogate posteriors with substantially fewer
high-fidelity training evaluations.
The choice between \ac{MFDA} and standard \ac{MLDA} depends on whether the offline training cost is compensated by the improvement in effective sample size and mixing during inference.

Further improvements could be achieved through the incorporation of adaptive training mechanisms for the \ac{NN}s during the online phase. While this adaptation has the potential to introduce bias, it could substantially reduce the dependence on the offline training phase. However, implementing online learning for \ac{NN}s remains a non-trivial challenge and may require significant methodological adjustments, potentially involving techniques such as deep kernel learning \cite{botteghi2022deep}. Figure~\ref{fig:MFDA_schematic} illustrates the workflow, and Algorithm~\ref{alg:MFDA} provides the full procedure.

\section{Numerical Experiments I: Isotropic Groundwater Flow - Transmissivity Reconstruction}
\label{sec:res1} 
The first test concerns the reconstruction of a spatially varying subsurface transmissivity field from hydraulic pressure measurements, using the benchmark configuration of~\cite{LykkegaardGWF2021}. This problem is governed by a linear, stationary elliptic equation and serves as a standard reference in inverse UQ. 


The evaluation focuses on two aspects:
\begin{enumerate}
    \item \textbf{Forward accuracy:} The ability of multi-fidelity \ac{NN}s to enhance the accuracy of coarse models, measured via the \ac{RMSE} with respect to the high-fidelity solver solution. 
    \item \textbf{Sampling efficiency:} The effectiveness of the \ac{MFDA} scheme for Bayesian inversion, compared with \ac{MH} and \ac{MLDA} algorithms.
    In particular, sampling efficiency is quantified using the time-to-\ac{ESS} ratio. For a Markov chain of $N$ samples, the \ac{ESS} is usually defined as
    \begin{equation}
    N_{\mathrm{eff}} = \frac{N}{1 + 2 \sum_{t = 1}^{\infty} \rho_t},
    \end{equation}
    where $\rho_t$ denotes the autocorrelation at lag $t$. This reflects the number of effectively independent samples produced by the chain. The objective is to obtain the highest possible \ac{ESS} for a given computational budget.
\end{enumerate}

\subsection{Problem description}
Let $\Omega = (0,1)^2$ denote the spatial domain with boundary $\Gamma = \partial \Omega$. The steady-state hydraulic head $h(\mathbf{x})$ satisfies the following diffusion equation
\begin{equation}
-\nabla \cdot \left(T(\mathbf{x}) \nabla h(\mathbf{x})\right) = g(\mathbf{x}), \quad \mathbf{x} \in \Omega,
\end{equation}
where $T(\mathbf{x})$ denotes the transmissivity field and $g(\mathbf{x})$ is the source term. The boundary conditions are
\begin{equation}
h(\mathbf{x}) = h_D(\mathbf{x}) \quad \text{on } \Gamma_D, \qquad \left(-T(\mathbf{x})\nabla h(\mathbf{x})\right) \cdot \mathbf{n} = q_N(\mathbf{x}) \quad \text{on } \Gamma_N,
\end{equation}
where $\Gamma = \Gamma_D \cup \Gamma_N$, $\Gamma_D \cap \Gamma_N = \emptyset$. In particular, $q_N({\bf x}) = 0$ on 
\[
\Gamma_N = \Gamma_N^{bottom} \cup \Gamma_N^{top} = \{(x_1,0) \, : \, x_1 \in (0,1) \} \cup 
\{(x_1,1) \, : \, x_1 \in (0,1) \}
\]
whereas on 
\[
\Gamma_D = \Gamma_D^{left} \cup \Gamma_D^{right} =  \{(0,x_2) \, : \, x_2 \in (0,1) \} \cup 
\{(1,x_2) \, : \, x_2 \in (0,1) \}
\]
we set $h_D(\mathbf{x}) = 1$ on  $\Gamma_D^{left}$ and  $h_D(\mathbf{x}) = 0$ on  $\Gamma_D^{right}$. 

A widely used model for the prior distribution of aquifer transmissivity in groundwater hydrology is the log-Gaussian random field~\cite{LykkegaardGWF2021}. In this approach, the logarithm of the transmissivity, $\log T(\mathbf{x})$, is modeled as a Gaussian random field \cite{adler2007random} characterized by a specified mean and covariance structure. The mean of $\log T(\mathbf{x})$ is set to $\mu=1$, and the covariance function is given by
\begin{equation}
C(\mathbf{x}_1, \mathbf{x}_2) = \sigma^2 \exp\left(-\frac{\|\mathbf{x}_1 - \mathbf{x}_2\|_2^2}{2\lambda^2}\right), \quad \mathbf{x}_1, \mathbf{x}_2 \in \Omega,
\end{equation}
where $\sigma = 0.1$ and $\lambda$ denotes the correlation length. 
To enable a finite-dimensional parameterization, the log-Gaussian random field is approximated using a truncated \ac{KL} \cite{Loeve1978} expansion. Specifically,
\begin{equation}
\log T(\mathbf{x}) = \mu + \sum_{i=1}^m \sqrt{\lambda_i} \psi_i(\mathbf{x})\, \theta_i,
\label{eq:transmissivity}
\end{equation}
where $\{\lambda_i\}_{i=1}^m$ and $\{\psi_i(\mathbf{x})\}_{i=1}^m$ are the $m$ largest eigenvalues and associated $L^2$-orthonormal eigenfunctions of the covariance operator with kernel $C(\mathbf{x}_1, \mathbf{x}_2)$, and $\theta_i \sim \mathcal{N}(0, 1)$ are independent standard normal random variables. 
In this study, we deal with $m=64$ modes to represent the spatially distributed random field, ensuring that approximately 95\% of the field variance is retained.
The parameter vector $\boldsymbol{\theta} = (\theta_1, \ldots, \theta_m)^\top \sim \mathcal{N}(0, I_{m})$ serves as the set of uncertain parameters in the stochastic PDE model.

Therefore, our objective is to infer the posterior distribution of $\boldsymbol{\theta}$, given noisy measurements of the hydraulic head $h$ at $d=25$ discrete sensor locations $\{\mathbf{x}_j\}_{j=1}^d \subset \Omega$. These measurements are collected in the observation vector $\mathbf{y}^{\mathrm{obs}} \in \mathbb{R}^d$. 

\begin{figure}[t]
    \centering
    \includegraphics[width=0.85\linewidth]{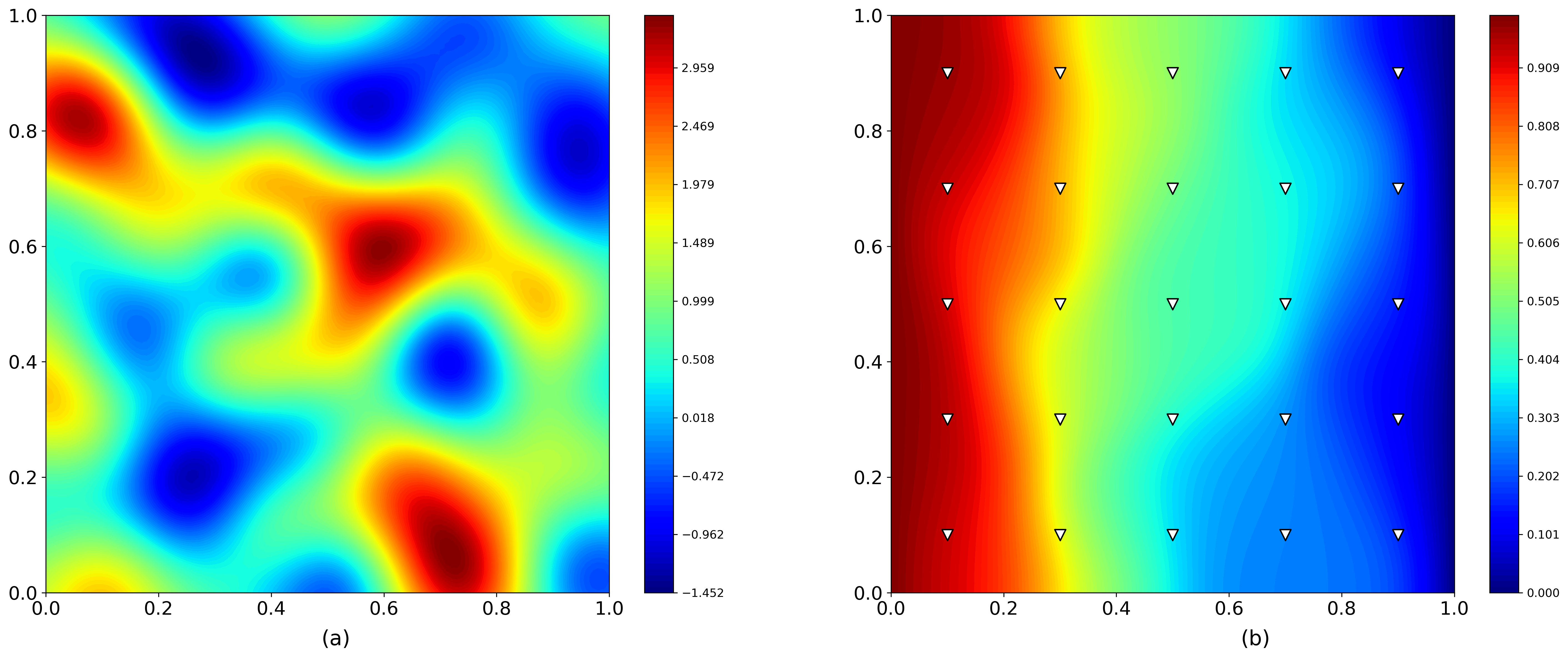}
    \caption{(a) Transmissivity field corresponding to mesh $\mathcal{T}_\mathrm{HF}$ and (b) hydraulic head using high-fidelity solver. Inverted triangles indicate sensor locations.}
    \label{fig:gwflowex}
\end{figure}

\begin{figure}[t]
    \centering
    \begin{subfigure}[b]{0.4\linewidth}
        \includegraphics[width=\linewidth]{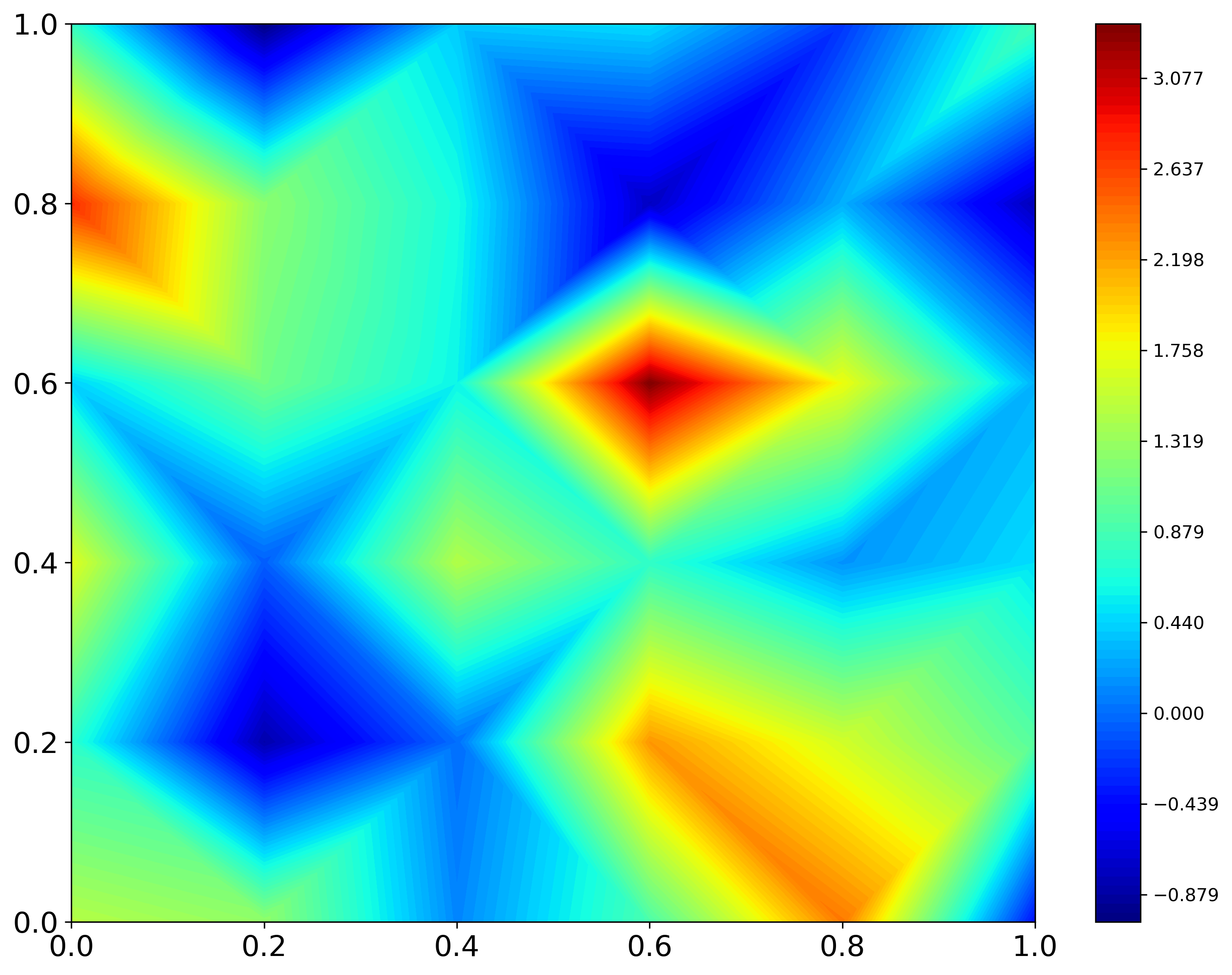}
        \caption{}
    \end{subfigure}%
    \begin{subfigure}[b]{0.4\linewidth}
        \includegraphics[width=\linewidth]{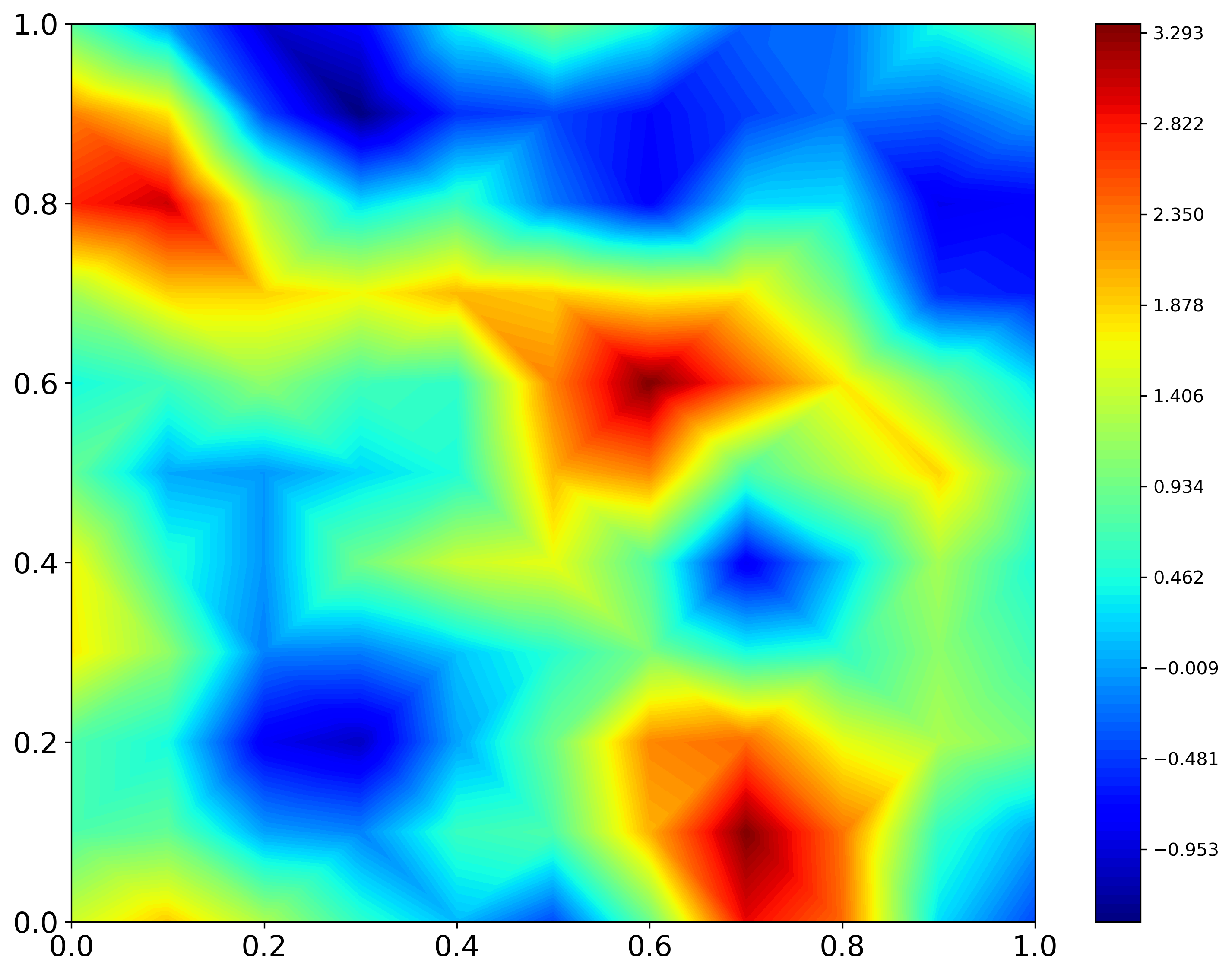}
        \caption{}
    \end{subfigure}
    \begin{subfigure}[b]{0.4\linewidth}
        \includegraphics[width=\linewidth]{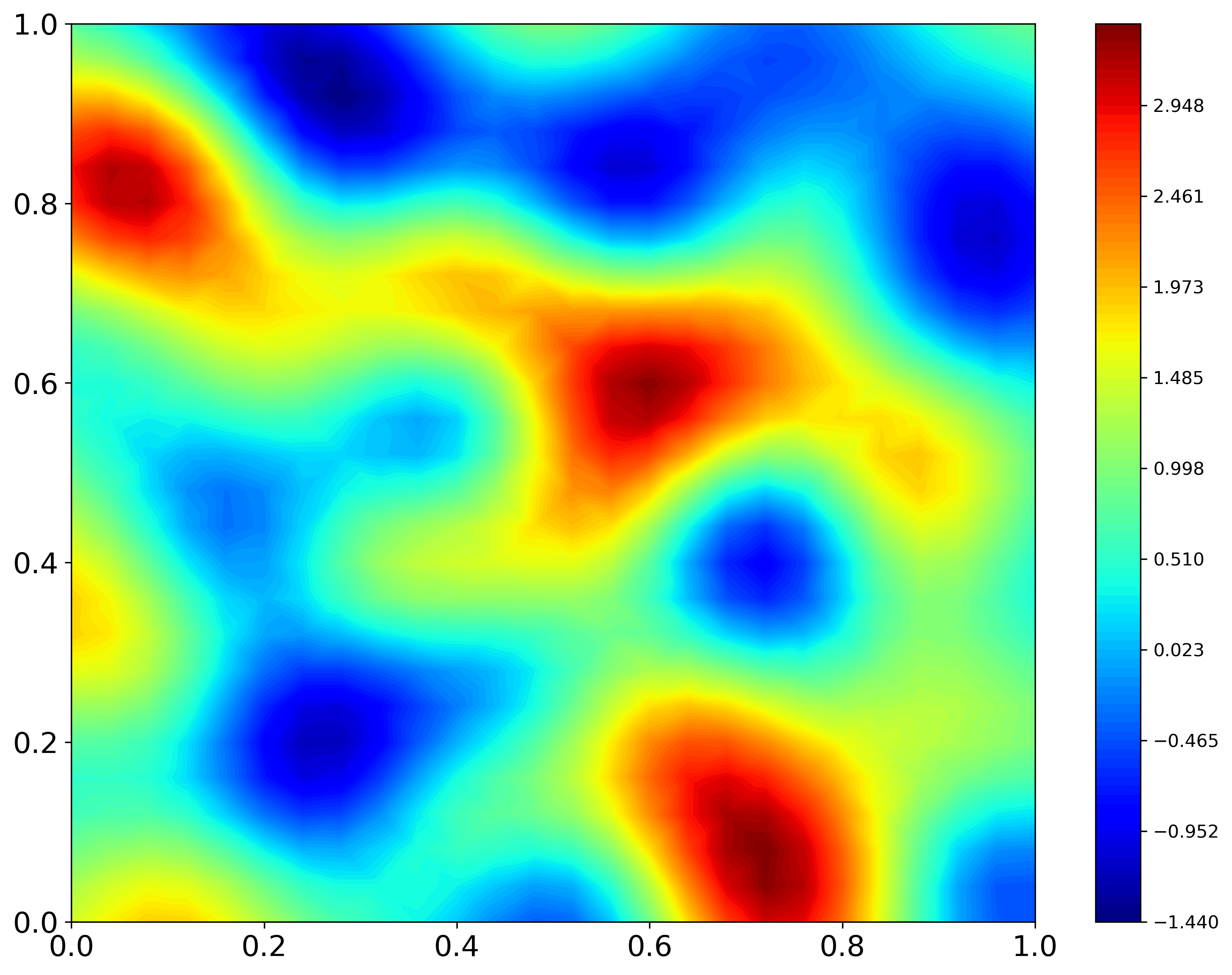}
        \caption{}
    \end{subfigure}%
    \begin{subfigure}[b]{0.4\linewidth}
        \includegraphics[width=\linewidth]{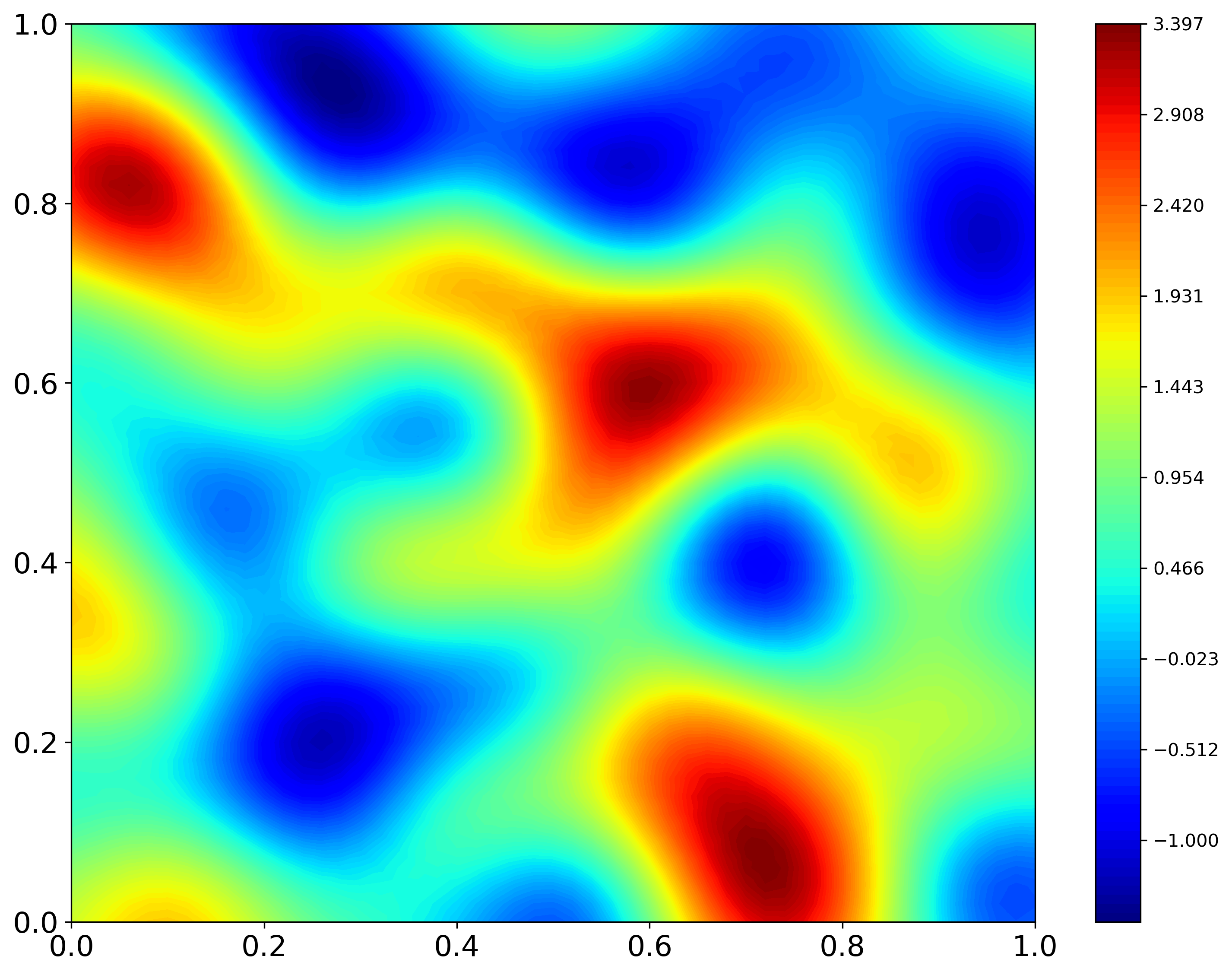}
        \caption{}
    \end{subfigure}
    \caption{Transmissivity fields corresponding to low-fidelity meshes: (a) $\mathcal{T}_1$, (b) $\mathcal{T}_2$, (c) $\mathcal{T}_3$, (d) $\mathcal{T}_4$.}
    \label{fig:transmissivity}
\end{figure}


\begin{table}[b!]
\renewcommand{\arraystretch}{1.5}
\centering
\begin{tabular}{lrrrr}
\toprule
\textbf{Model} & \textbf{Mesh Size [\#elements]} & \textbf{DoFs} & \textbf{Time / eval [s]} & \textbf{RMSE} \\
\midrule
$f_{\mathrm{HF}}$ 
& $100 \times 100$ 
& $10{,}201$ 
& $1.27 \times 10^{-1}$
& $0$ \\

$f_{\mathrm{LF}}^{(1)}$ 
& $5 \times 5$ 
& $36$ 
& $2.53 \times 10^{-3}$
& $5.1\times 10^{-2}$ \\

$f_{\mathrm{LF}}^{(2)}$ 
& $10 \times 10$ 
& $121$ 
& $3.22 \times 10^{-3}$ 
& $1.8\times 10^{-2}$ \\

$f_{\mathrm{LF}}^{(3)}$ 
& $25 \times 25$ 
& $676$ 
& $4.11 \times 10^{-3}$
& $4.4\times 10^{-3}$ \\

$f_{\mathrm{LF}}^{(4)}$ 
& $50 \times 50$ 
& $2{,}601$ 
& $1.09 \times 10^{-2}$ 
& $6.7\times 10^{-4}$ \\
\midrule
$f_{\mathrm{MF}}^{(1)}$ 
& --- & --- 
& $4.0 \times 10^{-4}$ 
& $7.0\times 10^{-3}$ \\

$f_{\mathrm{MF}}^{(2)}$ 
& --- & --- 
& $5.6 \times 10^{-4}$ 
& $4.9\times 10^{-3}$ \\

$f_{\mathrm{MF}}^{(3)}$ 
& --- & --- 
& $6.6 \times 10^{-4}$ 
& $6.7\times 10^{-4}$ \\

$f_{\mathrm{MF}}^{(4)}$ 
& --- & --- 
& $7.8 \times 10^{-4}$ 
& $1.9\times 10^{-4}$ \\
\bottomrule
\end{tabular}
\caption{Spatial discretization, degrees of freedom (DoFs), computational time per forward evaluation, 
and predictive accuracy (RMSE) of the high-fidelity, low-fidelity, and multi-fidelity surrogate models.}
\label{tab:model_performance1}
\end{table}

\subsection{MFDA: Setting}
For this test case we design a MFDA scheme with $L=4$ levels, defining the following models:

\begin{itemize}
    \item \textbf{High-fidelity solver}: The high-fidelity model employs a finite element discretization on a structured triangular mesh $\mathcal{T}_{\mathrm{HF}}$ consisting of $100$ elements per spatial direction, corresponding to $101$ nodes along each axis. Linear finite elements are used. For each parameter instance $\boldsymbol{\theta}$, it reconstructs the transmissivity field (see Eq.~\eqref{eq:transmissivity}) and computes the pressure head $h$ at the $d$ sensor locations $\{\mathbf{x}_j\}_{j=1}^d$. We have $\mathbf{f}_{\mathrm{HF}}: \mathbb{R}^m \rightarrow \mathbb{R}^d$. The PDE solution requires solving a linear system with $101^2$ degrees of freedom. For the solution, we use the GMRES solver with an incomplete LU preconditioner. Figure~\ref{fig:gwflowex} illustrates an example of the numerical solution and the corresponding transmissivity field. The high-fidelity solver generates reference data for the offline training. In addition, when solving the inverse problem, synthetic observations are produced by sampling random parameter instances $\boldsymbol{\theta}$, solving the high-fidelity model, and perturbing the resulting pressure fields with additive Gaussian noise, as detailed in Section~\ref{sec:online1}.
    \item \textbf{Low-fidelity solvers}: 
    Four low-fidelity models are obtained by uniform mesh coarsening, yielding meshes $\mathcal{T}_l$, $l=1,\dots,4$. The corresponding spatial resolutions and degrees of freedom are reported in Table~\ref{tab:model_performance1}. As shown in Fig.~\ref{fig:transmissivity}, the representation of the transmissivity field becomes progressively smoother as the mesh is refined.  All low-fidelity models employ linear finite elements and GMRES for the solution of the resulting linear systems. Each solver $\mathbf{f}_{\mathrm{LF}}^{(l)} : \mathbb{R}^m \rightarrow \mathbb{R}^d$ maps the parameter vector $\boldsymbol{\theta}$ to the pressure head evaluated at the $25$ sensor locations. The outputs of $\mathbf{f}_{\mathrm{LF}}^{(l)}$ are used as inputs to the multi-fidelity \ac{NN} at level $l$, and, when the end of the sub-chain is reached, are stored for use at subsequent finer levels.      
    \item \textbf{Multi-fidelity \ac{NN}:} At each level $l$, a multi-fidelity \ac{NN} $\mathbf{f}_{\mathrm{MF}}^{(l)}: \mathbb{R}^m \times [\mathbb{R}^d]^l \rightarrow \mathbb{R}^d$ is trained to approximate the high-fidelity pressure head solution, given the parameter vector $\boldsymbol{\theta}$ and the outputs of all low-fidelity solvers up to level $l$, according to \eqref{eq:MF_l}. Each multi-fidelity surrogate is then used during inference to compute an approximate likelihood $\pi_{\mathrm{MF}}^{(l)}$. Additional details for the training are provided in the next section.
\end{itemize}
All finite element simulations are performed using the FEniCS library\cite{alnaes_fenics_2015}, based on code adapted from Lykkegaard et al.~\cite{LykkegaardGWF2021} (available at \url{https://ore.exeter.ac.uk/repository/handle/10871/125704}). In the following,  the neural network surrogates are implemented in \texttt{TensorFlow}~\cite{tensorflow2015-whitepaper}, and the multi level \ac{MCMC} are implemented using the \texttt{TinyDA} library \cite{tinyda}.

\subsection{MFDA: offline training and models accuracy}

During the offline phase, $N_{\mathrm{train}}$ parameter samples $\{\boldsymbol{\theta}_j\}_{j=1}^{N_{\mathrm{train}}}$ are drawn from the prior, and all finite element solvers are evaluated at these instances. Each multi-fidelity network $\mathbf{f}_\mathrm{MF}^{(l)}$, $l=1,\ldots,4$, is trained to minimize the mean squared error with respect to the high-fidelity output:

\begin{equation}
\mathcal{L}(\mathbf{W}_l, \mathbf{b}_l) = \frac{1}{N_{\mathrm{train}}} \sum_{j=1}^{N_{\mathrm{train}}} \left\| \mathbf{f}_{\mathrm{HF}}(\boldsymbol{\theta}_j) - \mathbf{f}_\mathrm{MF}^{(l)}\left(\boldsymbol{\theta}_j, \mathbf{f}_\mathrm{LF}^{(1)}(\boldsymbol{\theta}_j), \ldots, \mathbf{f}_\mathrm{LF}^{(l)}(\boldsymbol{\theta}_j); \mathbf{W}_l, \mathbf{b}_l\right) \right\|^2, \quad l=1,\ldots,4.
\end{equation}
where $\mathbf{W}_l$ and $\mathbf{b}_l$ are the weights and biases of the multi-fidelity NN at level $l$, respectively. 
The Adam optimizer is used for minimization. Network architecture details are provided in \ref{app:nn_architectures}.

\begin{figure}[t]
    \centering
    \begin{subfigure}[b]{0.48\linewidth}
        \includegraphics[width=\linewidth]{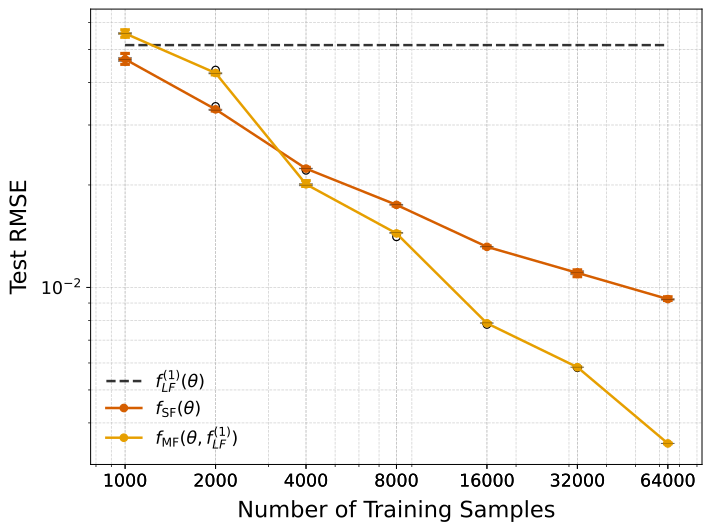}
        \caption{First-level surrogate, $\mathbf{f}_\mathrm{MF}^{(1)}$.}
    \end{subfigure}\hfill
    \begin{subfigure}[b]{0.48\linewidth}
        \includegraphics[width=\linewidth]{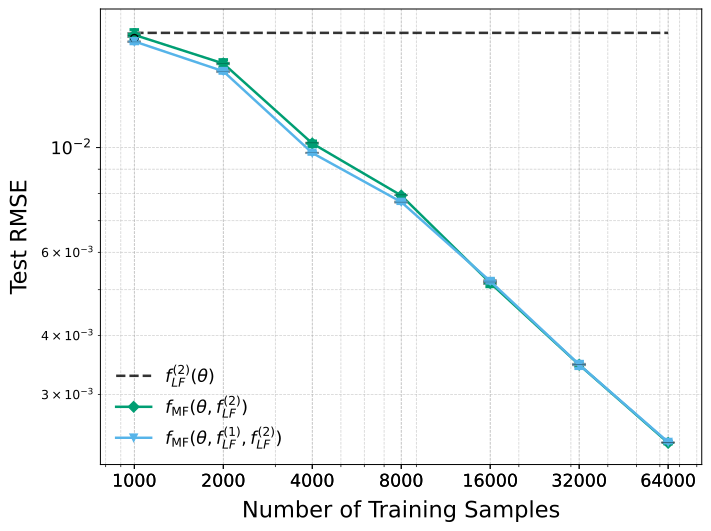}
        \caption{Second-level surrogate, $\mathbf{f}_\mathrm{MF}^{(2)}$.}
    \end{subfigure}

    \vspace{0.6em}

    \begin{subfigure}[b]{0.48\linewidth}
        \includegraphics[width=\linewidth]{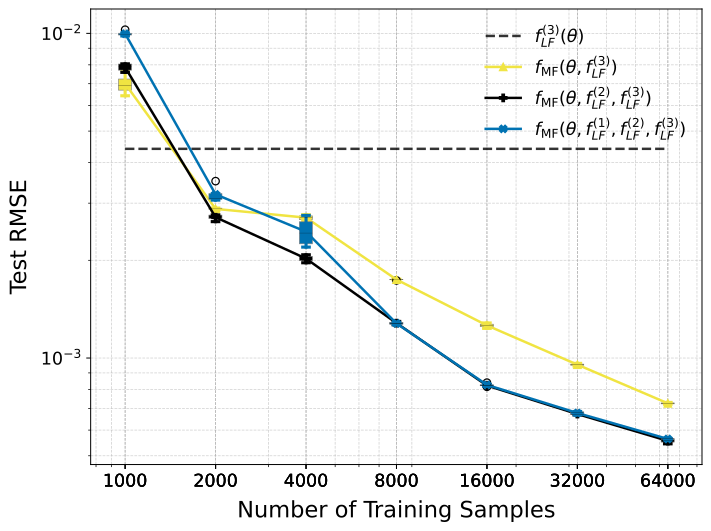}
        \caption{Third-level surrogate, $\mathbf{f}_\mathrm{MF}^{(3)}$.}
    \end{subfigure}\hfill
    \begin{subfigure}[b]{0.48\linewidth}
        \includegraphics[width=\linewidth]{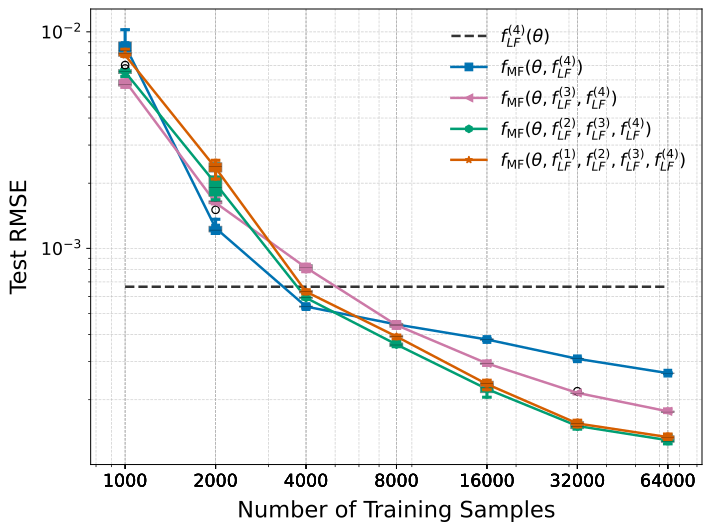}
        \caption{Fourth-level surrogate, $\mathbf{f}_\mathrm{MF}^{(4)}$.}
    \end{subfigure}
    \caption{Accuracy of the multi-fidelity surrogate models for different levels and training sample sizes, varying the amount of input coarse solver information.}
    \label{fig:MFNN_II}
\end{figure}

We assess the accuracy of the multi-fidelity models incrementally, from level $l=1$ to the level $l=4$.  For each level, we evaluate whether the corresponding multi-fidelity \ac{NN} enhances the predictive accuracy of the associated low-fidelity solver, and whether incorporating information from coarser levels contributes positively to the prediction performance.

In Figure~\ref{fig:MFNN_II}a, the multi-fidelity surrogate $\mathbf{f}_\mathrm{MF}^{(1)}$, which receives as input {$\boldsymbol{\theta}$ and evaluations} from the low-fidelity solver $\mathbf{f}_\mathrm{LF}^{(1)}$, achieves a \ac{RMSE} nearly one order of magnitude lower than that of the solver, for $N_{\mathrm{train}} = 64{,}000$. By comparison, a standard (or single-fidelity) \ac{NN} $\mathbf{f}_\mathrm{SF}$ trained using only $\boldsymbol{\theta}$, without low-fidelity inputs, yields a \ac{RMSE} almost three times higher, confirming the advantage of incorporating low-fidelity information.

Figure~\ref{fig:MFNN_II}b reports the results for the second-level surrogate $\mathbf{f}_\mathrm{MF}^{(2)}$, which takes as input both $\mathbf{f}_\mathrm{LF}^{(1)}$ and $\mathbf{f}_\mathrm{LF}^{(2)}$. Also in this case, the multi-fidelity network surpasses the predictive accuracy of $\mathbf{f}_\mathrm{LF}^{(2)}$. Notably, a version using only $\mathbf{f}_\mathrm{LF}^{(2)}$ as input provides nearly equivalent performance, suggesting no benefits when $\mathbf{f}_\mathrm{LF}^{(1)}$ is added.

The third-level surrogate $\mathbf{f}_\mathrm{MF}^{(3)}$ (Figure~\ref{fig:MFNN_II}c), which combines the outputs from $\mathbf{f}_\mathrm{LF}^{(1)}$, $\mathbf{f}_\mathrm{LF}^{(2)}$, and $\mathbf{f}_\mathrm{LF}^{(3)}$, again shows improved accuracy over $\mathbf{f}_\mathrm{LF}^{(3)}$. In contrast to the second level, here the inclusion of $\mathbf{f}_\mathrm{LF}^{(2)}$ proves beneficial, showing the usefulness of multi-fidelity fusion.

Finally, Figure~\ref{fig:MFNN_II}d illustrates the performance of the fourth-level surrogate $\mathbf{f}_\mathrm{MF}^{(4)}$, which aggregates all available low-fidelity solver evaluations. The resulting model achieves a \ac{RMSE} approximately four times lower than that of the best-performing low-fidelity solver at $N_{\mathrm{train}} = 64{,}000$. The figure also compares variants using only subsets of the low-fidelity inputs, confirming that including coarser solvers as inputs systematically enhances accuracy, with the only exclusion of model $\mathbf{f}_\mathrm{LF}^{(1)}$. This however might be reasonable, given the extremely coarse mesh the surrogate $\mathbf{f}_\mathrm{LF}^{(1)}$ has been built on.

Overall, the results consistently show that multi-fidelity \ac{NN}s outperform the corresponding low-fidelity solvers at all levels. The integration of information from multiple fidelities contributes significantly to predictive performance. Based on these findings, all subsequent experiments fix the number of training samples to $N_{\mathrm{train}} = 16{,}000$. For consistency, we retain model $\mathbf{f}_\mathrm{LF}^{(1)}$, as it is computationally inexpensive and beneficial at the first level. In addition, the inclusion of its information does not decrease the accuracy of the models.

A comprehensive summary of the spatial discretization, accuracy, and computational cost of the high-fidelity, low-fidelity, and multi-fidelity surrogate models is provided in Table~\ref{tab:model_performance1}.

\subsection{MFDA: Online Inference}\label{sec:online1}
\begin{figure}[t]
    \centering
    \begin{subfigure}[t]{0.35\linewidth}
        \centering
        \includegraphics[width=\linewidth]{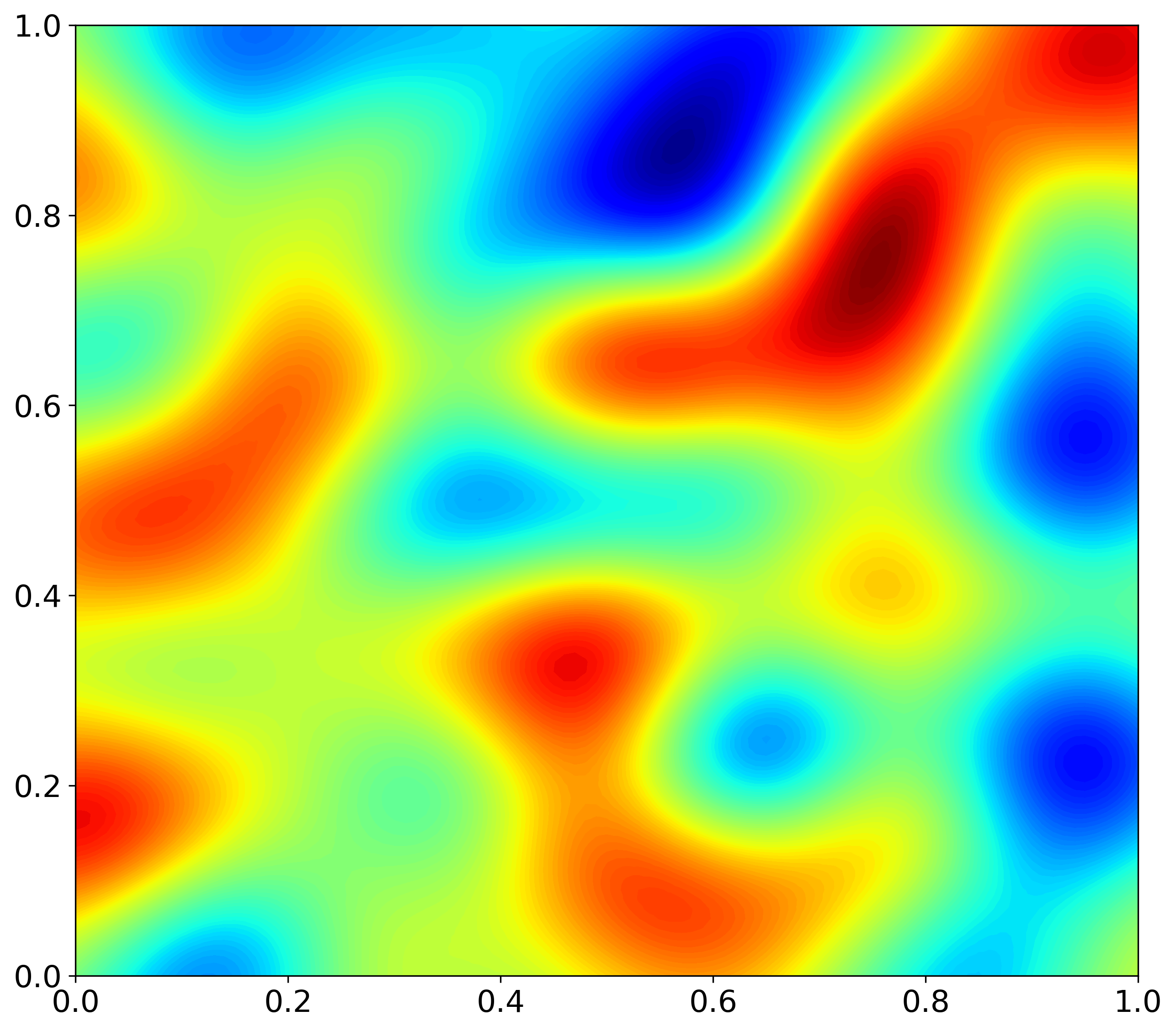}
        \caption{}
    \end{subfigure}
    \begin{subfigure}[t]{0.35\linewidth}
        \centering
        \includegraphics[width=\linewidth]{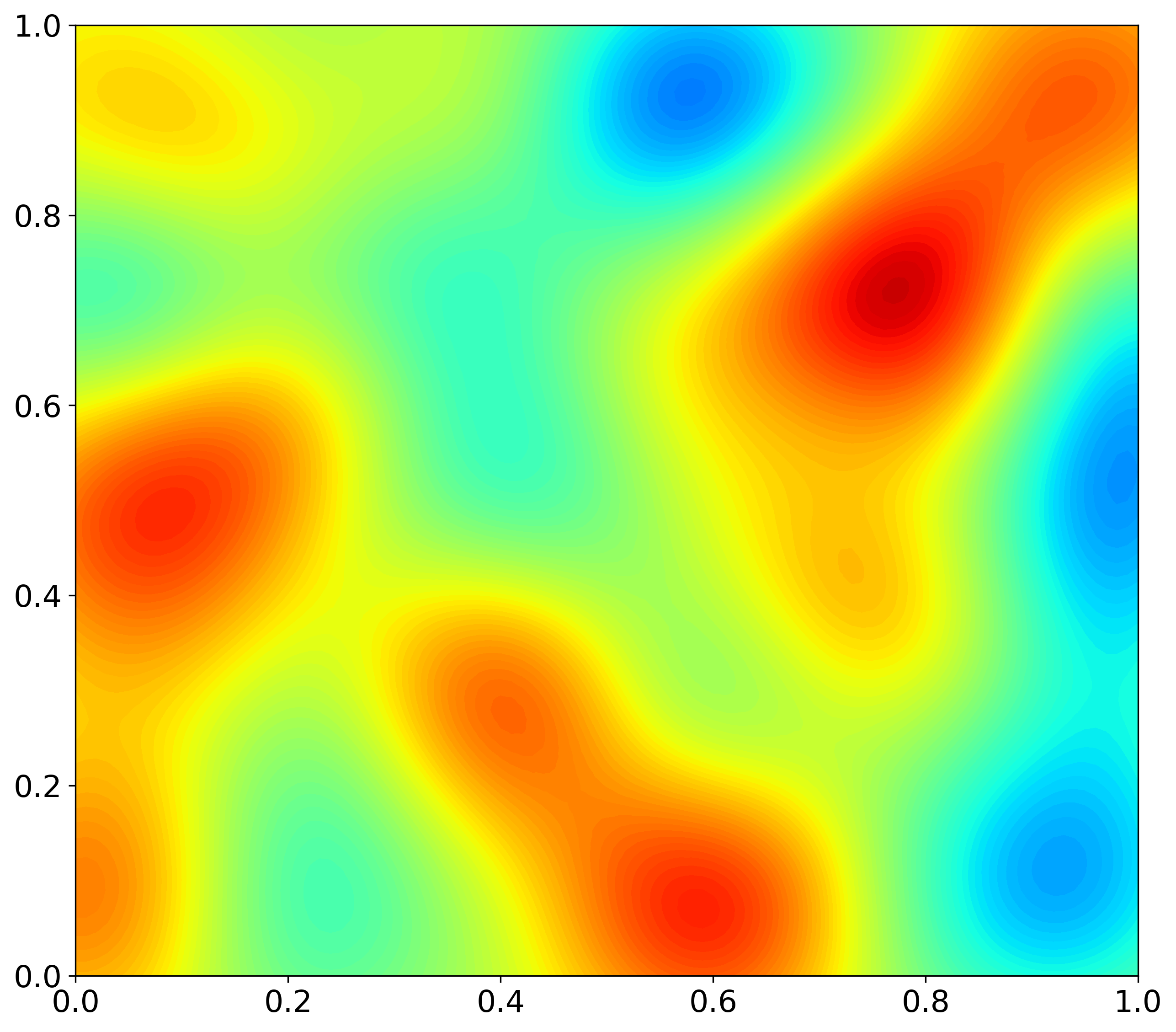}
        \caption{}
    \end{subfigure}
    \begin{subfigure}[t]{0.35\linewidth}
        \centering
        \includegraphics[width=\linewidth]{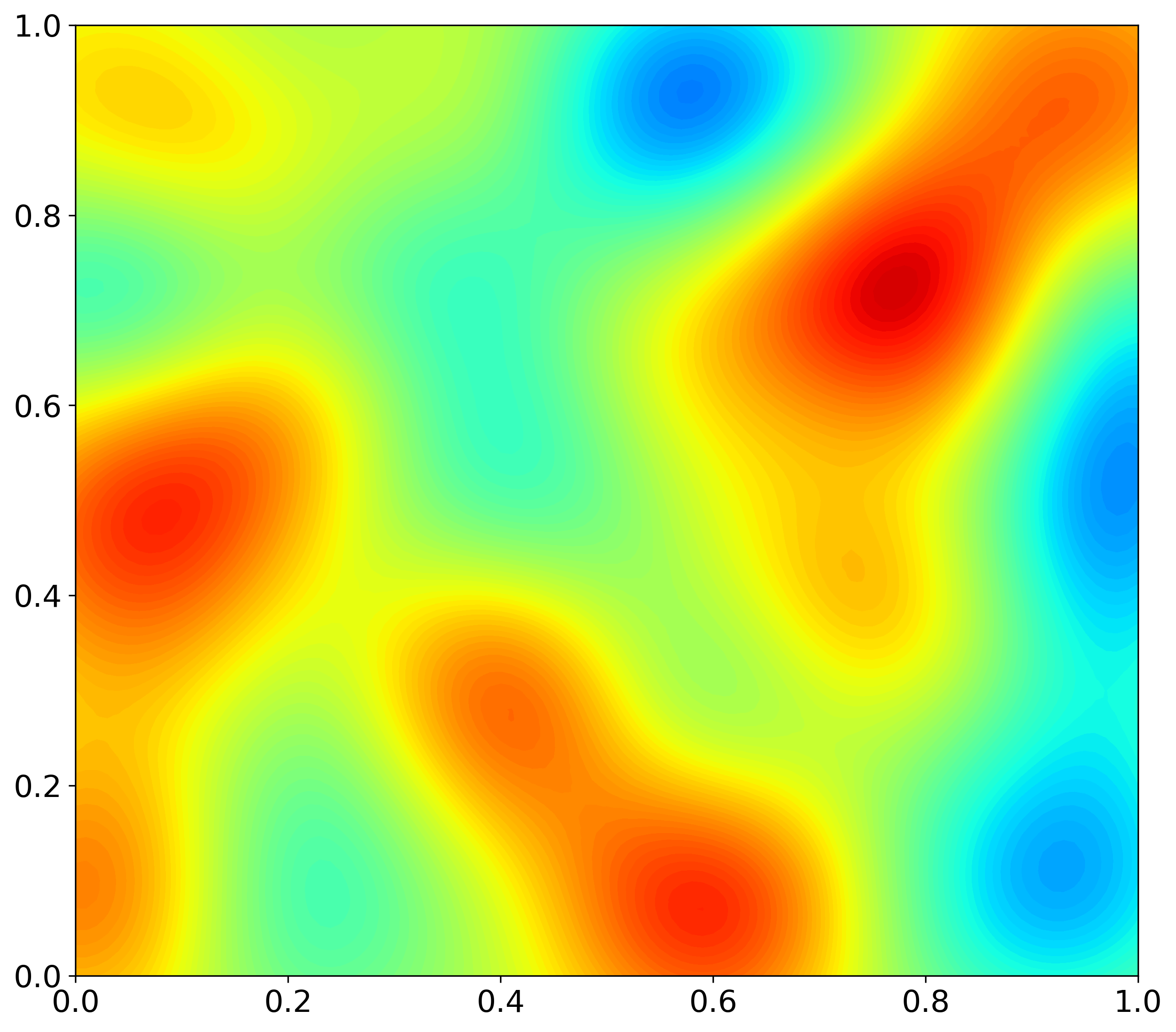}
        \caption{}
    \end{subfigure}
    \begin{subfigure}[t]{0.35\linewidth}
        \centering
        \includegraphics[width=\linewidth]{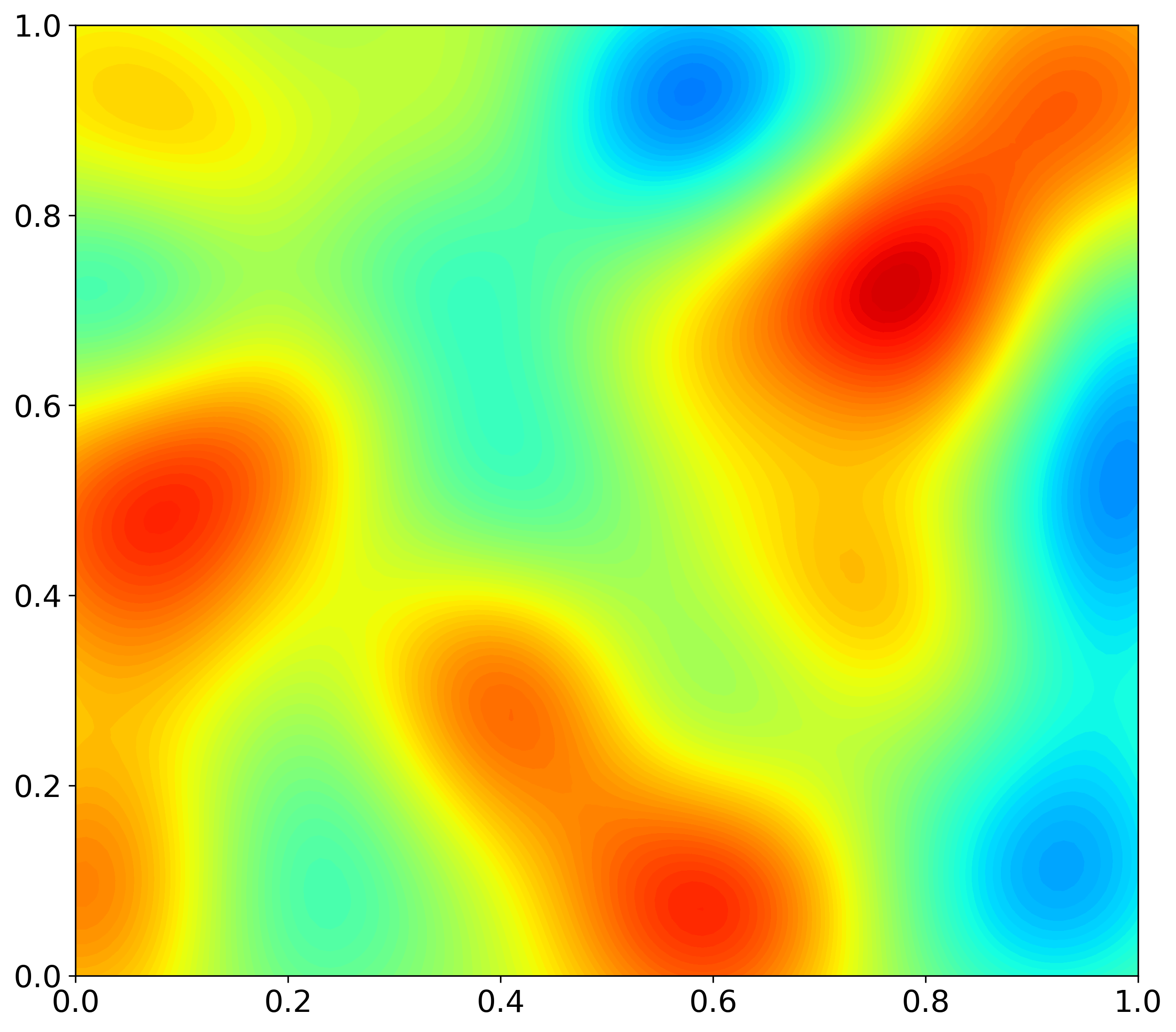}
        \caption{}
    \end{subfigure}
    \caption{
    (a) Exact transmissivity field $\boldsymbol{\theta}$ to be identified. 
    (b--d) Posterior mean transmissivity fields reconstructed using (b) MH, (c) MLDA, and (d) MFDA, respectively.
    }
    \label{fig:acc1}
\end{figure}

\begin{table}[b]
    \centering
    \begin{tabular}{lllll}
        \toprule
        \textbf{Scheme} & \textbf{Level} & \textbf{Forward model} & \textbf{Inputs} & \textbf{Sub-chain Length}\\
        \midrule
        MH & only 1 level & $\mathbf{f}_\mathrm{HF}$ & $\boldsymbol{\theta}$ & -\\

        \midrule
        \multirow{5}{*}{MLDA} 
            & 1 & $\mathbf{f}_\mathrm{LF}^{(1)}$ & $\boldsymbol{\theta}$ & 5 \\
            & 2 & $\mathbf{f}_\mathrm{LF}^{(2)}$ & $\boldsymbol{\theta}$ & 2 \\
            & 3 & $\mathbf{f}_\mathrm{LF}^{(3)}$ & $\boldsymbol{\theta}$ & 2 \\
            & 4 & $\mathbf{f}_\mathrm{LF}^{(4)}$ & $\boldsymbol{\theta}$ & 1 \\
            & 5 & $\mathbf{f}_\mathrm{HF}$ & $\boldsymbol{\theta}$ & -\\
                \midrule
        
        \multirow{4}{*}{MFDA} 
            & 1 & $\mathbf{f}_\mathrm{MF}^{(1)}$ & $\boldsymbol{\theta}, \mathbf{f}_\mathrm{LF}^{(1)}$ & 10 \\
            & 2 & $\mathbf{f}_\mathrm{MF}^{(2)}$ & $\boldsymbol{\theta}, \mathbf{f}_\mathrm{LF}^{(1)}$, $\mathbf{f}_\mathrm{LF}^{(2)}$ & 2 \\
            & 3 & $\mathbf{f}_\mathrm{MF}^{(3)}$ & $\boldsymbol{\theta}, \mathbf{f}_\mathrm{LF}^{(1)}$, $\mathbf{f}_\mathrm{LF}^{(2)}$, $\mathbf{f}_\mathrm{LF}^{(3)}$ & 1\\
            & 4 & $\mathbf{f}_\mathrm{MF}^{(4)}$ & $\boldsymbol{\theta}, \mathbf{f}_\mathrm{LF}^{(1)}$, $\mathbf{f}_\mathrm{LF}^{(2)}$, $\mathbf{f}_\mathrm{LF}^{(3)}$,
            $\mathbf{f}_\mathrm{LF}^{(4)}$ & -\\
            \bottomrule
    \end{tabular}
        \caption{Likelihood evaluation strategies for each scheme and level during the online phase. MFDA entries specify the surrogate and its required inputs.}
    \label{tab:online_phase_config}
\end{table}

\begin{figure}[t]
    \centering
    \begin{minipage}[t]{0.48\linewidth}
        \centering
        \includegraphics[width=\linewidth]{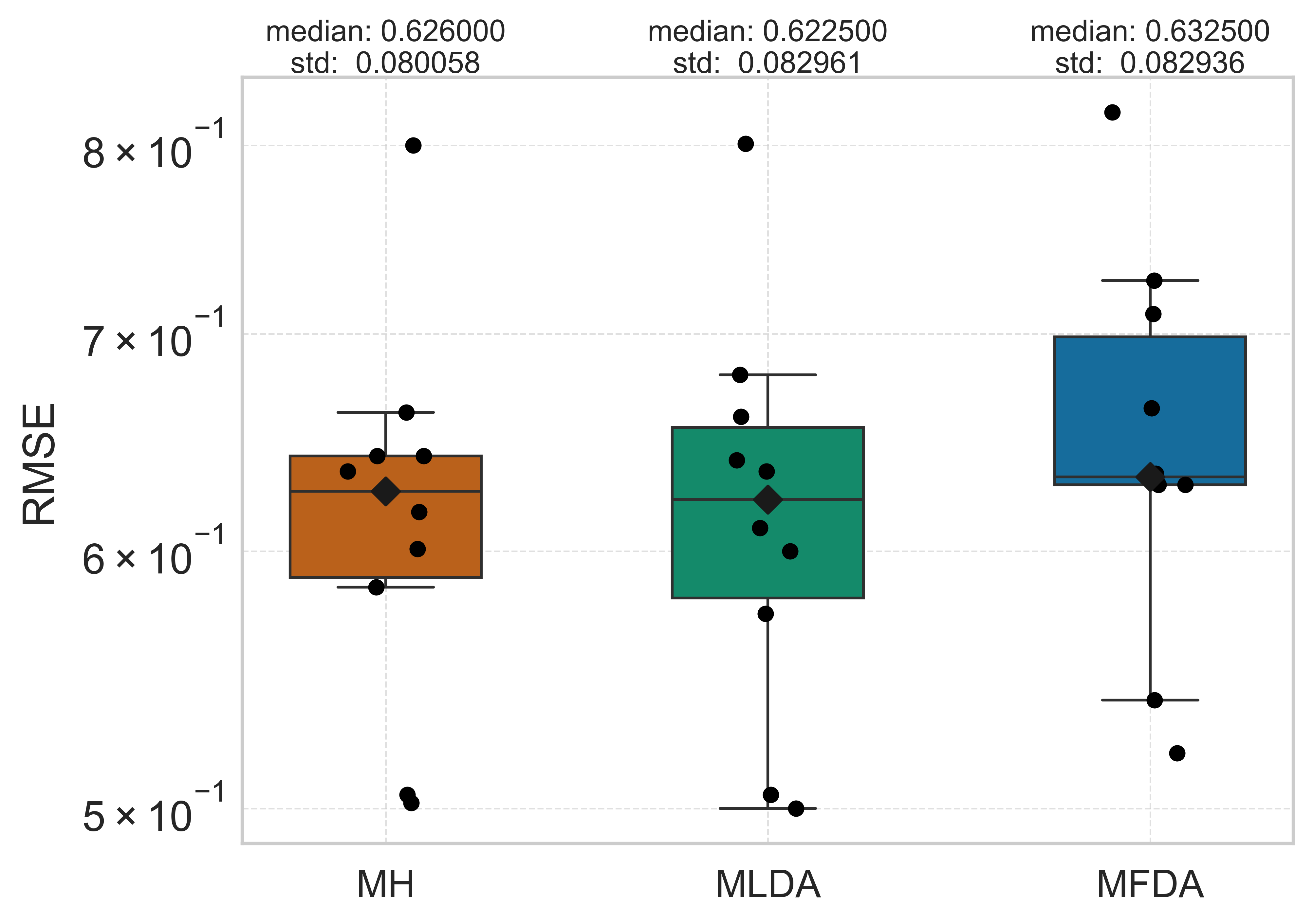}
        \captionof{figure}{Root mean square error (RMSE) of reconstructed parameters across 10 instances for all sampling schemes.}
        \label{fig:acc2}
    \end{minipage}%
    \hfill
    \begin{minipage}[t]{0.48\linewidth}
        \centering
        \includegraphics[width=\linewidth]{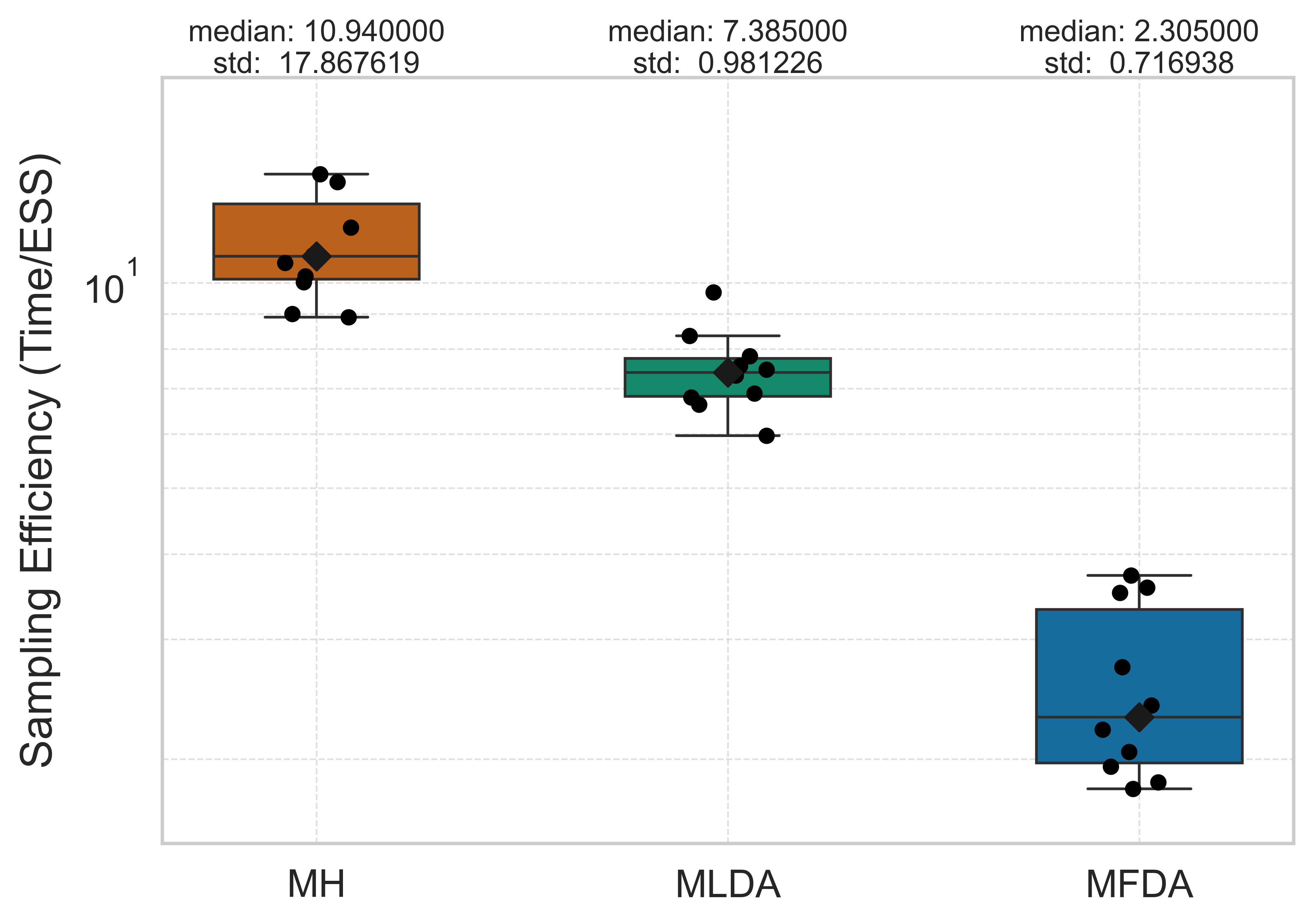}
        \captionof{figure}{Sampling efficiency (computation time per ESS) across 10 instances for all sampling schemes.}
        \label{fig:sampling_efficiency}
    \end{minipage}
\end{figure}

Synthetic observations at the sensor locations $\{\mathbf{x}_j\}_{j=1}^d$ are generated by evaluating the high-fidelity model and adding independent Gaussian noise,
\[
\mathbf{y}^{\mathrm{obs}} = \mathbf{f}_{\mathrm{HF}}(\boldsymbol{\theta}) + \boldsymbol{\varepsilon}, 
\qquad \boldsymbol{\varepsilon} \sim \mathcal{N}(0,0.01^2 I_d).
\]
The aim is to characterize the posterior distribution $\pi(\boldsymbol{\theta}\mid\mathbf{y}^{\mathrm{obs}})$ and to compare the sampling efficiency of MFDA with that of MH and MLDA.

In MFDA, the high-fidelity model is evaluated only during the offline stage to produce the training dataset. During inference, the likelihood is evaluated using the multi-fidelity surrogates $\mathbf{f}_{\mathrm{MF}}^{(l)}$, which combine $\boldsymbol{\theta}$ with low-fidelity forward evaluations. In contrast, MLDA evaluates a hierarchy of low- and high-fidelity solvers online, while MH relies solely on the high-fidelity model.

The sampling schemes, the forward models used at every level and the corresponding sub-chain lengths are summarized in Table~\ref{tab:online_phase_config}. Longer sub-chains are assigned to coarser levels so that most evaluations occur at inexpensive models. Five independent MCMC chains are run, and convergence is monitored every 100 iterations using the Gelman–Rubin statistic $\hat{R}$ \cite{rubin}. Sampling is terminated once the maximum $\hat{R}$ across all components of $\boldsymbol{\theta}$ falls below $1.01$.

Figure~\ref{fig:acc1} shows the posterior mean reconstructions of the transmissivity field obtained with MH, MLDA, and MFDA, together with the ground truth, for one random instance. All schemes recover the main spatial features with comparable accuracy. To assess robustness, posterior sampling is repeated for ten independent realizations of $\boldsymbol{\theta}$ and corresponding synthetic observations. The distribution of the resulting \ac{RMSE} values is reported in Fig.\ref{fig:acc2}, confirming that the reconstruction quality is similar across the three methods. In particular, MH and MLDA provide almost identical accuracy, while MFDA provide a median RMSE 1\% larger, which is fully acceptable given the computational gains. Sampling efficiency, measured as computation time per ESS, is shown in Fig.\ref{fig:sampling_efficiency}. The MFDA scheme achieves the best performance, reducing the cost per effective sample by about a factor of five relative to MH and by about a factor of 3 relative to MLDA.

To quantify the overall computational gains, Table~\ref{tab:cost_summary} reports the offline and online wall-clock costs associated with each scheme. The mean online wall-clock time per inference is reduced by about 75\% compared to MH and by about 70\% compared to MLDA. Although MFDA requires a one-time offline effort for data generation and surrogate training, the overall cost per inference remains roughly 50\% lower than MH and about 30\% lower than MLDA, even when this offline stage is included. Moreover, the offline cost is incurred only once: in scenarios requiring repeated inference, MFDA reuses the trained surrogates, and its effective cost reduces to the online stage alone.

In summary, the MFDA approach attains posterior reconstruction accuracy comparable to that of MLDA and standard MH, while delivering substantially greater computational efficiency in Bayesian inference for the considered groundwater flow model. 

\begin{table}[b!]
\centering
\renewcommand{\arraystretch}{1.25}
\begin{tabular}{lllll}
\toprule
\textbf{Scheme} & 
\textbf{Data Gen. [s]} & 
\textbf{Training [s]} &
\textbf{Online Cost [s]} &
\textbf{Total Cost [s]} \\
\midrule
MH   
& 0 & 0 
& $12078 \,\pm\, 1600$ 
& $12078 \,\pm\, 1600$ \\

MLDA 
& 0 & 0 
& $8794 \,\pm\, 2044$ 
& $8794 \,\pm\, 2044$ \\

MFDA 
& $2367$ 
& $1655$ 
& $2722 \,\pm\, 928$ 
& $6744 \,\pm\, 928$ \\
\bottomrule
\end{tabular}
\caption{Computational cost comparison among sampling schemes. 
MFDA incurs a one-time offline cost due to dataset generation and surrogate training, 
but achieves a substantially lower online and total inference cost. 
Online cost values report mean~$\pm$~standard deviation over 10 independent inverse problems (i.e., 10 realizations of $\boldsymbol{\theta}$ and $\mathbf{y}^{\mathrm{obs}}$).}
\label{tab:cost_summary}
\end{table}

\section{Numerical Experiments II:  Reaction--Diffusion Equation}
\label{sec:res2}

To further assess the performance of the MFDA algorithm, we consider a nonlinear, time-dependent reaction-diffusion system as a second benchmark \cite{champion2019,conti_multi-fidelity_2024}. This case study is characterized by higher-dimensional output, offering a challenging test for data-driven models and posterior inference. The governing equations are:
\begin{equation}
    \begin{aligned}
        \dot{u} &= \left(1 - (u^2 + v^2)\right) u + \mu_1 (u^2 + v^2) v + \mu_2\left(u_{xx} + u_{yy}\right), \\
        \dot{v} &= -\mu_1 (u^2 + v^2) u + \left(1 - (u^2 + v^2)\right) v + \mu_2\left(v_{xx} + v_{yy}\right),
    \end{aligned}
    \label{eq:RD_eqs}
\end{equation}
where $u(x, y, t)$ and $v(x, y, t)$ represent two interacting species. The parameters of interest are the nonlinear reaction coefficient $\mu_1\in (0.5, 1.5)$ and the diffusion coefficient $\mu_2 \in (0.01, 0.1)$, therefore $\boldsymbol{\mu} = [\mu_1,\mu_2] \in \mathcal{M}  = (0.5, 1.5) \times (0.01, 0.1) \subseteq \mathbb{R}^2$. The domain is $(x, y) \in (-L, L)^2$ with $L = 20$ and periodic boundary conditions. The system is integrated in time over $t \in [0, 50]$ with initial conditions:
\begin{equation*}
    u(x, y, 0) = v(x, y, 0) = \tanh\left(\sqrt{x^2 + y^2} \cos\left((x + i y) - \sqrt{x^2 + y^2}\right)\right).
\end{equation*}
Observations are collected at $13 \times 13$ spatial sensor locations for both $u$ and $v$. We set $N_s=13\times13\times2$. The measurements are taken every 0.2 seconds, resulting in $T=250$ time steps and a total observation vector of size $d=N_s\times T=13 \times 13 \times 2 \times 250$.

\subsection{MFDA: Setting}

\begin{figure}[t]
    \centering
    \includegraphics[width=\linewidth]{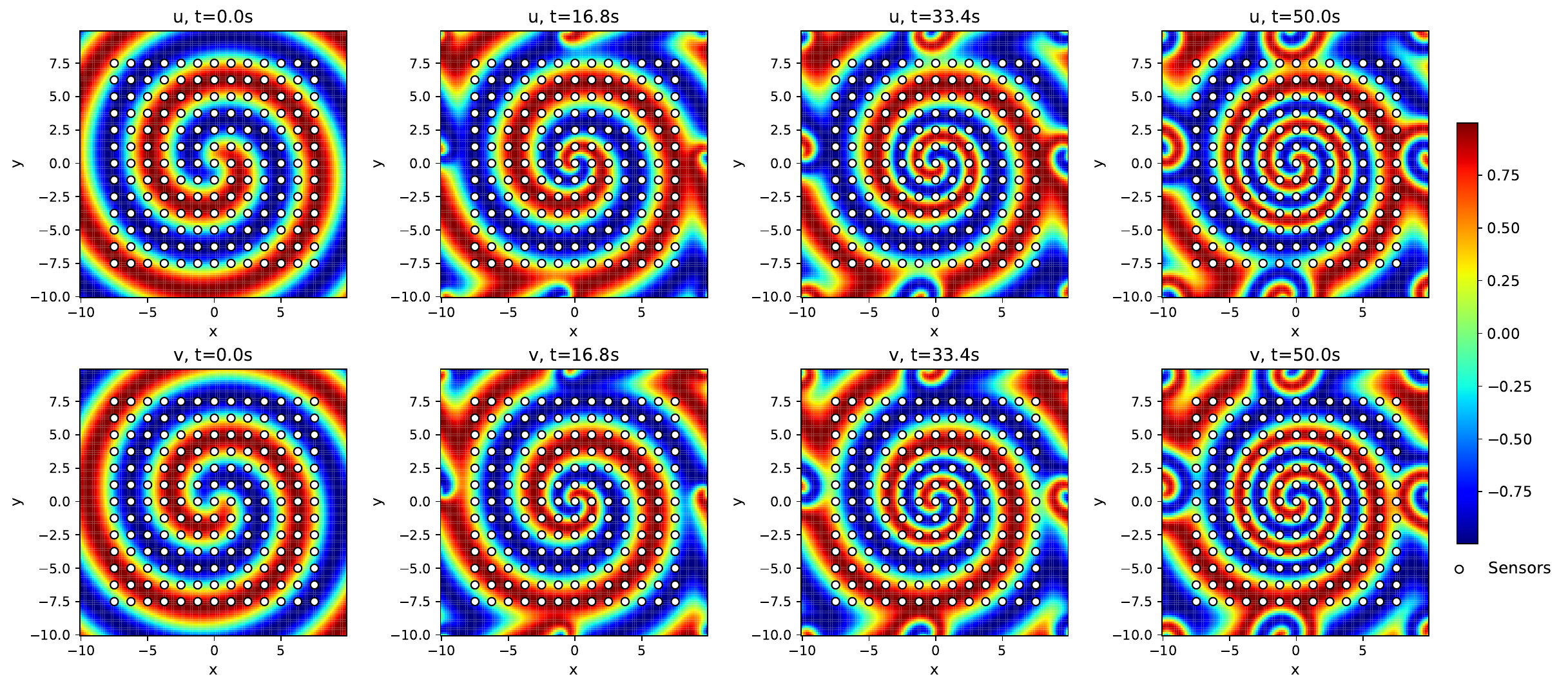}
    \caption{Reference solution for the diffusion reaction problem obtained using high fidelity model at four distinct time instants along with the observation sensor locations.}
    \label{fig:sensors}
\end{figure} 
\begin{figure}[t]
\centering

\begin{subfigure}{0.31\textwidth}
    \centering
    \includegraphics[width=\linewidth]{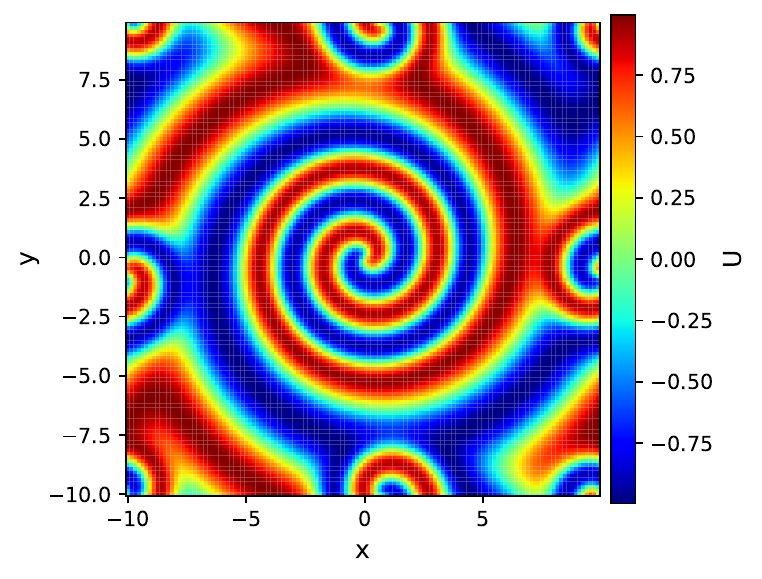}
    \caption{$\mathbf{g}_{\mathrm{HF}}$}
    \label{fig:U_hf}
\end{subfigure}

\begin{subfigure}{0.31\textwidth}
    \centering
    \includegraphics[width=\linewidth]{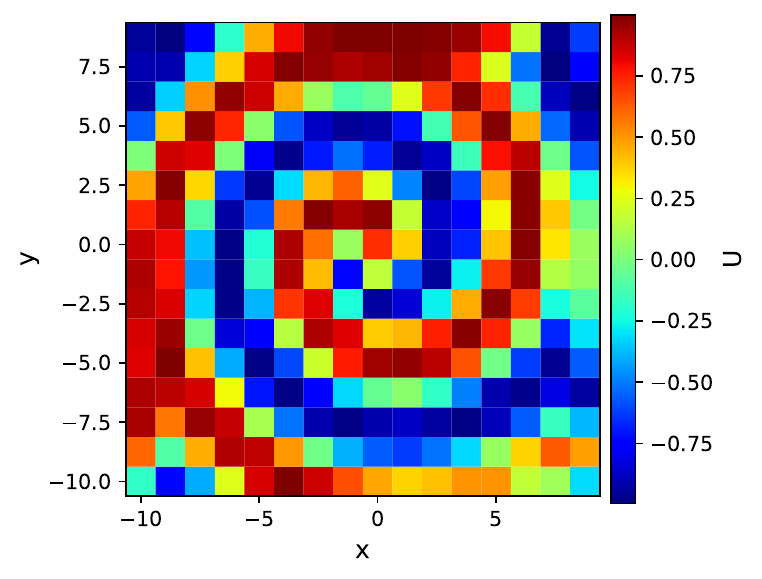}
    \caption{$\mathbf{g}_{\mathrm{LF}}^{(1)}$}
    \label{fig:U_lf1}
\end{subfigure}%
\begin{subfigure}{0.31\textwidth}
    \centering
    \includegraphics[width=\linewidth]{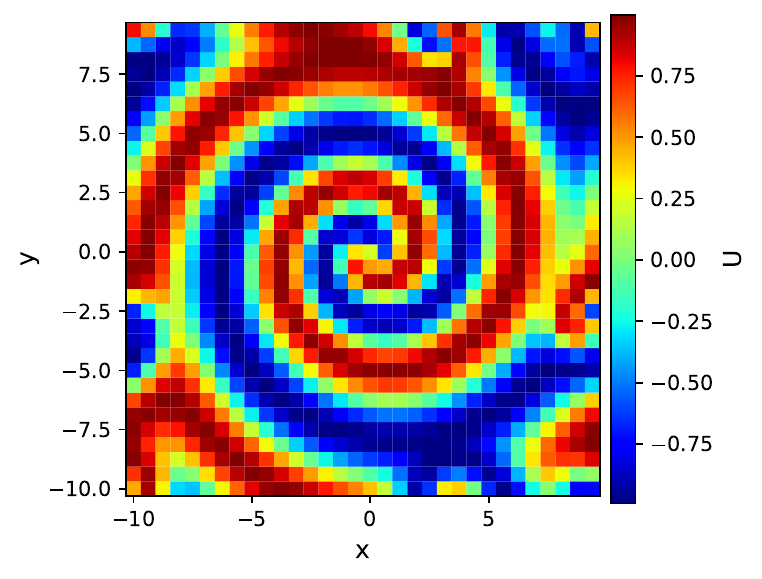}
    \caption{$\mathbf{g}_{\mathrm{LF}}^{(2)}$}
    \label{fig:U_lf2}
\end{subfigure}%
\begin{subfigure}{0.31\textwidth}
    \centering
    \includegraphics[width=\linewidth]{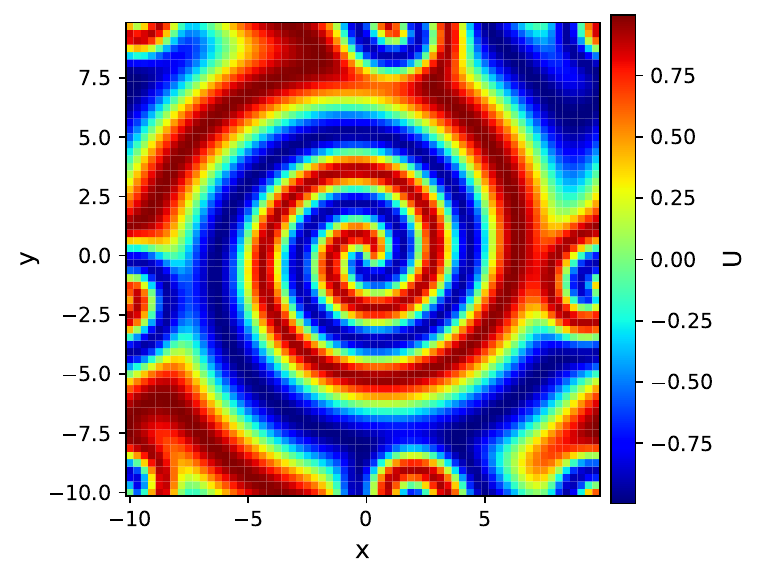}
    \caption{$\mathbf{g}_{\mathrm{LF}}^{(3)}$}
    \label{fig:U_lf3}
\end{subfigure}

\begin{subfigure}{0.31\textwidth}
    \centering
    \includegraphics[width=\linewidth]{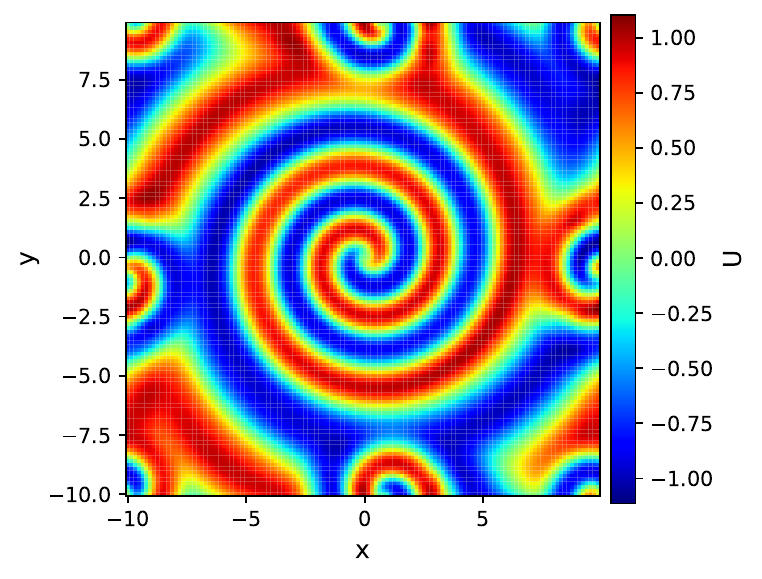}
    \caption{$\mathbf{g}_{\mathrm{MF}}^{(1)}$}
    \label{fig:U_mf1}
\end{subfigure}%
\begin{subfigure}{0.31\textwidth}
    \centering
    \includegraphics[width=\linewidth]{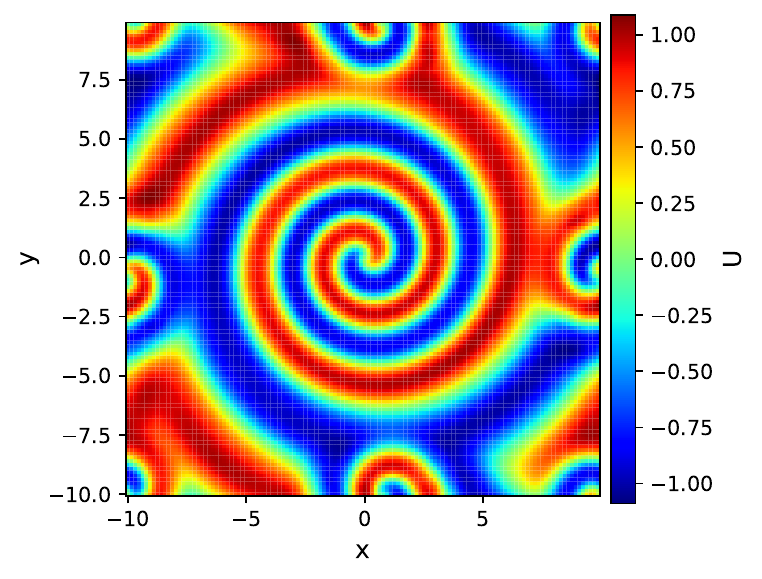}
    \caption{$\mathbf{g}_{\mathrm{MF}}^{(2)}$}
    \label{fig:U_mf2}
\end{subfigure}%
\begin{subfigure}{0.31\textwidth}
    \centering
    \includegraphics[width=\linewidth]{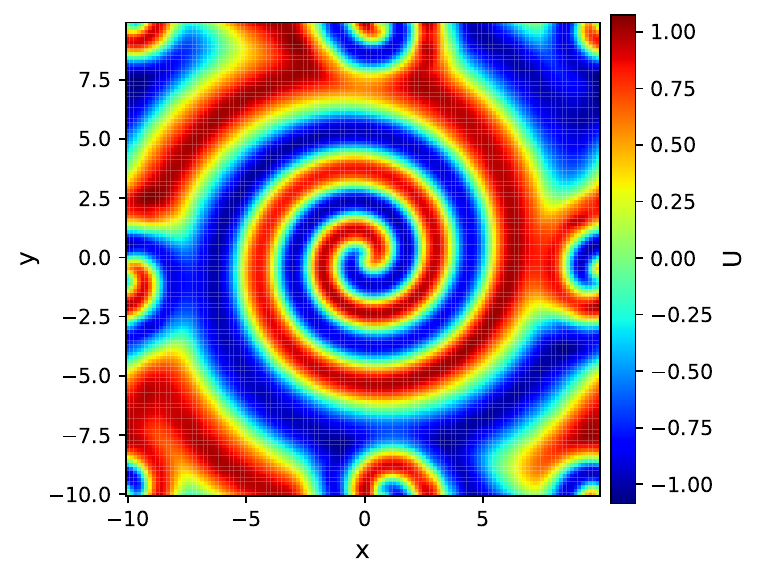}
    \caption{$\mathbf{g}_{\mathrm{MF}}^{(3)}$}
    \label{fig:U_mf3}
\end{subfigure}

\caption{Final-time full field numerical solutions for $u$. (a) HF reference; (b–d) LF solver outputs at increasing spatial resolution; (e–g) MF surrogate predictions corresponding to the LF levels. }
\label{fig:sol2}
\end{figure}

\begin{figure}[t]
\centering
\begin{subfigure}{0.31\textwidth}
    \centering
    \includegraphics[width=\linewidth]{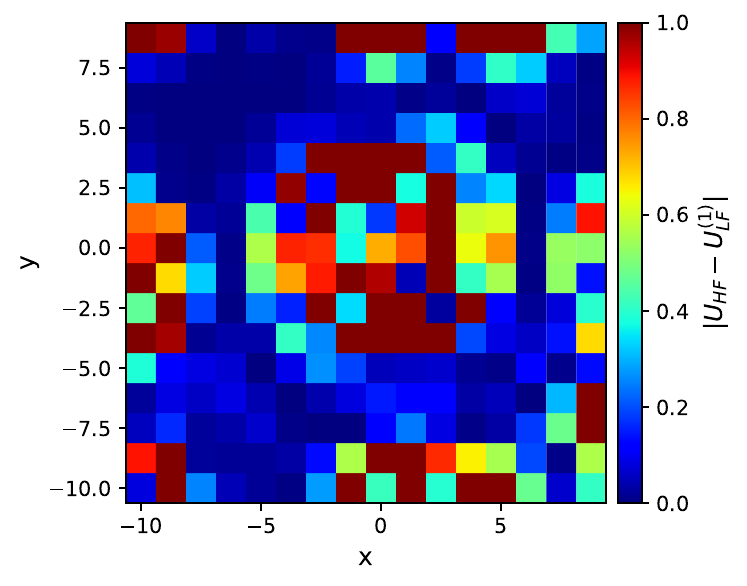}
    \caption{Error: $\mathbf{g}_{\mathrm{LF}}^{(1)}$}
    \label{fig:err_lf1}
\end{subfigure}%
\begin{subfigure}{0.31\textwidth}
    \centering
    \includegraphics[width=\linewidth]{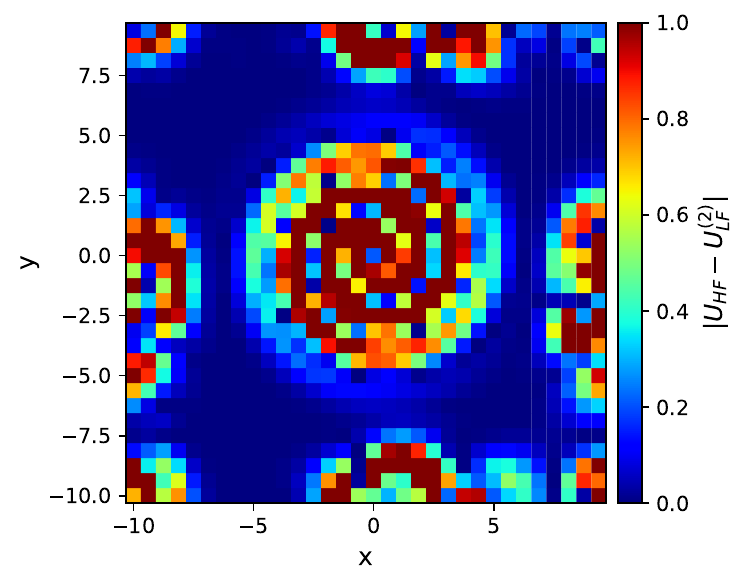}
    \caption{Error: $\mathbf{g}_{\mathrm{LF}}^{(2)}$}
    \label{fig:err_lf2}
\end{subfigure}%
\begin{subfigure}{0.31\textwidth}
    \centering
    \includegraphics[width=\linewidth]{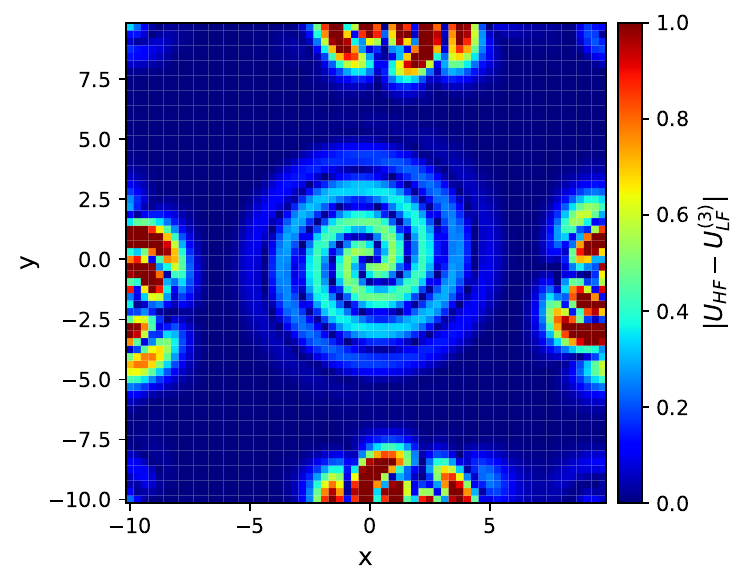}
    \caption{Error: $\mathbf{g}_{\mathrm{LF}}^{(3)}$}
    \label{fig:err_lf3}
\end{subfigure}

\begin{subfigure}{0.31\textwidth}
    \centering
    \includegraphics[width=\linewidth]{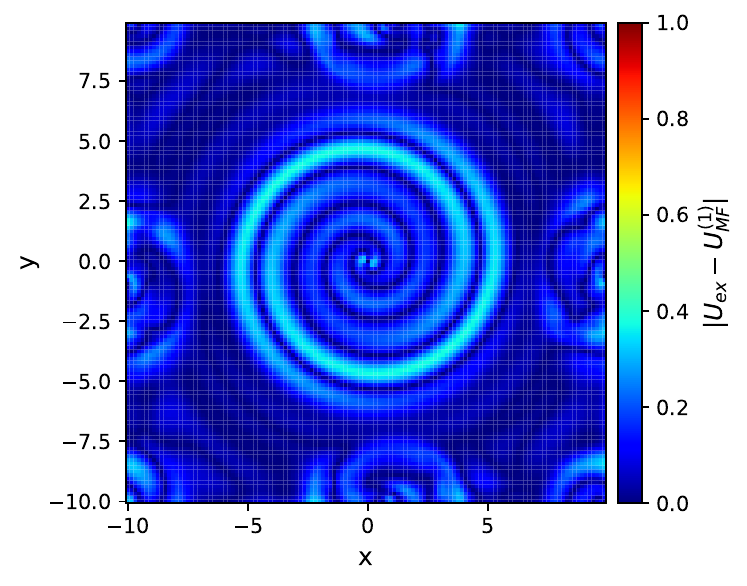}
    \caption{Error: $\mathbf{g}_{\mathrm{MF}}^{(1)}$}
    \label{fig:err_mf1}
\end{subfigure}%
\begin{subfigure}{0.31\textwidth}
    \centering
    \includegraphics[width=\linewidth]{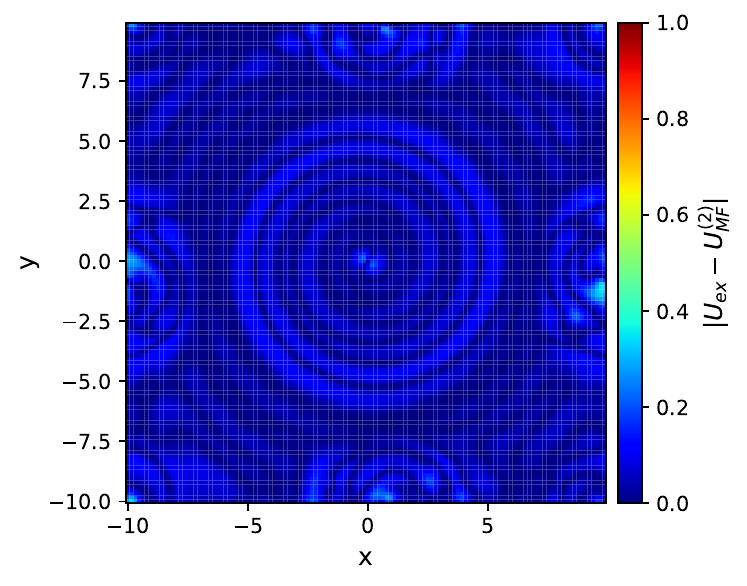}
    \caption{Error: $\mathbf{g}_{\mathrm{MF}}^{(2)}$}
    \label{fig:err_mf2}
\end{subfigure}%
\begin{subfigure}{0.31\textwidth}
    \centering
    \includegraphics[width=\linewidth]{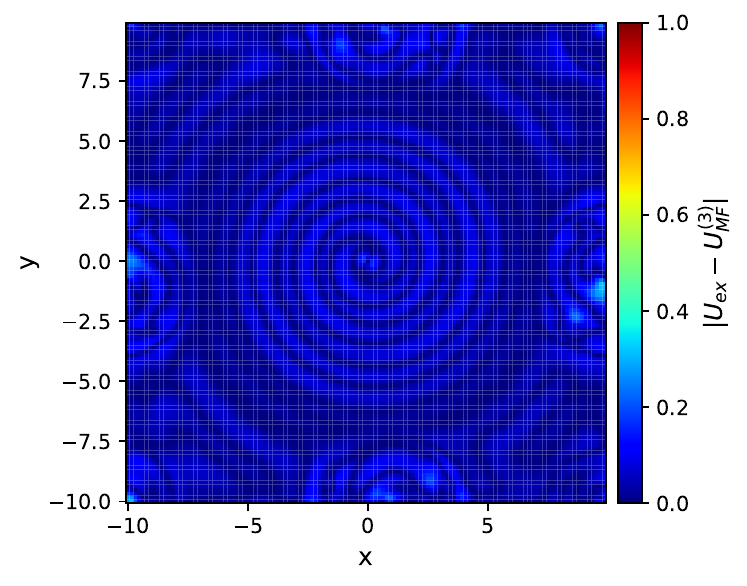}
    \caption{Error: $\mathbf{g}_{\mathrm{MF}}^{(3)}$}
    \label{fig:err_mf3}
\end{subfigure}
\caption{Absolute error fields at final time for the $u$ quantity. (a–c) LF solver errors; (d–f) MF surrogate errors for the corresponding LF levels. Identical color limits are used within each row for direct comparison.}
\label{fig:sol3}
\end{figure}

\begin{figure}[t]
    \centering
    \includegraphics[width=\linewidth]{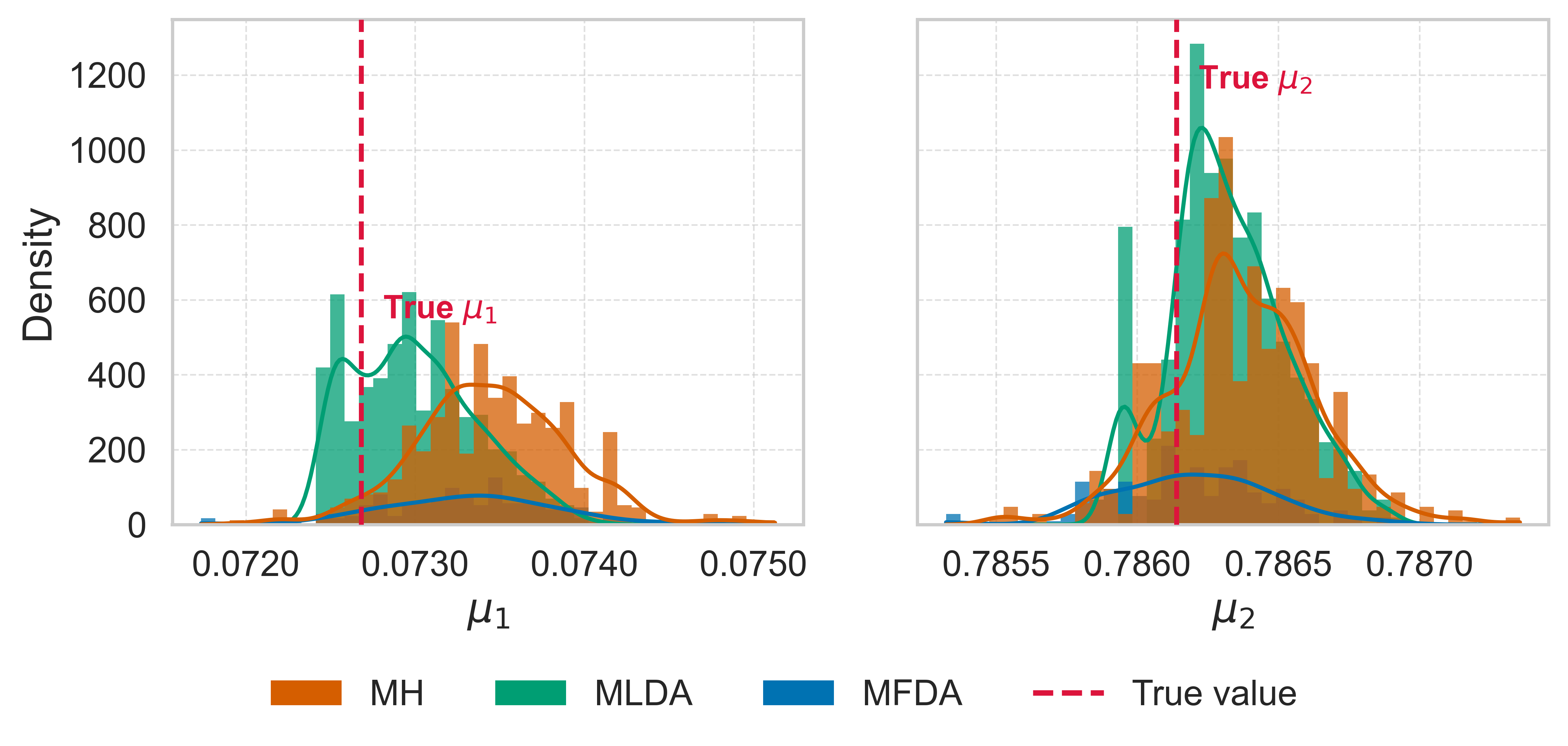   }
    \caption{Posterior distributions for the reaction–diffusion parameters using MH, MLDA, and MFDA.}
    \label{fig:post2}
\end{figure}

\begin{figure}[t]
    \centering
    \includegraphics[width=\linewidth]{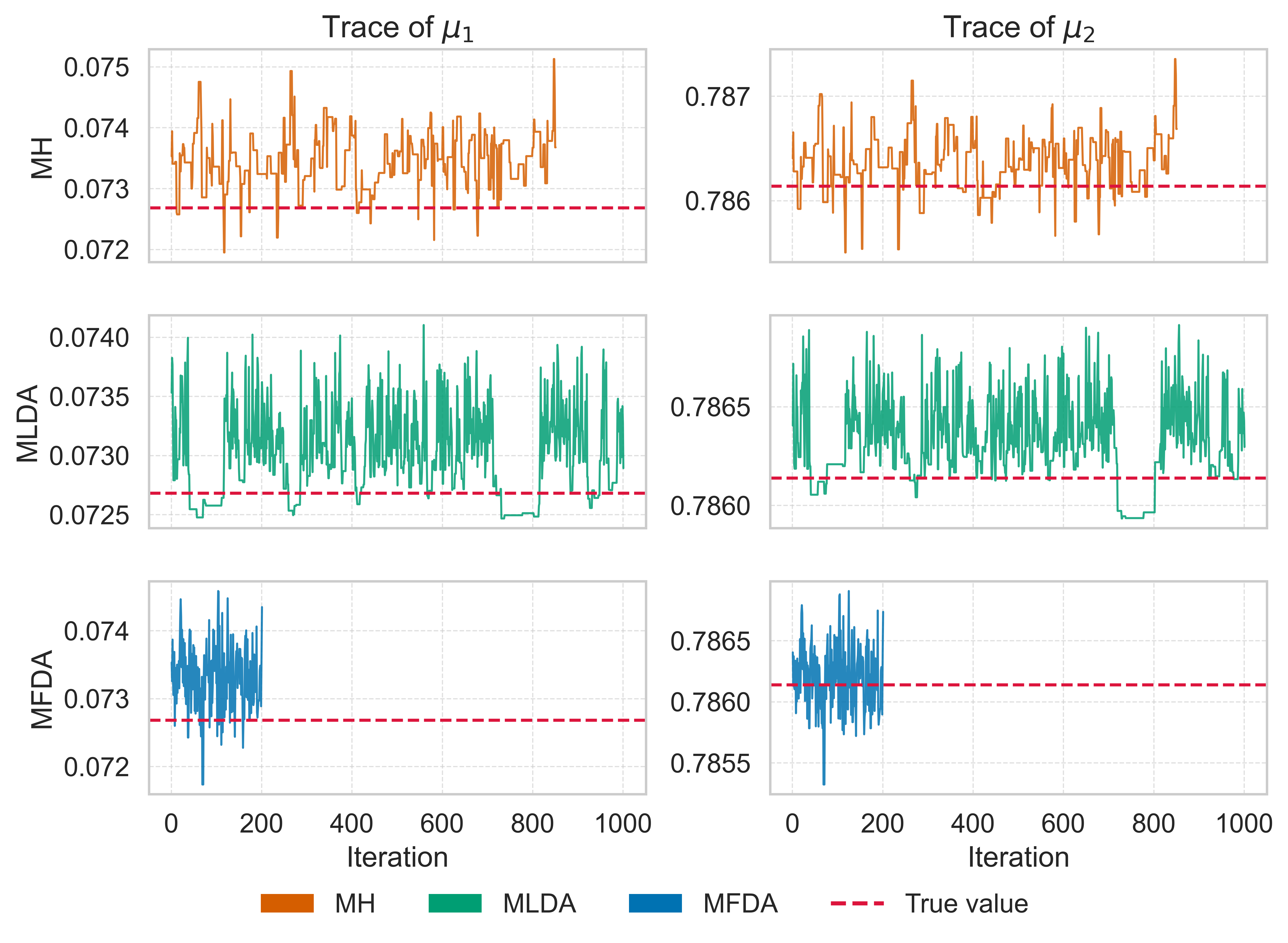}
    \caption{Trace plots for the reaction–diffusion posterior distributions of the parameters obtained using MH, MLDA, and MFDA.}
    \label{fig:trace2}
\end{figure}

\begin{figure}[t]
    \centering
    \begin{minipage}[t]{0.48\linewidth}
        \centering
        \includegraphics[width=\linewidth]{err_stats.png}
        \captionof{figure}{Root mean square error (RMSE) of reconstructed parameters across 10 test instances for all methods.}
        \label{fig:acc22}
    \end{minipage}%
    \hfill
    \begin{minipage}[t]{0.48\linewidth}
        \centering
        \includegraphics[width=\linewidth]{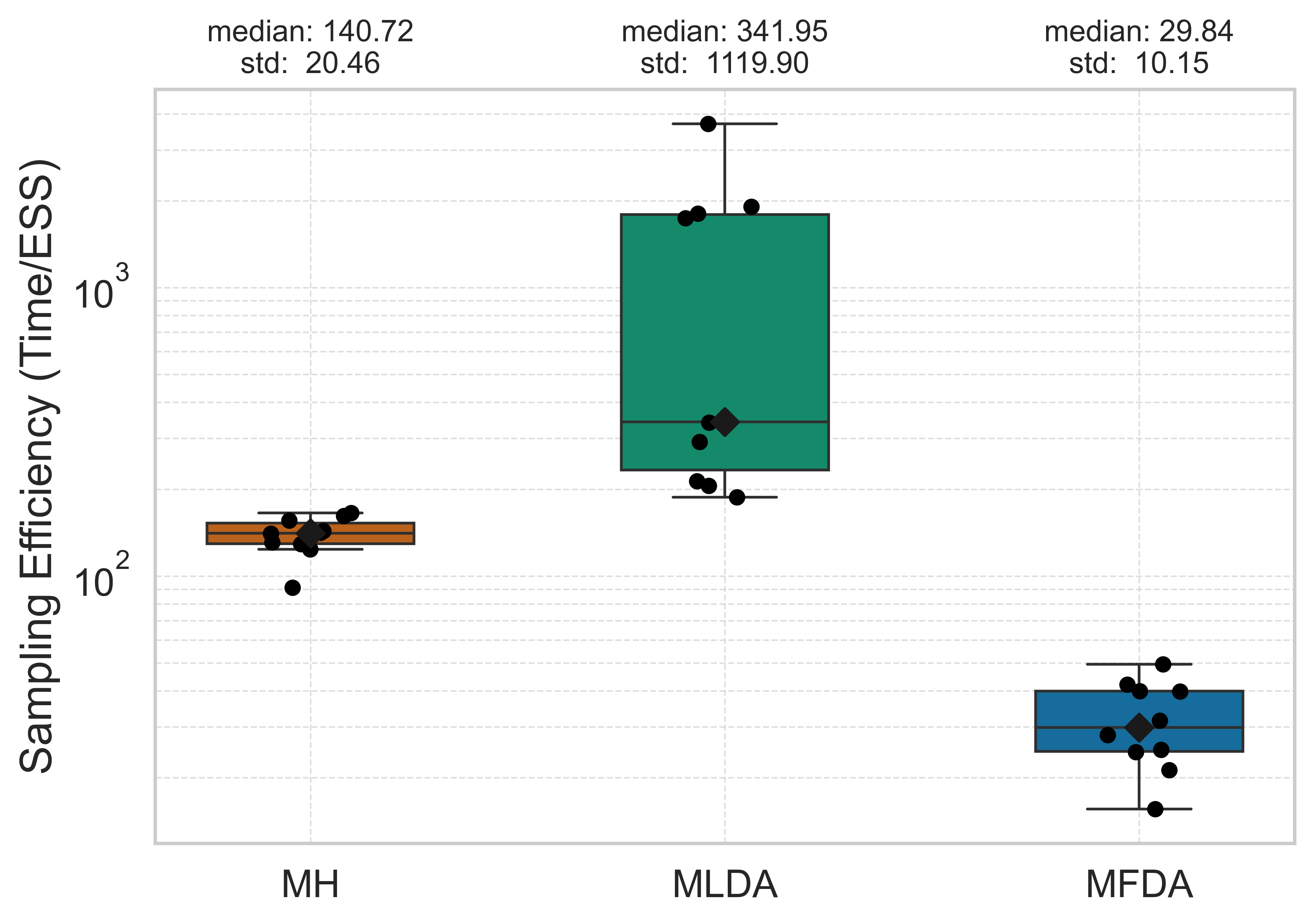}
        \captionof{figure}{Sampling efficiency: computation time per ESS across 10 test instances for all methods.}
        \label{fig:sampling_efficiency_1}
    \end{minipage}
\end{figure}

For this test case, we design a MFDA scheme with $L = 3$ levels. Analogously to before, we define:
\begin{itemize}
    \item \textbf{High-fidelity solver:} 
    The high-fidelity forward model solves problem~\eqref{eq:RD_eqs} numerically using a pseudo-spectral method on a $128 \times 128$ spatial grid, with a time step $\Delta t = 0.2$.
    We denote by $\mathbf{g}_{\mathrm{HF}}(\boldsymbol{\mu})$ the corresponding full-field solution (for both $u$ and $v$) on this grid. 
    The dominant computational cost comes from the two-dimensional Fast Fourier Transform \cite{heideman1984}, with per-step complexity $\mathcal{O}(N \log N)$, where $N = 128^2$. 
    Given $T = 50 / \Delta t = 250$ time steps, the total complexity of the solver is $\mathcal{O}(T \cdot N \log N)$. 
    As before, this high-fidelity solver is used to generate reference data for training the NNs. 
    Evaluating $\mathbf{g}_{\mathrm{HF}}(\boldsymbol{\mu})$ at the $N_s=13\times 13\times 2$ grid nodes corresponding to the sensor locations yields the sensor-level output $\mathbf{f}_{\mathrm{HF}}(\boldsymbol{\mu}) \in  \mathbb{R}^{N_s\times T}$. 
    In addition, it is used to generate the synthetic observations employed in the inverse problem, as outlined in Section \ref{sec:res3}. 
    The reference solution obtained using the high-fidelity model for a random instance of the parameters $\boldsymbol{\mu}$ is shown in Figure~\ref{fig:sensors} along with the sensor configuration at four distinct time instances.

    \item \textbf{Low-fidelity solvers:}  
    A hierarchy of three low-fidelity forward models is constructed by uniformly coarsening both the spatial and temporal discretisations of the high-fidelity solver. 
    Each low-fidelity solver computes a full-field solution $\mathbf{g}_{\mathrm{LF}}^{(l)}(\boldsymbol{\mu})$ on a coarser spectral grid and with a larger time step, $l=1,2,3$. 
    All solvers employ the same pseudo-spectral scheme; only the mesh resolution and time-step size vary across levels.  
    The spatial resolutions, number of time steps, forward evaluation times, and associated errors relative to the high-fidelity solution are summarised in Table~\ref{tab:model_performance2}. 
    Restricting $\mathbf{g}_{\mathrm{LF}}^{(l)}(\boldsymbol{\mu})$ to the sensor locations yields the corresponding sensor-level outputs $\mathbf{f}_{\mathrm{LF}}^{(l)}(\boldsymbol{\mu}) \in  \mathbb{R}^{N_s\times T} $.
    
    \item \textbf{Dimensionality reduction:} 
    To make NN training tractable and mitigate overfitting, we perform Proper Orthogonal Decomposition (POD) on the full-field high-fidelity solutions. 
    Define the snapshot matrix
    \begin{equation} 
    \mathbf{S} = \big[ \mathbf{g}_{\mathrm{HF}}(\boldsymbol{\mu}^{(1)}), \ldots, \mathbf{g}_{\mathrm{HF}}(\boldsymbol{\mu}^{(20)}) \big] 
    \in \mathbb{R}^{N \times (T\times 20)},
    \end{equation}
    where each column corresponds to a vectorised full-field solution at a given time instance and $N$ is the number of spatial degrees of freedom of the high-fidelity discretisation. 
    We compute the singular value decomposition
    \begin{equation}
    \mathbf{S} = \mathbf{U}\boldsymbol{\Sigma}\mathbf{V}^\top,
    \end{equation}
    and select the smallest $r$ such that
    \begin{equation}
    \frac{\sum_{i=1}^r \sigma_i^2}{\sum_{i} \sigma_i^2} \ge 0.95,
    \end{equation}
    where $\sigma_i$ denote the singular values. 
    The resulting POD basis on the full grid is
    \begin{equation}
    \boldsymbol{\Phi} = \mathbf{U}_{[:,1:r]} \in \mathbb{R}^{N \times r}, \qquad r \ll N.
    \end{equation}
    In this case, the above criterion yields $r = 25$ retained modes.

    Since the low-fidelity solvers are defined on coarser spatio-temporal grids, their full-field outputs are first mapped to the high-fidelity spatio-temporal resolution using an interpolation operator
    \begin{equation}    
    \mathcal{I}: \mathbb{R}^{N_l \times T_l} \longrightarrow \mathbb{R}^{N \times T},
    \end{equation}
    where $N_l$ is  the number of spatial degrees of freedom of the corresponding low-fidelity discretisation. The operator $\mathcal{I}$ applies linear interpolation in space and cubic spline resampling in time .
    For a given parameter vector $\boldsymbol{\mu}$ and fidelity level $l$, the reduced-order representation (POD coefficients) of the low-fidelity model is then defined as
    \begin{equation}
    \label{eq:lf-reduced-final}
    \mathbf{z}_{\mathrm{LF}}^{(l)}(\boldsymbol{\mu})
    = \boldsymbol{\Phi}^\top \,
    \mathcal{I}\big(\mathbf{g}_{\mathrm{LF}}^{(l)}(\boldsymbol{\mu})\big)
    \in \mathbb{R}^{r \times T}.
    \end{equation}
    An analogous projection applied to the high-fidelity full-field solution yields the corresponding high-fidelity POD coefficients $\mathbf{z}_{\mathrm{HF}}(\boldsymbol{\mu})$. 
    In this way, all models share a consistent reduced representation in the POD coefficient space.

    \item \textbf{Multi-fidelity NN:} 
    At each level $l=1,2,3$, a multi-fidelity NN $\mathbf{f}_{\mathrm{MF}}^{(l)}$ is trained to predict the high-fidelity POD coefficients using the parameter vector $\boldsymbol{\mu}$ and the reduced outputs from all low-fidelity solvers up to level $l$. 
    The multi-fidelity \ac{NN} acts in the reduced space and has the form
    \begin{equation}
    \hat{\mathbf{g}}_{\mathrm{MF}}^{(l)}: \mathbb{R}^2 \times \left[\mathbb{R}^{r\times T}\right]^l \rightarrow \mathbb{R}^{r\times T},
    \end{equation}
    where the output is the time sequence of POD coefficients associated with the high-fidelity solution.
    The corresponding reconstructed full-field prediction is obtained by
    \begin{equation}
    {\mathbf{g}}_{\mathrm{MF}}^{(l)}(\boldsymbol{\mu}) 
    = \boldsymbol{\Phi} \, \hat{\mathbf{g}}_{\mathrm{MF}}^{(l)}\left(\boldsymbol{\mu}, \left\{ \mathbf{z}_{\mathrm{LF}}^{(j)}(\boldsymbol{\mu}) \right\}_{j=1}^l \right)
    \in \mathbb{R}^{N\times T}.
    \end{equation}
    Evaluating ${\mathbf{g}}_{\mathrm{MF}}^{(l)}(\boldsymbol{\mu})$ at the sensor grid nodes yields the sensor-level surrogate output
    \begin{equation}
    {\mathbf{f}}_{\mathrm{MF}}^{(l)}(\boldsymbol{\mu}) \in \mathbb{R}^{N_s\times T},
    \end{equation}
    which is the quantity used in the likelihood evaluation. 
    This architecture efficiently leverages low-fidelity evaluations to approximate the high-fidelity response, combining model reduction with hierarchical information fusion. 
    It can be regarded as a multi-fidelity extension of the POD–NN framework~\cite{Hesthaven2018}. The architecture details are provided in \ref{app:nn_second_test}.
\end{itemize}

\subsection{MFDA: Offline training and models accuracy}
We sample $N_{train}=500$ parameter instances, denoted by $\{\boldsymbol{\mu}_i\}_{i=1}^{N_{train}}$, using Latin Hypercube Sampling within the admissible parameter domain $\mu$. For each sampled instance, all the high-fidelity and low-fidelity models are evaluated. A dimensionality reduction is applied using $r = 25$ modes. The three multi-fidelity neural networks are then trained using the Adam optimizer by minimizing the MSE:
\begin{equation}
\mathcal{L}(\mathbf{W}_l, \mathbf{b}_l) = \frac{1}{N_{\mathrm{train}}} \sum_{j=1}^{N_{\mathrm{train}}} \left\| \mathbf{f}_{\mathrm{HF}}(\boldsymbol{\mu}_j) - \hat{\mathbf{f}}_\mathrm{MF}^{(l)}\left(\boldsymbol{\mu}_j, \boldsymbol{\Phi}^\top \, \mathcal{I}(\mathbf{f}_\mathrm{LF}^{(1)}(\boldsymbol{\mu}_j)), \ldots, \boldsymbol{\Phi}^\top \, \mathcal{I}(\mathbf{f}_\mathrm{LF}^{(l)}(\boldsymbol{\mu}_j)); \mathbf{W}_l, \mathbf{b}_l\right) \right\|^2, \quad l=1,2,3.
\end{equation}
An additional set of 20 parameter instances is reserved for testing. Table~\ref{tab:model_performance2} reports the computational cost and predictive accuracy, measured in terms of RMSE in the testing set, for each solver and surrogate model. The results indicate that the multi-fidelity NNs consistently achieve significantly lower RMSE compared to any individual low-fidelity model, while maintaining comparable computational costs. The performance gain is particularly notable when using the coarsest solvers: for instance, the surrogate $f_{\mathrm{MF}}^{(1)}$ reduces the RMSE by approximately one order of magnitude relative to its corresponding low-fidelity model $f_{\mathrm{LF}}^{(1)}$.

Figure~\ref{fig:sol2} presents the predicted solutions for the low-fidelity solvers and the multi-fidelity neural networks at all three levels, evaluated at the final time step for the $u$ quantity in a benchmark case. The corresponding errors are presented in Figure~\ref{fig:sol3}. As shown, the multi-fidelity neural networks substantially enhance the reconstruction accuracy. The resulting error is significantly reduced and primarily concentrated in the central region of the domain, indicating the effectiveness of the surrogate models in capturing the underlying dynamics even at coarse resolutions. The source of the errors is mostly related to higher frequencies modes, not included in dimensionality reduction.

\begin{table}[b]
    \renewcommand{\arraystretch}{1.3}
    \centering
    \begin{tabular}{lrrrr}
        \toprule
        \textbf{Model} & \textbf{Mesh Size [\#elements]} & \textbf{Time Steps} & \textbf{Computational Time [s]} & \textbf{RMSE} \\
        \midrule
        $f_{\mathrm{HF}}$        & $128\times128$  & 250  & 19.34  & 0       \\
        $f_{\mathrm{LF}}^{(1)}$  & $16\times16$    & 50   & 0.407  & $1.74\times 10^{-1}$   \\
        $f_{\mathrm{LF}}^{(2)}$  & $32\times32$    & 100   & 0.809  & $8.72\times 10^{-2}$   \\
        $f_{\mathrm{LF}}^{(3)}$  & $64\times64$    & 250   & 3.293  & $1.34\times 10^{-2}$   \\
        \midrule
        $f_{\mathrm{MF}}^{(1)}$  & --              & --    & 0.0588 & $4.0\times 10^{-2}$   \\
        $f_{\mathrm{MF}}^{(2)}$  & --              & --    & 0.5180 & $3.9\times 10^{-2}$   \\
        $f_{\mathrm{MF}}^{(3)}$  & --              & --    & 1.556  & $2.8\times 10^{-2}$   \\
        \bottomrule
    \end{tabular}
    \caption{Computation time and \ac{RMSE} for high-fidelity, low-fidelity, and multi-fidelity surrogate models.}
    \label{tab:model_performance2}
\end{table}

\subsection{MFDA: Online Inference}\label{sec:res3}

Synthetic observations are generated by evaluating the high-fidelity solver at the $13\times13$ sensor locations and adding independent Gaussian noise,
\begin{equation}
\mathbf{y}^{\mathrm{obs}} = \mathbf{f}_{\mathrm{HF}}(\boldsymbol{\mu}) + \boldsymbol{\varepsilon},
\qquad \boldsymbol{\varepsilon} \sim \mathcal{N}(0,\sigma_\varepsilon^2 I_{d}),
\end{equation}
where $\sigma_\varepsilon = 0.2$, measurements are collected every $0.2$ time units. The sensor configuration is shown in Fig.~\ref{fig:sensors}. The aim is to characterize the posterior distribution $\pi(\boldsymbol{\mu}\mid\mathbf{y}^{\mathrm{obs}})$ and compare the sampling efficiency of MFDA with that of MH and MLDA.

In MFDA, the high-fidelity model is used only offline to both construct the POD basis and train the multi-fidelity surrogates. During inference, the likelihood is evaluated exclusively using $\mathbf{f}_{\mathrm{MF}}^{(l)}$, which combines $\boldsymbol{\mu}$ with reduced low-fidelity model outputs. In contrast, MLDA evaluates a hierarchy of low- and high-fidelity solvers online, while MH relies solely on the high-fidelity solver.

The sampling schemes, together with the forward models and sub-chain lengths at each level, are summarized in Table~\ref{tab:online_phase_config_RD}. Sub-chain lengths are chosen so that the number of coarse evaluations per fine-level proposal in MLDA matches that of MFDA. 
Analogously to previous test case, five independent MCMC chains are run, and convergence is monitored every 100 iterations using the Gelman–Rubin diagnostic; sampling is terminated once $\widehat{R}<1.01$ across all components of $\boldsymbol{\mu}$. For all cases, the chains are
initialized by solving a least-squares problem for the observations using the highest-level
model and setting the resulting estimate as the initial parameter sample. The proposal
distribution is a Gaussian random walk with covariance given by a Gauss-Newton approximation
of the least-squares Hessian at this point.
The MCMC initialization and proposal construction follow the same procedure as in
Section~\ref{sec:res1}.
\begin{table}[b!]
\centering
\renewcommand{\arraystretch}{1.2}
\begin{tabular}{llllc}
\toprule
\textbf{Scheme} & \textbf{Level} & \textbf{Forward Model} & \textbf{Inputs} & \textbf{Subchain Length} \\
\midrule
\multirow{3}{*}{MFDA}
    & 1 & $\mathbf{f}_{\mathrm{MF}}^{(1)}$ & $\boldsymbol{\mu}, \mathbf{z}_{\mathrm{LF}}^{(1)}$ & 5 \\
    & 2 & $\mathbf{f}_{\mathrm{MF}}^{(2)}$ & $\boldsymbol{\mu}, \mathbf{z}_{\mathrm{LF}}^{(1)}, \mathbf{z}_{\mathrm{LF}}^{(2)}$ & 5 \\
    & 3 & $\mathbf{f}_{\mathrm{MF}}^{(3)}$ & $\boldsymbol{\mu}, \mathbf{z}_{\mathrm{LF}}^{(1)}, \mathbf{z}_{\mathrm{LF}}^{(2)}, \mathbf{z}_{\mathrm{LF}}^{(3)}$ & -- \\
\midrule
\multirow{4}{*}{MLDA}
    & 1 & $\mathbf{f}_{\mathrm{LF}}^{(1)}$ & $\boldsymbol{\mu}$ & 5 \\
    & 2 & $\mathbf{f}_{\mathrm{LF}}^{(2)}$ & $\boldsymbol{\mu}$ & 5 \\
    & 3 & $\mathbf{f}_{\mathrm{LF}}^{(3)}$ & $\boldsymbol{\mu}$ & 1 \\
    & 4 & $\mathbf{f}_{\mathrm{HF}}$       & $\boldsymbol{\mu}$ & -- \\
\midrule
MH & 1 & $\mathbf{f}_{\mathrm{HF}}$ & $\boldsymbol{\mu}$ & -- \\
\bottomrule
\end{tabular}
\caption{Likelihood evaluation strategies during online inference for the reaction--diffusion test case. MFDA uses POD-based surrogate evaluations with reduced-model inputs, see Eq. \eqref{eq:lf-reduced-final} for definition of $\mathbf{z}_{LF}^{(\cdot)}$; MLDA sequentially filters proposals across low- to high-fidelity solvers; MH directly evaluates the high-fidelity model.}
\label{tab:online_phase_config_RD}
\end{table}

Computed posterior distributions for a given reference value of the parameter vector  are shown in Figure~\ref{fig:post2}. All schemes yield accurate and well localized posterior distributions. Among them, MLDA produces the distribution most closely aligned with the true parameter values, while both MH and MFDA also deliver estimates in close agreement with the ground truth. The corresponding trace plots are reported in Figure~\ref{fig:trace2}. The MFDA scheme exhibits the best mixing properties, whereas MLDA occasionally shows stretches of high autocorrelation, likely arising in regions of the posterior where the coarser models deviate from the fine-level model.

The procedure is repeated over ten independent realizations of the parameter vector $\boldsymbol{\mu}$. Figure~\ref{fig:acc22} reports the distribution of the RMSE between the posterior means and the corresponding true parameters. All methods achieve accurate recovery of the unknown parameters. Among them, MLDA exhibits the lowest estimation error and the smallest variability across realizations. The standard MH sampler achieves slightly higher errors with comparable dispersion. MFDA displays a moderate increase in the error due to the absence of the high fidelity solver in the sampling scheme. Nonetheless, the estimation accuracy of MFDA remains well within the range typically considered acceptable for inverse UQ tasks, with a final RMSE around $4\times10^{-4}$. We notice anyway that the error is one order of magnitude more than MLDA and 4 times more than MH, and the main source of error probably is due to the dimensionality reduction.

Finally, the sampling efficiency of the schemes is investigated in Figure~\ref{fig:sampling_efficiency_1}. The MFDA scheme demonstrates the highest efficiency, achieving approximately a fourfold speedup relative to standard MH. Conversely, MLDA provides no computational gain in this setting and results in the poorest performance, over an order of magnitude less efficient than MFDA. This loss of efficiency can be attributed to the limited accuracy of the coarsest-level model.

To quantify the computational gains, Table~\ref{tab:cost_summary_RD} reports the offline and online wall-clock costs associated with each sampling scheme. The offline cost includes the high-fidelity data generation used to construct the POD basis and the surrogate training phase (MFDA only). The online cost corresponds to the total wall-clock time required to complete one inference run, averaged over the five MCMC chains.
The MFDA scheme achieves a substantial reduction in online cost relative to standard MH, and is significantly more efficient than MLDA. When the offline stage is included, the total cost of MFDA remains lower than both MH and MLDA. As in the previous test case, the offline cost is incurred only once: in settings where inference must be repeated for multiple observational datasets, the effective cost per inference for MFDA reduces to its online cost alone.

Overall, these results indicate that MFDA yields posterior estimates that are statistically consistent with those obtained by the high-fidelity and MLDA samplers, exhibiting only a negligible loss in accuracy. At the same time, MFDA achieves a substantial gain in sampling efficiency.

\begin{table}[t!]
\centering
\renewcommand{\arraystretch}{1.25}
\begin{tabular}{lcccc}
\toprule
\textbf{Scheme} &
\textbf{Data Gen. [s]} &
\textbf{Training [s]} &
\textbf{Online Cost [s]} &
\textbf{Total Cost [s]} \\
\midrule
MH
& $0$ & $0$ 
& $91181 \pm 12115$
& $91181 \pm 12115$ \\

MLDA
& $0$ & $0$ 
& $62591 \pm 12066$
& $62591 \pm 12066$ \\

MFDA
& $14615$
& $13200$
& $9220\pm 2753$
& $38035 \pm 2753$ \\
\bottomrule
\end{tabular}
\caption{Computational cost comparison among sampling schemes for the reaction–diffusion test case.
MFDA incurs a one-time offline cost for POD construction and surrogate training,
but provides a substantially lower online and total inference cost.
Online cost values correspond to mean $\pm$ standard deviation over the independent MCMC chains.}
\label{tab:cost_summary_RD}
\end{table}

\section{Conclusions}
\label{sec:conclusion}

In this work we have presented a novel MFDA framework for Bayesian inverse problems governed by partial differential equations. The proposed approach integrates the hierarchical sampling strategy of MLDA with neural network–based multi-fidelity fusion, enabling the combination of information from multiple low-fidelity solvers to construct enhanced surrogates that approximate high-fidelity model evaluations with high accuracy and low cost. In doing so, the MFDA framework retains the sampling accuracy of high-fidelity models while reducing the computational cost traditionally associated with Markov chain Monte Carlo (MCMC)–based inference in large-scale PDE settings.

From a methodological perspective, the MFDA framework offers three key advances. First, it introduces multi-fidelity neural networks capable of incorporating outputs from multiple low-fidelity models simultaneously, thereby overcoming the limitations of traditional auto-regressive or pairwise multi-fidelity regression schemes. Second, it confines the use of computationally expensive high-fidelity solvers to an offline training phase, making the online sampling stage entirely reliant on low fidelity solvers and neural network corrections. Third, it strongly improves the accuracy of the coarsest levels, allowing to rely on the cheapest models for long sub-chains without diverging too much from fine level posterior distributions.

The performance of MFDA has been demonstrated on two benchmark problems: the reconstruction of transmissivity fields in a groundwater Darcy flow model, and the parameter inference for a nonlinear, time-dependent reaction–diffusion system. Across both case studies, MFDA achieved posterior estimates statistically comparable from those obtained using high-fidelity MH and MLDA samplers, while substantially improving sampling efficiency. In the groundwater flow example, MFDA reduced the computation time per effective sample by factors of approximately five relative to MH and 3 relative to MLDA. In the reaction–diffusion case, MFDA delivered similar gains w.r.t. to MH and much higher w.r.t. to MLDA, despite the added complexity of nonlinear dynamics and high-dimensional spatio-temporal observations. The results also confirmed that the neural network surrogates consistently improved the predictive accuracy of all low-fidelity solvers, with the benefits being most pronounced at coarser resolutions.

The proposed methodology is general and can be applied to a broad class of Bayesian inverse problems involving expensive forward models, provided that lower-fidelity approximations are available. The modular design allows flexibility in the choice of surrogate architectures and subchain configurations, making it extendable to problem-specific constraints and computational budgets. Overall, the MFDA framework provides an effective and scalable tool for accelerating Bayesian inference in PDE-constrained inverse problems, achieving a favorable balance between computational efficiency and posterior accuracy.

Future research directions include the development of adaptive online training strategies to update neural network surrogates during sampling, potentially reducing offline training costs. Additionally, investigating MFDA to settings where low-fidelity models differ in physics or dimensionality, rather than solely in numerical resolution, could further highlight its applicability.

\section*{Code availability}
The source code implementing the MFDA framework and supporting the numerical experiments of this work is available at \url{https://github.com/filippozacchei/MFDA}.

\section*{Acknowledgements}

FZ acknowledges the support of the JRC STEAM STM-Politecnico di Milano agreement. PC and AM acknowledges the PRIN 2022 Project “Numerical approximation of uncertainty quantification problems for PDEs by multi-fidelity methods (UQ-FLY)” (No. 202222PACR), funded by the European Union - NextGenerationEU.

AM acknowledges the project “Dipartimento di Eccellenza” 2023-2027 funded by MUR, the project FAIR (Future Artificial Intelligence Research), funded by the NextGenerationEU program within the PNRR-PE-AI scheme (M4C2, Investment 1.3, Line on Artificial Intelligence), and the Project “Reduced Order Modeling and Deep Learning for the real-time approximation of PDEs (DREAM)” (Starting Grant No. FIS00003154), funded by the Italian Science Fund (FIS) - Ministero dell'Università e della Ricerca. 

AF acknowledges the PRIN 2022 Project “DIMIN- DIgital twins of nonlinear MIcrostructures with iNnovative model-order-reduction strategies” 
(No. 2022XATLT2) funded by the European Union - NextGenerationEU.

The authors acknowledge Luca Caroselli for useful discussions and preliminary testing of the methodology proposed in this paper. 

\bibliography{references.bib}

@article{ZaccheiMEMS2024,
       title = {Neural networks based surrogate modeling for efficient uncertainty quantification and calibration of MEMS accelerometers},
        journal = {International Journal of Non-Linear Mechanics},
        volume = {167},
        pages = {104902},
        year = {2024},
        doi = {https://doi.org/10.1016/j.ijnonlinmec.2024.104902},
        author = {Filippo Zacchei and Francesco Rizzini and Gabriele Gattere and Attilio Frangi and Andrea Manzoni},
    }

@article{LykkegaardGWF2021,
       title={Accelerating uncertainty quantification of groundwater flow modelling using a deep neural network proxy},
       volume={383},
       DOI={10.1016/j.cma.2021.113895},
       journal={Computer Methods in Applied Mechanics and Engineering},
       author={Lykkegaard, Mikkel B. and Dodwell, Tim J. and Moxey, David},
       year={2021},
       pages={113895} 
    }

@article{Hesthaven2018,
    author = {Hesthaven, Jan and Ubbiali, Stefano},
    year = {2018},
    pages = {},
    title = {Non-intrusive reduced order modeling of nonlinear problems using neural networks},
    volume = {363},
    journal = {Journal of Computational Physics},
    doi = {10.1016/j.jcp.2018.02.037}
    }

@article{hastings,
     author = {W. K. Hastings},
     journal = {Biometrika},
     number = {1},
     pages = {97--109},
     title = {Monte Carlo Sampling Methods Using Markov Chains and Their Applications},
     volume = {57},
     year = {1970}
    }

@article{CO2_2023,
    author = {Sugimoto, Saeki and Takakura, Yuya and Kajiro, Hiroshi and Fujiki, Junpei and Dashti, Hossein and Yajima, Tomoyuki and Kawajiri, Yoshiaki},
    year = {2023},
    pages = {},
    title = {Modeling, parameter estimation, and uncertainty quantification for CO2 adsorption process using flexible metal–organic frameworks by Bayesian Monte Carlo methods},
    volume = {5},
    journal = {Journal of Advanced Manufacturing and Processing},
    doi = {10.1002/amp2.10165}
    }

@article{Willcox2008,
    author = {Bui-Thanh, Tan and Wilcox, Karen and Ghattas, Omar},
    year = {2008},
    pages = {2520-2529},
    title = {Parametric Reduced-Order Models for Probabilistic Analysis of Unsteady Aerodynamic Applications},
    volume = {46},
    journal = {AIAA Journal},
    doi = {10.2514/1.35850}
    }

@article{Durr_2023,
       title={Bayesian Calibration of MEMS Accelerometers},
       volume={23},
       DOI={10.1109/jsen.2023.3272907},
       number={12},
       journal={IEEE Sensors Journal},
       author={Dürr, Oliver and Fan, Po-Yu and Yin, Zong-Xian},
       year={2023},
       pages={13319–13326} }

@article{rubin,
    author = {Gelman, Andrew and Rubin, Donald B.},
    title = {{Inference from Iterative Simulation Using Multiple Sequences}},
    volume = {7},
    journal = {Statistical Science},
    number = {4},
    pages = {457 - 472},
    year = {1992},
    doi = {10.1214/ss/1177011136},
    }

@article{Hoffman2011,
    author = {Hoffman, Matthew and Gelman, Andrew},
    year = {2011},
    pages = {},
    title = {The No-U-Turn Sampler: Adaptively Setting Path Lengths in Hamiltonian
    Monte Carlo},
    volume = {15},
    journal = {Journal of Machine Learning Research}
    }

@article{dodwell2015,
    author = {Dodwell, Tim J. and Ketelsen, Christian and Scheichl, Robert and Teckentrup, Aretha L.},
    title = {A Hierarchical Multilevel Markov Chain Monte Carlo Algorithm with  Applications to Uncertainty Quantification in Subsurface Flow},
    journal = {SIAM/ASA Journal on Uncertainty Quantification},
    volume = {3},
    number = {1},
    pages = {1075-1108},
    year = {2015},
    doi = {10.1137/130915005},
    }

@article{Fox2005,
    author = {Christen, J. Andrés and Fox, Colin},
    title = {Markov chain Monte Carlo Using an Approximation},
    journal = {Journal of Computational and Graphical Statistics},
    volume = {14},
    number = {4},
    pages = {795--810},
    year = {2005},
    doi = {10.1198/106186005X76983},
    }

@article{Willcox2018,
    author = {Peherstorfer, Benjamin and Willcox, Karen and Gunzburger, Max},
    title = {Survey of Multifidelity Methods in Uncertainty Propagation, Inference, and Optimization},
    journal = {SIAM Review},
    volume = {60},
    number = {3},
    pages = {550-591},
    year = {2018},
    doi = {10.1137/16M1082469},
    }

@article{Guo_2022,
       title={Multi-fidelity regression using artificial neural networks: Efficient approximation of parameter-dependent output quantities},
       volume={389},
       DOI={10.1016/j.cma.2021.114378},
       journal={Computer Methods in Applied Mechanics and Engineering},
       author={Guo, Mengwu and Manzoni, Andrea and Amendt, Maurice and Conti, Paolo and Hesthaven, Jan S.},
       year={2022},
       pages={114378} }

@article{Meng_2021,
       title={Multi-fidelity Bayesian neural networks: Algorithms and applications},
       volume={438},
       DOI={10.1016/j.jcp.2021.110361},
       journal={Journal of Computational Physics},
       author={Meng, Xuhui and Babaee, Hessam and Karniadakis, George Em},
       year={2021},
       pages={110361} }

@article{rosafalco2024ekf,
    	author = {Rosafalco, Luca and Conti, Paolo and Manzoni, Andrea and Mariani, Stefano and Frangi, Attilio},
    	doi = {10.1016/j.cma.2024.117264},
    	journal = {Computer Methods in Applied Mechanics and Engineering},
    	pages = {117264},
    	title = {{EKF}--{SINDy}: {Empowering} the extended {Kalman} filter with sparse identification of nonlinear dynamics},
    	volume = {431},
    	year = {2024},
    }

@article{tokdar_importance_2010,
    	title = {Importance sampling: a review},
    	volume = {2},
    	doi = {10.1002/wics.56},
    	number = {1},
    	journal = {WIREs Computational Statistics},
    	author = {Tokdar, Surya T. and Kass, Robert E.},
    	year = {2010},
    	pages = {54--60},
    }

@article{lykkegaard_multilevel_2023,
      author = {Lykkegaard, Mikkel Bue and Dodwell, Tim J. and Fox, Colin and Mingas, Grigorios and Scheichl, Robert},
      title = {Multilevel Delayed Acceptance MCMC},
      journal = {SIAM/ASA Journal on Uncertainty Quantification},
      volume = {11},
      number = {1},
      pages = {1--30},
      year = {2023},
      doi = {10.1137/22M1476770},
    }

@article{nelson_control_1987,
    	title = {On control variate estimators},
    	volume = {14},
    	doi = {10.1016/0305-0548(87)90024-4},
    	number = {3},
    	journal = {Computers \& Operations Research},
    	author = {Nelson, Barry L.},
    	month = jan,
    	year = {1987},
    	pages = {219--225},
    }

@article{myers_matrix_1982,
    	title = {Matrix formulation of co-kriging},
    	volume = {14},
    	doi = {10.1007/BF01032887},
    	number = {3},
    	journal = {Journal of the International Association for Mathematical Geology},
    	author = {Myers, Donald E.},
    	year = {1982},
    	pages = {249--257},
    }

@article{perdikaris_multi-fidelity_2015,
      author = {Perdikaris, Paris and Venturi, Daniele and Royset, J. O. and Karniadakis, George Em},
      title = {Multi-fidelity modelling via recursive co-kriging and Gaussian–Markov random fields},
      journal = {Proceedings of the Royal Society A: Mathematical, Physical and Engineering Sciences},
      volume = {471},
      number = {2179},
      pages = {20150018},
      year = {2015},
      doi = {10.1098/rspa.2015.0018},
    }

@article{conti_multi-fidelity_2024,
    	title = {Multi-fidelity reduced-order surrogate modelling},
    	volume = {480},
    	doi = {10.1098/rspa.2023.0655},
    	number = {2283},
    	journal = {Proceedings of the Royal Society A: Mathematical, Physical and Engineering Sciences},
    	author = {Conti, Paolo and Guo, Mengwu and Manzoni, Andrea and Frangi, Attilio and Brunton, Steven L. and Nathan Kutz, J.},
    	year = {2024},
    	pages = {20230655},}

@article{meng_composite_2020,
    	title = {A composite neural network that learns from multi-fidelity data: {Application} to function approximation and inverse {PDE} problems},
    	volume = {401},
    	doi = {10.1016/j.jcp.2019.109020},
    	journal = {Journal of Computational Physics},
    	author = {Meng, Xuhui and Karniadakis, George Em},
    	year = {2020},
    	pages = {109020}
    }

@article{motamed_multi-fidelity_2020,
    	title = {A {Multi}-{Fidelity} {Neural} {Network} {surrogate} {sampling} {method} {for} {Uncertianty} {Quantification}},
    	volume = {10},
    	doi = {10.1615/Int.J.UncertaintyQuantification.2020031957},
    	number = {4},
    	journal = {International Journal for Uncertainty Quantification},
    	author = {Motamed, Mohammad},
    	year = {2020},
    }

@article{liu_multi-fidelity_2019,
    	title = {Multi-{Fidelity} {Physics}-{Constrained} {Neural} {Network} and {Its} {Application} in {Materials} {Modeling}},
    	volume = {141},
    	issn = {1050-0472},
    	doi = {10.1115/1.4044400},
    	number = {121403},
    	urldate = {2024-11-12},
    	journal = {Journal of Mechanical Design},
    	author = {Liu, Dehao and Wang, Yan},
    	month = sep,
    	year = {2019},
    }

@article{bj1080222083,
    author = {Haario, Heikki  and Saksman, Eero and Tamminen, Johanna},
    title = {{An adaptive Metropolis algorithm}},
    volume = {7},
    journal = {Bernoulli},
    number = {2},
    publisher = {Bernoulli Society for Mathematical Statistics and Probability},
    pages = {223 -- 242},
    year = {2001},
    }

@article{girolami2011riemann,
      title={Riemann manifold langevin and hamiltonian monte carlo methods},
      author={Girolami, Mark and Calderhead, Ben},
      journal={Journal of the Royal Statistical Society Series B: Statistical Methodology},
      volume={73},
      number={2},
      pages={123--214},
      year={2011},
      publisher={Oxford University Press}
    }

@article{giles2015multilevel,
      title={Multilevel monte carlo methods},
      author={Giles, Michael B},
      journal={Acta numerica},
      volume={24},
      pages={259--328},
      year={2015},
      publisher={Cambridge University Press}
    }

@article{beskos2017multilevel,
      title={Multilevel sequential monte carlo samplers},
      author={Beskos, Alexandros and Jasra, Ajay and Law, Kody and Tempone, Raul and Zhou, Yan},
      journal={Stochastic Processes and their Applications},
      volume={127},
      number={5},
      pages={1417--1440},
      year={2017},
      publisher={Elsevier}
    }

@article{Rosafalco2021,
      author    = {Rosafalco, Luca and Torzoni, Matteo and Manzoni, Andrea and Mariani, Stefano and Corigliano, Alberto},
      title     = {Online Structural Health Monitoring by Model Order Reduction and Deep Learning Algorithms},
      journal   = {Computers \& Structures},
      volume    = {255},
      pages     = {106604},
      year      = {2021},
      month     = {October},
      doi       = {10.1016/j.compstruc.2021.106604},
      url       = {https://doi.org/10.1016/j.compstruc.2021.106604}
    }

@article{blei2017,
      title     = {Variational Inference: A Review for Statisticians},
      author    = {Blei, David M. and Kucukelbir, Alp and McAuliffe, Jon D.},
      journal   = {Journal of the American Statistical Association},
      volume    = {112},
      number    = {518},
      pages     = {859--877},
      year      = {2017},
      doi       = {10.1080/01621459.2017.1285773}
    }

@article{cliffe2011multilevel,
      title={Multilevel Monte Carlo methods and applications to elliptic PDEs with random coefficients},
      author={Cliffe, K Andrew and Giles, Mike B and Scheichl, Robert and Teckentrup, Aretha L},
      journal={Computing and Visualization in Science},
      volume={14},
      pages={3--15},
      year={2011},
      publisher={Springer}
    }

@article{manzoni2016accurate,
    author ={Manzoni, Andrea and Pagani, Stefano and Lassila, Toni},
      title={Accurate solution of Bayesian inverse uncertainty quantification problems combining reduced basis methods and reduction error models},
      journal={SIAM/ASA Journal on Uncertainty Quantification},
      volume={4},
      number={1},
      pages={380--412},
      year={2016},
      publisher={SIAM}
    }

@article{pagani2017efficient,
      title={Efficient state/parameter estimation in nonlinear unsteady PDEs by a reduced basis ensemble Kalman filter},
      author={Pagani, Stefano and Manzoni, Andrea and Quarteroni, Alfio},
      journal={SIAM/ASA Journal on Uncertainty Quantification},
      volume={5},
      number={1},
      pages={890--921},
      year={2017},
      publisher={SIAM}
    }

@article{madrigal2023analysis,
      title={Analysis of a class of multilevel Markov chain Monte Carlo algorithms based on independent Metropolis--Hastings},
      author={Madrigal-Cianci, Juan P and Nobile, Fabio and Tempone, Raul},
      journal={SIAM/ASA Journal on Uncertainty Quantification},
      volume={11},
      number={1},
      pages={91--138},
      year={2023},
      publisher={SIAM}
    }

@article{perdikaris2017nonlinear,
      title={Nonlinear information fusion algorithms for data-efficient multi-fidelity modelling},
      author={Perdikaris, Paris and Raissi, Maziar and Damianou, Andreas and Lawrence, Neil D and Karniadakis, George Em},
      journal={Proceedings of the Royal Society A: Mathematical, Physical and Engineering Sciences},
      volume={473},
      number={2198},
      pages={20160751},
      year={2017},
      publisher={The Royal Society Publishing}
    }

@article{seelinger2025democratizing,
      title={Democratizing uncertainty quantification},
      author={Seelinger, Linus and Reinarz, Anne and Lykkegaard, Mikkel B and Akers, Robert and Alghamdi, Amal MA and Aristoff, David and Bangerth, Wolfgang and B{\'e}n{\'e}zech, Jean and Diez, Matteo and Frey, Kurt and others},
      journal={Journal of Computational Physics},
      volume={521},
      pages={113542},
      year={2025},
      publisher={Elsevier}
    }

@article{botteghi2022deep,
      title={Deep kernel learning of dynamical models from high-dimensional noisy data},
      author={Botteghi, Nicol{\`o} and Guo, Mengwu and Brune, Christoph},
      journal={Scientific reports},
      volume={12},
      number={1},
      pages={21530},
      year={2022},
      publisher={Nature Publishing Group UK London}
    }

@article{heideman1984,
      author  = {Heideman, Michael T. and Johnson, Don H. and Burrus, C. Sidney},
      title   = {Gauss and the history of the fast Fourier transform},
      journal = {IEEE ASSP Magazine},
      volume  = {1},
      number  = {4},
      pages   = {14--21},
      year    = {1984},
      doi     = {10.1109/MASSP.1984.1162257},
    }

@article{lee2025delayed,
      title={A Delayed Acceptance Auxiliary Variable MCMC for Spatial Models with Intractable Likelihood Function},
      author={Lee, Jong Hyeon and Kim, Jongmin and Lee, Heesang and Park, Jaewoo},
      journal={arXiv preprint arXiv:2504.17147},
      year={2025}
    }

@article{
    champion2019,
    author = {Champion, Kathleen and Lusch, Bethany  and Kutz, J. Nathan   and Brunton, Steven L.  },
    title = {Data-driven discovery of coordinates and governing equations},
    journal = {Proceedings of the National Academy of Sciences},
    volume = {116},
    number = {45},
    pages = {22445-22451},
    year = {2019},
    doi = {10.1073/pnas.1906995116},
    }

@article{schoot2014,
    author = {van de Schoot, Rens and Kaplan, David and Denissen, Jaap and Asendorpf, Jens B. and Neyer, Franz J. and van Aken, Marcel A.G.},
    title = {A Gentle Introduction to Bayesian Analysis: Applications to Developmental Research},
    journal = {Child Development},
    volume = {85},
    number = {3},
    pages = {842-860},
    doi = {https://doi.org/10.1111/cdev.12169},
    eprint = {https://srcd.onlinelibrary.wiley.com/doi/pdf/10.1111/cdev.12169},
    year = {2014}
    }

@article{HU2024112970,
    title = {A MCMC method based on surrogate model and Gaussian process parameterization for infinite Bayesian PDE inversion},
    journal = {Journal of Computational Physics},
    volume = {507},
    pages = {112970},
    year = {2024},
    doi = {https://doi.org/10.1016/j.jcp.2024.112970},
    author = {Zheng, Hu and Hongqiao, Wang and Qingping, Zhou},
    }

@article{cui2014,
      author  = {Cui, Tiangang and Martin, Jonathan and Marzouk, Youssef M. and Solonen, Antti and Spantini, Alessio},
      title   = {Likelihood-informed dimension reduction for nonlinear inverse problems},
      journal = {Inverse Problems},
      volume  = {30},
      number  = {11},
      pages   = {114015},
      year    = {2014},
      month   = {oct},
      doi     = {10.1088/0266-5611/30/11/114015},
      url     = {https://doi.org/10.1088/0266-5611/30/11/114015},
      publisher = {IOP Publishing},
    }

@article{Behrou,
        author = {Behrou, Reza and Mansourifar, Hadi and Zhou, Yuqing and Wang, Siwen and Schmalenberg, Paul D. and Ling, Chen and Dede, Ercan M.},
        title = {Physics-informed multi-fidelity surrogate modeling of fluid flow in porous media},
        journal = {APL Machine Learning},
        volume = {3},
        number = {3},
        pages = {036116},
        year = {2025},
        doi = {10.1063/5.0279064},
    }

@article{CLEEMAN2023118125,
    title = {Partial-physics-informed multi-fidelity modeling of manufacturing processes},
    journal = {Journal of Materials Processing Technology},
    volume = {320},
    pages = {118125},
    year = {2023},
    doi = {https://doi.org/10.1016/j.jmatprotec.2023.118125},
    author = {Jeremy, Cleeman and Kian, Agrawala and Evan, Nastarowicz and Rajiv Malhotra},
    }

@article{Ebers,
    author = {Ebers, Megan R. and Steele, Katherine M. and Kutz, J. Nathan},
    title = {Discrepancy Modeling Framework: Learning Missing Physics, Modeling Systematic Residuals, and Disambiguating between Deterministic and Random Effects},
    journal = {SIAM Journal on Applied Dynamical Systems},
    volume = {23},
    number = {1},
    pages = {440-469},
    year = {2024},
    doi = {10.1137/22M148375X},
    }

@article{KIM2023109163,
    title = {Multi-fidelity approach for transitional boundary layer},
    journal = {International Journal of Heat and Fluid Flow},
    volume = {102},
    pages = {109163},
    year = {2023},
    issn = {0142-727X},
    doi = {https://doi.org/10.1016/j.ijheatfluidflow.2023.109163},
    author = {Kim, Minwoo and Lim, Jiseop and Jee, Solkeun and Park, Donghun},}

@article{Shahriari2016TakingHumanOutBO,
      author  = {Shahriari, Bobak and Swersky, Kevin and Wang, Ziyu and Adams, Ryan P. and de Freitas, Nando},
      title   = {Taking the Human Out of the Loop: A Review of {B}ayesian Optimization},
      journal = {Proceedings of the IEEE},
      volume  = {104},
      number  = {1},
      pages   = {148--175},
      year    = {2016}
    }

@article{conti2025progressive,
      title={Progressive multi-fidelity learning for physical system predictions},
      author={Conti, Paolo and Guo, Mengwu and Frangi, Attilio and Manzoni, Andrea},
      journal={arXiv preprint arXiv:2510.13762},
      year={2025}
    }

@misc{krouglova2025multifidelity,
    Author        = {Anastasia N. Krouglova and Hayden R. Johnson and Basile Confavreux and Michael Deistler and Pedro J. Gonçalves},
    Title         = {Multifidelity Simulation-based Inference for Computationally Expensive Simulators},
    Eprint        = {2502.08416v3},
    ArchivePrefix = {arXiv},
    PrimaryClass  = {stat.ML},
    Year          = {2025},
    Month         = {Feb},
    Url           = {http://arxiv.org/abs/2502.08416v3}
    }

@misc{betancourt2018,
          title={A Conceptual Introduction to Hamiltonian Monte Carlo}, 
          author={Michael Betancourt},
          year={2018},
          eprint={1701.02434},
          archivePrefix={arXiv},
          primaryClass={stat.ME},
          url={https://arxiv.org/abs/1701.02434}, 
    }

@book{Kaipio2005,
      title     = {Statistical and Computational Inverse Problems},
      author    = {Jari P. Kaipio and Erkki Somersalo},
      series    = {Applied Mathematical Sciences},
      edition   = {1},
      publisher = {Springer New York, NY},
      year      = {2005},
      doi       = {10.1007/b138659},
      isbn      = {978-0-387-22073-4},
      pages     = {XVI, 340},
      url       = {https://doi.org/10.1007/b138659},
      note      = {Mathematics and Statistics, Mathematics and Statistics (R0)},
      eisbn     = {978-0-387-27132-3},
      issn      = {0066-5452},
      eissn     = {2196-968X},
    }

@book{tarantola2005,
    author = {Tarantola, Albert},
    title = {Inverse Problem Theory and Methods for Model Parameter Estimation},
    publisher = {Society for Industrial and Applied Mathematics},
    year = {2005},
    doi = {10.1137/1.9780898717921},
    address = {},
    edition   = {},
    URL = {https://epubs.siam.org/doi/abs/10.1137/1.9780898717921},
    eprint = {https://epubs.siam.org/doi/pdf/10.1137/1.9780898717921}
    }

@book{gelman2013bayesian,
      title        = {Bayesian Data Analysis},
      edition      = {3},
      author       = {Gelman, Andrew and Carlin, John B. and Stern, Hal S. and Dunson, David B. and Vehtari, Aki and Rubin, Donald B.},
      year         = {2013},
      publisher    = {Chapman and Hall/CRC},
      doi          = {10.1201/b16018}
    }

@book{doucet2001,
      title     = {Sequential Monte Carlo Methods in Practice},
      author    = {Doucet, Arnaud and de Freitas, Nando and Gordon, Neil},
      publisher = {Springer},
      year      = {2001},
      isbn      = {9780387951461}
    }

@book{bratley_guide_1987,
    	address = {New York, NY},
    	title = {A {Guide} to {Simulation}},
    	copyright = {http://www.springer.com/tdm},
    	isbn = {978-1-4612-6457-6 978-1-4419-8724-2},
    	url = {http://link.springer.com/10.1007/978-1-4419-8724-2},
    	language = {en},
    	urldate = {2024-11-12},
    	publisher = {Springer},
    	author = {Bratley, Paul and Fox, Bennett L. and Schrage, Linus E.},
    	year = {1987},
    	doi = {10.1007/978-1-4419-8724-2},
    }

@book{adler2007random,
      title={Random fields and geometry},
      author={Adler, Robert J and Taylor, Jonathan E},
      year={2007},
      publisher={Springer}
    }

@book{Loeve1978,
      author    = {Lo\`eve, Michel},
      title     = {Probability Theory. Volume II},
      series    = {Graduate Texts in Mathematics},
      volume    = {46},
      edition   = {4},
      publisher = {Springer-Verlag},
      address   = {New York},
      year      = {1978},
      isbn      = {978-0-387-90262-3}
    }

@book{hammersley_monte_1964,
    	address = {Dordrecht},
    	title = {Monte {Carlo} {Methods}},
    	copyright = {http://www.springer.com/tdm},
    	isbn = {978-94-009-5821-0 978-94-009-5819-7},
    	url = {http://link.springer.com/10.1007/978-94-009-5819-7},
    	language = {en},
    	publisher = {Springer Netherlands},
    	author = {Hammersley, John M. and Handscomb, David C.},
    	year = {1964},
    	doi = {10.1007/978-94-009-5819-7},
    }

@book{kirsch_introduction_2021,
    	address = {Cham},
    	series = {Applied {Mathematical} {Sciences}},
    	title = {An {Introduction} to the {Mathematical} {Theory} of {Inverse} {Problems}},
    	volume = {120},
    	language = {en},
    	publisher = {Springer International Publishing},
    	author = {Kirsch, Andreas},
    	year = {2021},
    	doi = {10.1007/978-3-030-63343-1},
    }

@book{robert2011short,
      title        = {Monte Carlo Statistical Methods},
      author       = {Christian P. Robert and George Casella},
      edition      = {2},
      series       = {Springer Texts in Statistics},
      publisher    = {Springer},
      address      = {New York, NY},
      year         = {2004},
      doi          = {10.1007/978-1-4757-4145-2},
      isbn         = {978-0-387-21239-5},
      isbn10soft   = {978-1-4419-1939-7},
      isbn10ebook  = {978-1-4757-4145-2},
      note         = {Springer Science+Business Media New York},
    }

@misc{tensorflow2015-whitepaper,
    title={ {TensorFlow}: Large-Scale Machine Learning on Heterogeneous Systems},
    url={https://www.tensorflow.org/},
    note={Software available from tensorflow.org},
    author={
        Mart\'{i}n~Abadi and
        Ashish~Agarwal and
        Paul~Barham and
        Eugene~Brevdo and
        Zhifeng~Chen and
        Craig~Citro and
        Greg~S.~Corrado and
        Andy~Davis and
        Jeffrey~Dean and
        Matthieu~Devin and
        Sanjay~Ghemawat and
        Ian~Goodfellow and
        Andrew~Harp and
        Geoffrey~Irving and
        Michael~Isard and
        Yangqing Jia and
        Rafal~Jozefowicz and
        Lukasz~Kaiser and
        Manjunath~Kudlur and
        Josh~Levenberg and
        Dandelion~Man\'{e} and
        Rajat~Monga and
        Sherry~Moore and
        Derek~Murray and
        Chris~Olah and
        Mike~Schuster and
        Jonathon~Shlens and
        Benoit~Steiner and
        Ilya~Sutskever and
        Kunal~Talwar and
        Paul~Tucker and
        Vincent~Vanhoucke and
        Vijay~Vasudevan and
        Fernanda~Vi\'{e}gas and
        Oriol~Vinyals and
        Pete~Warden and
        Martin~Wattenberg and
        Martin~Wicke and
        Yuan~Yu and
        Xiaoqiang~Zheng},
      year={2015},
    }

@article{alnaes_fenics_2015,
    	title = {The {FEniCS} {Project} {Version} 1.5},
    	volume = {3},
    	copyright = {Copyright (c) 2015 Archive of Numerical Software},
    	issn = {2197-8263},
    	doi = {10.11588/ans.2015.100.20553},
    	language = {en},
    	number = {100},
    	journal = {Archive of Numerical Software},
    	author = {Alnæs, Martin and Blechta, Jan and Hake, Johan and Johansson, August and Kehlet, Benjamin and Logg, Anders and Richardson, Chris and Ring, Johannes and Rognes, Marie E. and Wells, Garth N.},
    	month = dec,
    	year = {2015},
    	note = {Number: 100}
    }

@incollection{geyerIntroductionMarkovChain2011,
      title = {Introduction to {M}arkov {C}hain {M}onte {C}arlo},
      booktitle = {Handbook of {M}arkov {C}hain {M}onte {C}arlo},
      author = {Geyer, Charles J.},
      year = {2011},
      publisher = {{Chapman and Hall/CRC}},
      doi={10.1201/b10905-2}
    }

@incollection{Simon2006,
      author    = {Dan Simon},
      title     = {Nonlinear Kalman Filtering},
      booktitle = {Optimal State Estimation: Kalman, H Infinity, and Nonlinear Approaches},
      publisher = {John Wiley \& Sons, Ltd},
      year      = {2006},
      chapter   = {13},
      pages     = {393--431},
      isbn      = {9780470045343},
      doi       = {10.1002/0470045345.ch13},
      url       = {https://onlinelibrary.wiley.com/doi/10.1002/0470045345.ch13},
      note      = {Accessed via Wiley Online Library}
    }

@incollection{annels_geostatistical_1991,
    	address = {Dordrecht},
    	title = {Geostatistical {Ore}-reserve {Estimation}},
    	isbn = {978-94-011-9714-4},
    	language = {en},
    	urldate = {2024-11-12},
    	booktitle = {Mineral {Deposit} {Evaluation}: {A} practical approach},
    	publisher = {Springer Netherlands},
    	author = {Annels, Alwyn E.},
    	editor = {Annels, Alwyn E.},
    	year = {1991},
    	doi = {10.1007/978-94-011-9714-4_4},
    	pages = {175--245},
    }

@inproceedings{kingma2015adam,
      title        = {Adam: A Method for Stochastic Optimization},
      author       = {Kingma, Diederik P. and Ba, Jimmy},
      booktitle    = {Proceedings of the 3rd International Conference on Learning Representations (ICLR)},
      year         = {2015}
    }

@inproceedings{Akiba2019Optuna,
      author    = {Akiba, Takuya and Sano, Shotaro and Yanase, Toshihiko and Ohta, Takeru and Koyama, Masanori},
      title     = {Optuna: A Next-generation Hyperparameter Optimization Framework},
      booktitle = {Proceedings of the 25th ACM SIGKDD International Conference on Knowledge Discovery \& Data Mining},
      pages     = {2623--2631},
      year      = {2019}
    }

@misc{tinyda,
  title        = {TinyDA},
  author       = {{TinyDA developers}},
  howpublished = {\url{https://github.com/tiny-DA/tinyDA}},
  year         = {2024},
  note         = {Python library for (multi-level) delayed-acceptance MCMC and related sampling methods},
}

@article{alarra2023,
doi = {10.1088/2632-2153/acca60},
year = {2023},
month = {apr},
publisher = {IOP Publishing},
volume = {4},
number = {2},
pages = {025012},
author = {Larra{\~n}aga, Ana and Brunton, Steven L. and Mart{\'\i}nez, Javier and Chapela, Sergio and Porteiro, Jacobo},
title = {Data-driven prediction of the performance of enhanced surfaces from an extensive CFD-generated parametric search space},
journal = {Machine Learning: Science and Technology},
}

@article{cao2025derivative,
  title={Derivative-informed neural operator acceleration of geometric MCMC for infinite-dimensional Bayesian inverse problems},
  author={Cao, Lianghao and O'Leary-Roseberry, Thomas and Ghattas, Omar},
  journal={Journal of Machine Learning Research},
  volume={26},
  number={78},
  pages={1--68},
  year={2025}
}

@article{OLEARYROSEBERRY2024112555,
title = {Derivative-Informed Neural Operator: An efficient framework for high-dimensional parametric derivative learning},
journal = {Journal of Computational Physics},
volume = {496},
pages = {112555},
year = {2024},
issn = {0021-9991},
doi = {https://doi.org/10.1016/j.jcp.2023.112555},
author = {Thomas O'Leary-Roseberry and Peng Chen and Umberto Villa and Omar Ghattas},
}

\appendix

\section{Neural Network Architecture Selection for the Groundwater Flow Test Case}
\label{app:nn_architectures}

A comparative analysis of neural network architectures was performed for the test cases in Section~\ref{sec:res1}, using 16\,000 samples and focusing on $\mathbf{f}_{\mathrm{MF}}^{(3)}$.  
Because of the multi-input, multi-fidelity setting, the search was restricted to architectures with one branch per input source: each branch processes a single input (e.g.\ equation parameters or low-fidelity data), the resulting latent representations are concatenated in a fusion layer, and a final fully connected output block maps the concatenated representation to the output.

Architectures within this family were compared using Bayesian optimisation~\cite{Shahriari2016TakingHumanOutBO} as implemented in Python package \textit{Optuna}~\cite{Akiba2019Optuna}.  
Each candidate model was trained for a fixed duration of one minute, favouring compact and computationally efficient networks.  
The architectures for $\mathbf{f}_{\mathrm{MF}}^{(1)}$ and $\mathbf{f}_{\mathrm{MF}}^{(2)}$ were obtained by reusing the same design as for $\mathbf{f}_{\mathrm{MF}}^{(3)}$, while removing the unused low-fidelity input branches.

The hyperparameter search varied the number of hidden layers in each input branch and in the shared/output branch between 0 and 6.  
The special case with 0 hidden layers in all input branches corresponds to a standard fully connected network acting on the concatenation of all inputs, without any branch structure.  
The number of neurons per layer and the activation function were explored over \(\{32, 64, 128, 256\}\) and \(\{\text{GeLU}, \text{tanh}, \text{ReLU}, \text{sigmoid}\}\), respectively, and a regularisation coefficient was also tuned.

The final architecture for $\mathbf{f}_{\mathrm{MF}}^{(3)}$ employs a high-fidelity branch (Input~1) with four fully connected layers of 128 neurons, and three additional low-fidelity branches (Inputs~2--4), each mapped to 128 neurons.  
The resulting latent representations are concatenated and passed through a shared processing block with two fully connected layers and a 25-neuron linear output layer.

The complete configuration of this architecture is reported in Table~\ref{tab:gw_fmf3}.  
The networks used for $\mathbf{f}_{\mathrm{MF}}^{(1)}$ and $\mathbf{f}_{\mathrm{MF}}^{(2)}$ are obtained from the same structure by retaining only the branches associated with their available inputs.

\begin{table}[h!]
\centering
\begin{tabular}{llll}
\toprule
\textbf{Component} & \textbf{Layer type} & \textbf{Neurons} & \textbf{Activation} \\
\midrule
Parameter branch ($\boldsymbol{\theta}$) 
    & Input & 64  & --     \\
    & Dense & 128 & GeLU   \\
    & Dense & 128 & GeLU   \\
    & Dense & 128 & GeLU   \\
    & Dense & 128 & GeLU   \\
    & Dense & 128 & Linear \\[0.3em]
\midrule
Low-fidelity branch 1 ($\mathbf{f}_{\mathrm{LF}}^{(1)}$) 
    & Input & 25  & --     \\
    & Dense & 128 & Linear \\[0.3em]
\midrule
Low-fidelity branch 2 ($\mathbf{f}_{\mathrm{LF}}^{(2)}$) 
    & Input & 25  & --     \\
    & Dense & 128 & Linear \\[0.3em]
\midrule
Low-fidelity branch 3 ($\mathbf{f}_{\mathrm{LF}}^{(3)}$) 
    & Input & 25  & --     \\
    & Dense & 128 & Linear \\[0.3em]
\midrule
Fusion layer          
    & Concatenation & 512 & -- \\[0.3em]
\midrule
Output block        
    & Dense   & 128 & GeLU   \\
    & Dense   & 128 & GeLU   \\  
    & Dense & 25 & Linear \\
\bottomrule
\end{tabular}
\caption{Schematic architecture of the multi-fidelity \ac{NN} $\mathbf{f}_{\mathrm{MF}}^{(3)}$ for the groundwater flow test case. 
The fusion layer concatenates the outputs of the input branches, which are then processed by the output block to produce the final prediction.}
\label{tab:gw_fmf3}
\end{table}

\section{Neural Network Architecture for the Reaction–Diffusion Test Case}
\label{app:nn_second_test}

For the results in Section~\ref{sec:res2}, a hand-tuning phase was performed, building on the insights from the groundwater flow case.  
The goal was not to identify a single globally optimal architecture, but to show that a limited, targeted tuning effort already leads to substantial accuracy gains for the low-fidelity models without excessive offline cost.

All architectures process sequential inputs with a length equal to the number of time steps.   
All the branches are concatenated and further processed by a fully connected layer, followed by two stacked LSTM layers and two additional fully connected layers.  
The network concludes with a fully connected output layer.

The detailed architecture for $\mathbf{f}_{\mathrm{MF}}^{(3)}$ (using all inputs) is reported in Table~\ref{tab:rd_fmf3}.  
The models for $\mathbf{f}_{\mathrm{MF}}^{(1)}$ and $\mathbf{f}_{\mathrm{MF}}^{(2)}$ share the same design, restricted to the branches corresponding to their available inputs.

\begin{table}[h!]
\centering
\begin{tabular}{llll}
\toprule
\textbf{Component} & \textbf{Layer type} & \textbf{Neurons} & \textbf{Activation} \\
\toprule
Parameter branch ($\boldsymbol{\mu}$)  & Input & 3  & --    \\

                 & Dense & 64 & GeLU  \\
                 & Dense & 64 & GeLU  \\
                 & Dense & 64 & GeLU  \\
                 & Dense & 64 & Linear  \\[0.3em]
\midrule
Low-fidelity branch 1 ($\mathbf{z}_{\mathrm{LF}}^{(1)}$)  & Input & 25 & --    \\

                      & Dense & 64 & Linear  \\[0.3em]
\midrule
Low-fidelity branch 2 ($\mathbf{z}_{\mathrm{LF}}^{(2)}$) & Input & 25 & --    \\
 
                      & Dense & 64 & Linear  \\[0.3em]
\midrule
Low-fidelity branch 3 ($\mathbf{z}_{\mathrm{LF}}^{(3)}$) & Input & 25 & --    \\
 
                      & Dense & 64 & Linear  \\[0.3em]
\midrule
Fusion layer          & Concatenation & 256 & --    \\
\midrule
Output block        & Dense & 64 & GeLU  \\
                      & LSTM  & 64 & Tanh    \\
                      & LSTM  & 64 & Tanh    \\
                      & Dense & 64 & GeLU  \\
          & Dense & 25 & Linear \\
\bottomrule
\end{tabular}
\caption{Schematic architecture of the multi-fidelity \ac{NN} $\mathbf{f}_{\mathrm{MF}}^{(3)}$ for the reaction--diffusion test case. 
The fusion layer concatenates the outputs of the input branches, which are then processed by the output block to produce the final prediction.}
\label{tab:rd_fmf3}
\end{table}

\end{document}